\documentclass[12pt,letterpaper]{article}
\usepackage[a4paper, total={7in, 10in}]{geometry}

\usepackage{graphicx}
\usepackage{helvet}
\usepackage{authblk}
\usepackage{hyperref}
\usepackage{amsmath} 
\usepackage{amssymb} 
\usepackage{orcidlink} 

\usepackage{booktabs} % new
\usepackage{longtable}
\usepackage{tikz}
\usepackage{svg}
\usepackage{float}
\usepackage{listings}
\usepackage{subcaption}
\usepackage{tablefootnote}
\usepackage{multirow}
\usepackage{shellesc}
\usepackage{comment}
\definecolor{mygreen}{RGB}{28,172,28}
\definecolor{mylilas}{RGB}{170,55,241}
\definecolor{Blue}{rgb}{0,0,0.5}
\definecolor{Green}{rgb}{0,0.75,0.0}
\definecolor{LightGray}{rgb}{0.6,0.6,0.6}
\definecolor{DarkGray}{rgb}{0.3,0.3,0.3}

\usepackage{hyperref}       % hyperlinks

\usepackage[super,comma,sort&compress]  
   {natbib}
\makeatletter
\renewcommand{\maketitle}{\bgroup\setlength{\parindent}{0pt}
\begin{flushleft}
  \textbf{\@title}
  
  \@author
\end{flushleft}\egroup}
\makeatother

\title{Evaluating Large Language Models Beyond Textual Understanding with ChildPlay}
\date{}

\author[1,2,\orcidlink{0000-0001-8776-4852}]{Gonçalo Carvalho}
\author[1,2]{Oscar Knap}
\author[1,2,*,\orcidlink{0000-0001-8836-6266}]{Robert Pollice}

\affil[1]{Stratingh Institute for Chemistry, University of Groningen, Nijenborgh 3, 9747 AG Groningen, The Netherlands.}
\affil[2]{These authors contributed equally}

\affil[*]{Correspondence: r.pollice@rug.nl}

\begin{document}

%\begin{comment}

\maketitle

\section*{Abstract}
We developed a benchmark set to assess the generalization of state-of-the-art large language models on problems beyond linguistic tasks and evaluate it on a systematic progression of GPT models (GPT-3.5, GPT-4, GPT-4o, GPT-4o-mini). Using simple games like Tic-Tac-Toe, Connect Four, Battleship, and a Shape Recognition Game, all encoded in ASCII, we test strategic capabilities and spatial reasoning, core abilities any artificial intelligence would need to master for solving problems in chemistry. To probe generalization, we introduce two new games for spatial logic: LEGO Connect Language (LCL) and Guess-the-SMILES (GtS), a operationally simple chemistry benchmark. Our results show that GPT models provide meaningful responses for several tasks but, generally, perform poorly. A systematic performance progression with increased model capabilities (GPT-3.5, GPT-4, GPT-4o) is only observed for 4 out of the 7 benchmark tasks. All models consistently struggle with Battleship, LCL, and GtS. This suggests that while GPT models can emulate conversational proficiency and basic rule comprehension, they have limited generalization with respect to strategy and spatial reasoning. Particularly poor performance is observed for interpreting molecular graphs when encoded in ASCII. The results provided by our open-source benchmark suite (\href{https://github.com/BlueVelvetSackOfGoldPotatoes/child-play}{\texttt{ChildPlay} GitHub Repository}) caution against claims of emergent intelligence in GPT models, which appear more specialized than general.

\section*{Keywords}
Large Language Models, Benchmarking, Non-linguistic Reasoning, Zero-shot Learning, Molecular graphs, SMILES

\section*{Introduction}
Current Large Language Models (LLMs) are transformer-based. They process input and generate output coherently and contextually appropriate \cite{brown}. They use self-attention to weigh the relative importance of different words in a sentence \cite{vaswani2023attention, brown}. Input text is tokenized, converted into vectors using embeddings, and processed through transformer layers that calculate attention scores to focus on relevant tokens \cite{vaswani2023attention, brown, Fields2024ASO}. The model then selects the next token based on learned distributions \cite{vaswani2023attention, brown, Fields2024ASO}. With their large parameter count, from Alpaca (7~billion) \cite{alpaca} to LLaMA (65~billion) \cite{touvron2023llama} or PaLM (540~billion) \cite{chowdhery2022palm}, these models have learned complex linguistic abstractions, capturing patterns in syntax, semantics, pragmatics, and elements of both style and tone \cite{brown, Bubeck2023SparksOA, ouyang2022training}.

\textbf{Current Benchmarks} \quad Benchmark datasets like \texttt{SQuAD}, \texttt{GLUE}, \texttt{BIG-bench}, \texttt{SuperGLUE}, \texttt{ANLI}, \texttt{TruthfulQA}, \texttt{HellaSwag}, and \texttt{lm-evaluation-harness} offer various forms of assessment, including multiple-choice questions, reading comprehension, natural language understanding, common sense reasoning, factual knowledge, and dialogue completion \cite{rajpurkar2016squad, wang2019glue, srivastava2023beyond, wang2019glue, nie2020adversarial, lin2022truthfulqa, zellers2019hellaswag, eleutherai}.

In addition, recent studies have tested LLMs in non-linguistic tasks. Liga and Pasetto used Tic-Tac-Toe in ASCII form, pitting LLMs against the minimax algorithm to explore emergent features, suggested to resemble consciousness \cite{Liga2023TestingSR}. While performing respectably, LLMs still often only achieved draws or lost \cite{Liga2023TestingSR}. Topsakal and Harper \cite{Topsakal2024BenchmarkingLL} found GPT-4 to win more often than GPT-3.5 but still not to play in an optimal manner. Some studies, as those by Lappin et al. and Zečević et al., caution against overestimating the capabilities of LLMs, stating they may excel at language but seem to struggle with reasoning and causal understanding \cite{Lappin2023Strengths, Zecevic2023Causal}. Both Bender et al. and Schaeffer et al. critique the transparency of evaluation metrics, arguing that some so-called emergent abilities may be artifacts of evaluation rather than genuine behavior \cite{bender2021dangers, schaeffer2023mirage}.

Another issue in evaluating LLMs is training-test set cross-contamination \cite{brown}. Massive training sets, often untraceable, make replication difficult and risk rendering benchmarks meaningless due to contamination \cite{brown, Floridi2020GPT3IN}. N-Gram Overlap is one method to detect contamination but has limitations, such as failing to account for context and synonyms \cite{blodgett2020language}. %Furthermore, the arbitrary 200-character window used during GPT-3.5 training may not accurately reflect the influence of surrounding text \cite{brown}.

This critique of evaluation metrics invites exploration of general intelligence and how it can be reliably measured through rigorous and realistic tests beyond linguistics that include broader cognitive functions. For general intelligence (GI), the so-called g-factor quantifies the ability to reason, plan, solve problems, think abstractly, and learn quickly across many domains \cite{spearman1904general, binet1916development, wechsler1939measurement, cattell1963theory, carroll1993human}. GI then involves higher-order cognitive-equivalent processes beyond specific skills or knowledge domains \cite{gottfredson1997why, jensen1998factor}. %Hence, nuance is needed in current debates on Artificial General Intelligence (AGI) and a pragmatic approach on understanding capabilities of LLMs. 
To approximate measurements of GI in an AI system, it is key to build benchmarks that gauge generalization and reasoning, rather than relying only on pattern matching and statistical correlations \cite{vanDijk2023Nuance}.

\textbf{This Work} \quad We present \texttt{ChildPlay}, a scalable suite of non-language games comprising Tic-Tac-Toe, Connect-Four, Battleship, Lego Connect Language, and Shape Recognition. It also includes the hidden chemistry task Guess-the-SMILES (GtS), which probes the translation of ASCII depictions of chemical graphs into the SMILES representation. For this last task, the source code is not provided to obviate future model overfitting. These games assess LLM strategy, symbolic reasoning, spatial logic, and pattern recognition beyond linguistics. Notably, these are all aspects that are particularly relevant for applying LLMs in the context of chemistry when structure-property relationships need to be accounted for. %Completing all benchmarks only takes a few hours.
We chose games because they offer a simple structure with clear success criteria that is suitable for evaluating strategic thinking and decision-making in dynamic, adversarial scenarios \cite{srivastava2022beyond, Liga2023TestingSR, Topsakal2024BenchmarkingLL}. All games are ASCII-encoded to minimize dataset contamination\cite{brown, Liga2023TestingSR}. The game rules are provided in the prompts, which is critical to evaluate how well LLMs can generalize from instructions. We also proposed \texttt{ChildPlay} for inclusion in BIG-bench \cite{srivastava2023beyond}, which aims to standardize challenging LLM benchmarks. 

%While this does not guarantee that models are applying these rules as a human would, it still probes generalization ability. While our benchmarks attempt to probe the understanding of models indirectly, gameplay is inherently limited by reliance on probabilistic patterns rather than genuine reasoning, something currently contentious which we attempt to probe. 

\section*{Methods}
\textbf{General Methodology} \quad Tasks in BIG-bench \cite{srivastava2023beyond} are categorized as zero-, one-, or multi-shot \cite{brown}. Our tasks, which are available at the \href{https://github.com/BlueVelvetSackOfGoldPotatoes/child-play}{\texttt{ChildPlay} Repository},\cite{childplay} are all zero-shot. Accordingly, models get an explanation at inference time with no examples. %To test LLMs' beyond training data, we tested spatial reasoning with ASCII. An agent capable of abstraction should interpret these sequences based on the provided rules.

%\texttt{ChildPlay} encompasses simple games (Tic-Tac-Toe, Connect-Four, Battleship, Shape Recognition) and spatial reasoning. 
To study a systematic series, we tested \texttt{gpt-3.5-turbo-1106} (GPT-3.5), \texttt{gpt-4-1106-preview} (GPT-4), \texttt{gpt-4o-2024-08-06} (GPT-4o), and \texttt{gpt-4o-mini-2024-07-18} (GPT-4o-mini) across multiple temperatures (cf. Supporting Information, Section~\ref{app:others} for preliminary results with other models). When queried, all the tested models explained the tasks and generated valid board states. In addition, they also explained optimal strategies when prompted (cf. Supporting Information, Section~\ref{app:prompts_optimal}). Unless noted otherwise, error bars are based on the binomial distribution and represent the corresponding standard deviation.

\textbf{Board Games: Battleship, Tic-Tac-Toe, and Connect-Four (BTC)} \quad For the BTC games, each board state was introduced with a game explanation. The LLMs, always playing as the first player, received the current board state and faced an opponent making random moves, which corresponds to a weak opponent. As these games are fully observable, the initial instruction and the current state allow for optimal play. Inputs are two scalars for the row-column pair or a single scalar for the column in Connect-Four. In Battleship, ships ('S') were randomly initialized horizontally, varying from 2 to 5 cells in size. Hits were marked with an 'X' and misses with an 'O'. The symbol '\text{$\sim$}' denoted an unexplored square.

\begin{figure}[H]
    \centering
    \begin{subfigure}[t]{0.23\linewidth}
        \centering
        \fbox{\includegraphics[width=0.5\linewidth]{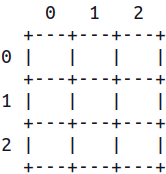}}
        %\caption{Tic-tac-toe board.}
        \label{fig:tic-tac-toe}
    \end{subfigure}%
    \hspace{10pt} % <-- Add horizontal space between subfigures
    \begin{subfigure}[t]{0.23\linewidth}
        \centering
        \fbox{\includegraphics[width=0.5\linewidth]{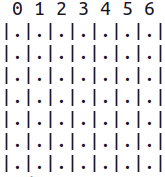}}
        %\caption{Connect-four board.}
        \label{fig:connect-four}
    \end{subfigure}
    \hspace{10pt}
    \begin{subfigure}[t]{0.45\linewidth}
        \centering
        \fbox{\includegraphics[width=0.6\linewidth]{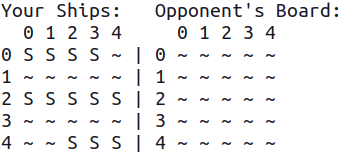}}
        %\caption{Battleship board.}
        \label{fig:battleship}
    \end{subfigure}
    \caption{Initial board states presented to LLMs. Left: Tic-Tac-Toe board. Middle: Connect-Four board. Right: Battleship board. Ship placements are randomized with every initialization.}
    \label{fig:board_games}
\end{figure}

\begin{figure}[H]
    \centering
    % \begin{subfigure}[t]{0.20\linewidth}
    %     \centering
    %     \fbox{\includegraphics[width=\linewidth]{shapes_circle.png}}
    %     \caption{The circle.}
    %     \label{fig:circle}
    % \end{subfigure}%
    % \hspace{10pt} % <-- Add horizontal space between subfigures
    \begin{subfigure}[t]{0.3\linewidth}
        \centering
        \fbox{\includegraphics[width=0.3\linewidth]{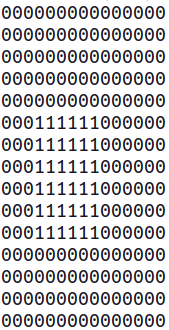}}
        %\caption{The square.}
        \label{fig:square}
    \end{subfigure}
    \hspace{10pt}
    \begin{subfigure}[t]{0.3\linewidth}
        \centering
        \fbox{\includegraphics[width=0.3\linewidth]{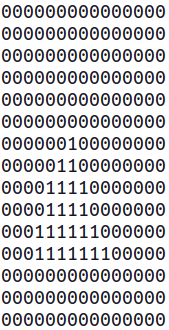}}
        %\caption{The triangle.}
        \label{fig:triangle_shapes}
    \end{subfigure}%
    \hspace{10pt} % <-- Add horizontal space between subfigures
    \begin{subfigure}[t]{0.3\linewidth}
        \centering
        \fbox{\includegraphics[width=0.3\linewidth]{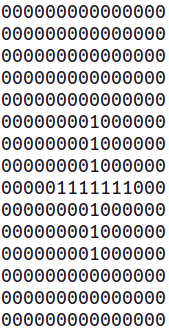}}
        %\caption{The cross.}
        \label{fig:cross}
    \end{subfigure}
    \caption{Matrices of the Shape Recognition Game. From Left to Right: Square, Triangle, Cross.}
    \label{fig:shapes}
\end{figure}

\textbf{Shape Recognition Game} \quad We first probed what geometric shapes the LLMs considered basic. The shapes consistently mentioned were square, circle, and triangle. Accordingly, the Shape Recognition Game consists of identifying a basic geometric shape "hidden" behind 1s in a matrix of 0s in a multiple-choice fashion with four options: circle, rectangle, triangle, and cross. Only the latter three were ever shown (cf. Figure~\ref{fig:shapes}).

\textbf{Lego Connect Language}\label{sec:lcl_rules} \quad We invented the LEGO Connect Language (LCL). We propose $LCL_2$ as a language to instruct assembly in 2D. In the current implementation, only 2x4 pieces are allowed for a construction, $M$, which is valid \textit{if} no pieces are overlapping and all pieces are connected to other pieces. A Lego piece can only be connected through interlocking pegs. Two Lego pieces overlap when they share the same y-coordinate and any part of their length has the same x-coordinate (cf. Supporting Information, Section~\ref{app:def_lcl} for a formal definition).

\begin{figure}[H]
    \centering
    \begin{subfigure}[t]{0.45\linewidth}
        \centering
        \includegraphics[width=0.45\linewidth]{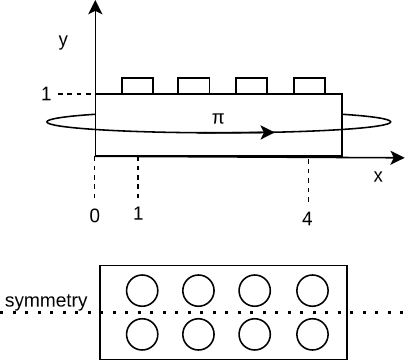}
        %\caption{A standard LEGO piece in $LCL_2$.}
        \label{fig:rotation}
    \end{subfigure}
    \hspace{10pt} % <-- Add horizontal space between subfigures
    \begin{subfigure}[t]{0.45\linewidth}
        \centering
        \includegraphics[width=0.6\linewidth]{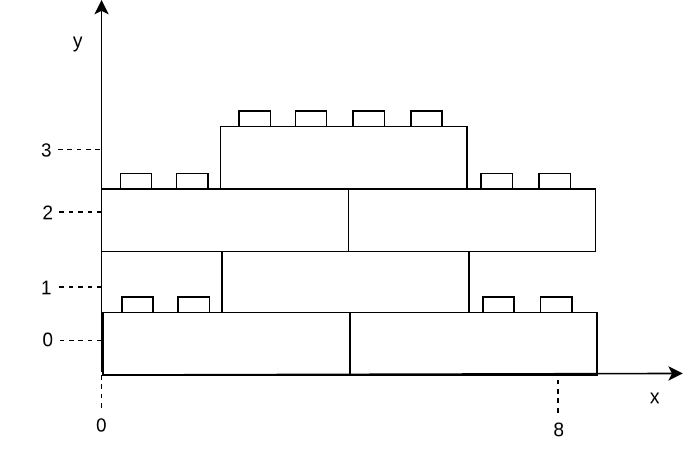}
        %\caption{A \textit{valid} humanoid construct in $LCL_2$.}
        \label{fig:subfig_humanoid}
    \end{subfigure}%
    \caption{Left: A standard LEGO piece in $LCL_2$. Right: A \textit{valid} construct in $LCL_2$.}
    \label{fig:valid_lcl}
\end{figure}

LCL includes two games, validity testing, which evaluates the ability to assess constructs as valid or invalid, and construct generation, where models are tasked to generate valid LCL constructs from prompts. Constructs are described by tuples specifying the coordinates and colors of pieces. In both games, we evaluate the proportion of correct answers. We generated 800 constructs for the validity test (400 valid, 400 invalid) and, for the construct generation, queried each model to produce 100 constructs. A construct example encoded in ASCII is $[(1, 1, 'red'), (2, 1, 'blue'), (3, 1, 'green')]$, where each tuple represents a brick's position and color. 

\begin{figure}[H]
\centerline{\includegraphics[width=0.60\columnwidth]{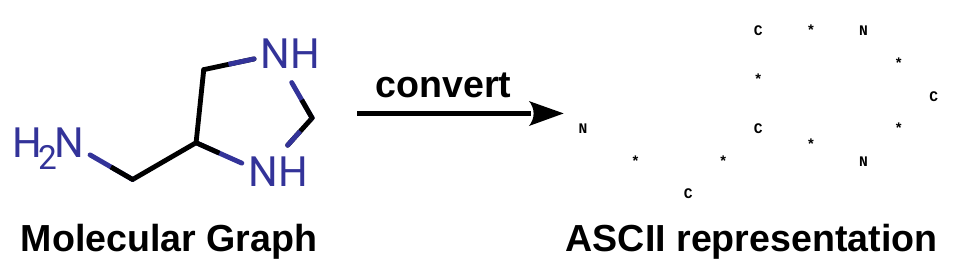}}
  \caption{Depicting a molecular graph in an ASCII representation.}
  \label{fig:smiles_ascii}
\end{figure}

\textbf{Guess-the-SMILES} \quad Finally, we have a hidden task with undisclosed source code about interpreting ASCII drawings of molecules (cf. Figure \ref{fig:smiles_ascii} and Supporting Information, Section~\ref{app:GtS}). The goal is to provide a corresponding SMILES, an established string representation of molecular graphs \cite{weininger1988smiles}. This task evaluates both spatial and chemical logic of the LLM. It allows to assess whether LLMs understand the relative spatial arrangement of atoms as encoded in a SMILES string, which is critical for structure-property-relationships.

\begin{figure}[H]
\centerline{\includegraphics[width=0.60\columnwidth]{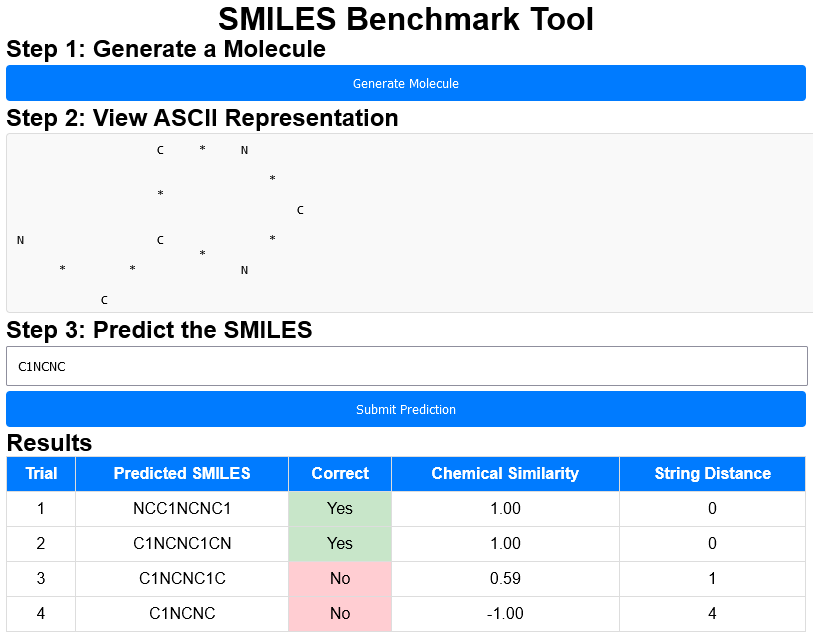}}
  \caption{Guess-the-SMILES game UI accessible through a \href{https://child-play.onrender.com/}{dedicated website}.}
  \label{fig:smiles_game}
\end{figure}

When prompted, all tested models are able to explain SMILES of molecules and manipulate them. GtS is implemented as Flask web application. We provide an API allowing to request ASCII drawings, submit predicted SMILES, and collect results. We provide a user interface (UI) to test the game (Figure \ref{fig:smiles_game}), available via this \href{https://child-play.onrender.com/}{website}. 

%%%  The methods (experimental procedures) should appear 
%%%  immediately after the discussion. Subheadings 
%%%  may be customized. Please consult your handling 
%%%  editor if you have questions about what content 
%%%  should appear here. (Note: Our production team 
%%%  will ensure that this your manuscript has the 
%%%  journal's heading, "methods" or "experimental 
%%%  procedures," before the proof stage.

% \newpage

\section*{Results}
Table \ref{tab:win_rates} summarizes the win and lose rates of GPT-3.5, GPT-4, GPT-4o, and GPT-4o-mini in the three board games Battleship, Tic-Tac-Toe, and Connect-Four against a random player, evaluated at various temperature parameter values.

\begin{table}[H]
\centering
\resizebox{\textwidth}{!}{
\begin{tabular}{lcccccccc}
\toprule
\textbf{Model} & \textbf{Temp.} & \multicolumn{2}{c}{\textbf{Battleship}} & \multicolumn{2}{c}{\textbf{Tic-Tac-Toe}} & \multicolumn{2}{c}{\textbf{Connect-Four}} \\ 
\cmidrule(lr){3-4} \cmidrule(lr){5-6} \cmidrule(lr){7-8}
 & & \textbf{Win Rate (\%)} & \textbf{Lose Rate (\%)} & \textbf{Win Rate (\%)} & \textbf{Lose Rate (\%)} & \textbf{Win Rate (\%)} & \textbf{Lose Rate (\%)} \\ 
\midrule
\textbf{GPT-3.5}    & 0.0 & \textbf{10.00} & 90.00 & \textbf{53.00} & 47.00 & 64.00 & 36.00 \\
\textbf{GPT-3.5}    & 0.5 & 8.00  & 92.00 & 24.00 & 76.00 & \textbf{76.00} & 24.00 \\
\textbf{GPT-3.5}    & 1.0 & 3.00  & 97.00  & 24.00 & 76.00 & 75.00 & 25.00 \\
\textbf{GPT-3.5}    & 1.5 & 3.00  & 97.00 & 19.00 & 81.00 & 75.00 & 25.00 \\ 
\cmidrule(lr){1-8}
\textbf{GPT-4}      & 0.0 & 0.00  & 100.00 & 71.00 & 25.00 & \textbf{80.00} & 20.00 \\
\textbf{GPT-4}      & 0.5 & 0.00  & 100.00 & \textbf{77.00} & 18.00 & \textbf{80.00} & 20.00 \\
\textbf{GPT-4}      & 1.0 & \textbf{4.00}  & 96.00 & 69.00 & 28.00 & 70.00 & 30.00 \\
\textbf{GPT-4}      & 1.5 & 0.00  & 100.00 & 50.00 & 46.00 & 39.00 & 61.00 \\ 
\cmidrule(lr){1-8}
\textbf{GPT-4o}     & 0.0 & \textbf{0.00}  & 100.00 & 50.00 & 39.00 & 79.00 & 21.00 \\
\textbf{GPT-4o}     & 0.5 & \textbf{0.00}  & 100.00 & \textbf{92.00} & 6.00  & \textbf{80.00} & 20.00 \\
\textbf{GPT-4o}     & 1.0 & \textbf{0.00}  & 100.00 & 85.00 & 11.00 & 73.00 & 27.00 \\
\textbf{GPT-4o}     & 1.5 & \textbf{0.00}  & 100.00 & 88.00 & 9.00  & 67.00 & 33.00 \\ 
\cmidrule(lr){1-8}
\textbf{GPT-4o-mini}& 0.0 & \textbf{0.00}  & 100.00 & 10.00 & 89.00 & 75.00 & 25.00 \\
\textbf{GPT-4o-mini}& 0.5 & \textbf{0.00}  & 100.00 & 59.00 & 25.00 & \textbf{79.00} & 21.00 \\
\textbf{GPT-4o-mini}& 1.0 & \textbf{0.00}  & 100.00 & \textbf{61.00} & 29.00 & 73.00 & 27.00 \\
\textbf{GPT-4o-mini}& 1.5 & \textbf{0.00}  & 100.00 & 59.00 & 24.00 & 69.00 & 31.00 \\ 
\bottomrule
\end{tabular}
}
\caption{LLM win and lose rates in Battleship, Tic-Tac-Toe, and Connect-Four at different temperatures.}
\label{tab:win_rates}
\end{table}

\begin{figure}[H]
    \centering
    \includegraphics[width=0.475\linewidth]{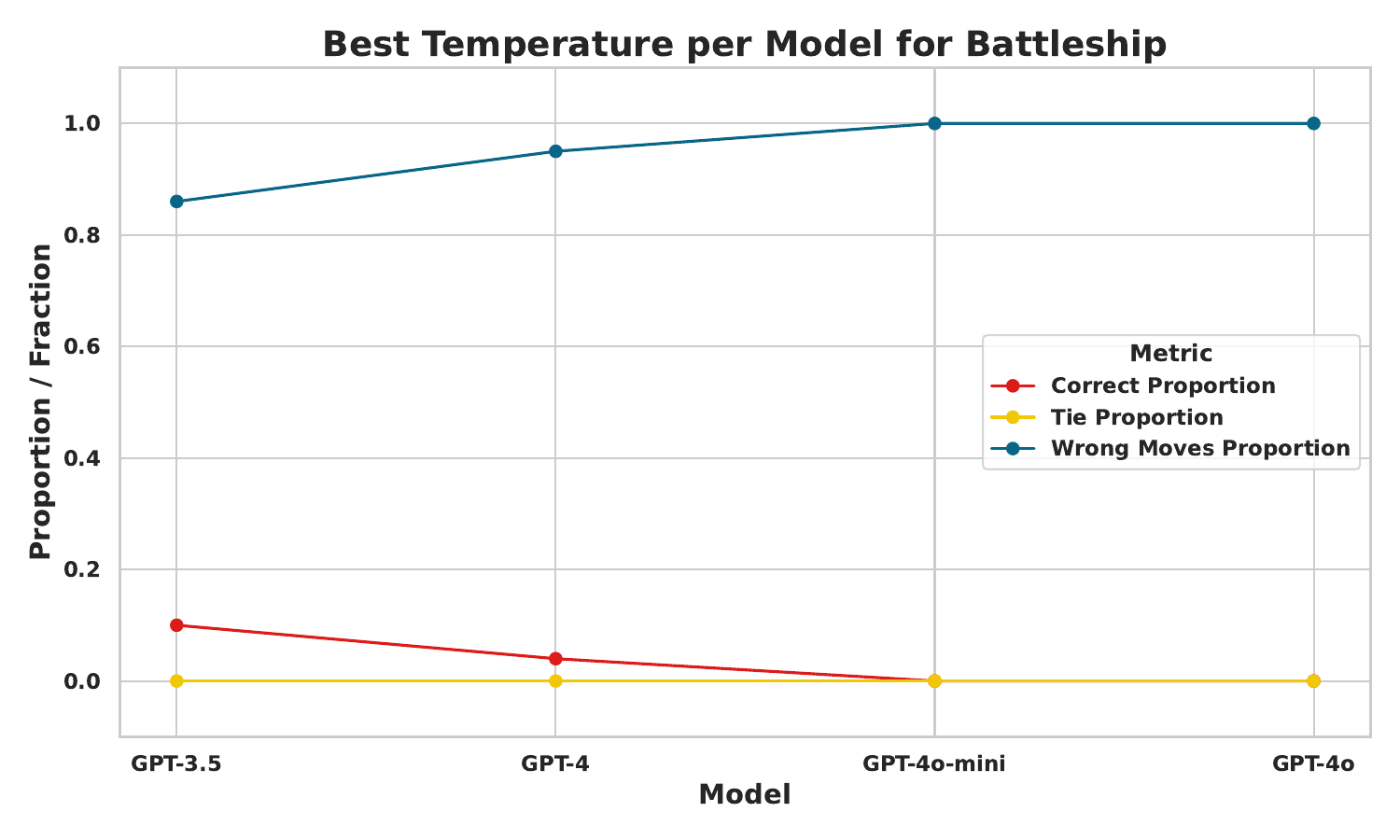} %
    \includegraphics[width=0.475\linewidth]{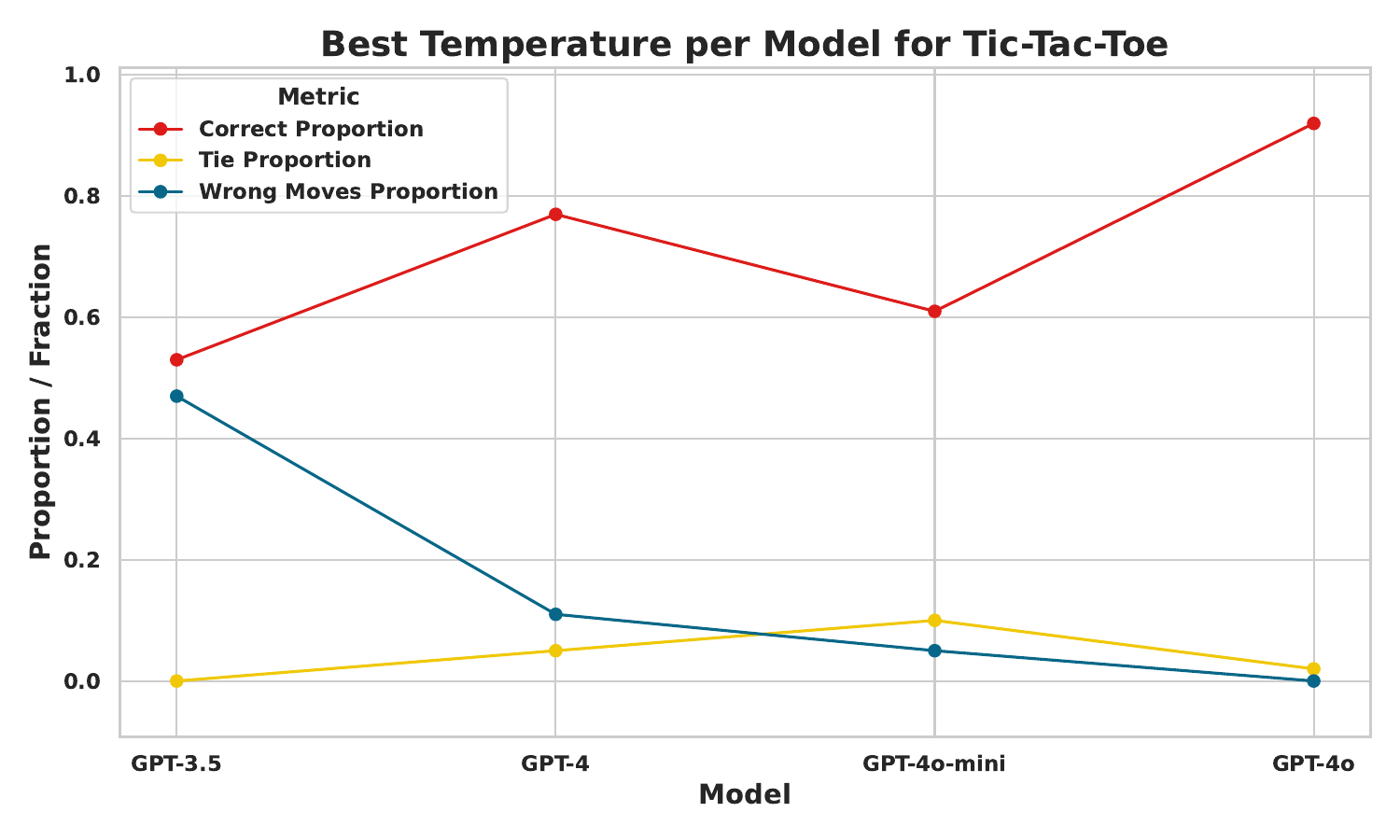}
    \caption{Comparison of best model performances in Battleship (left) and Tic-Tac-Toe (right).}
    \label{fig:battleship_Tic-Tac-best}
\end{figure}

\textbf{Battleship} \quad The LLMs did not follow the rules and lost most games. GPT-4 only won very few games, both GPT-4o and GPT-4o-mini did not win any game due to wrong moves (cf. Figure~\ref{fig:battleship_Tic-Tac-best}). GPT-3.5 performed the best and won 10\% of its games at a low temperature, but none at higher temperature parameter values. Surprisingly, LLM performance degrades on this benchmark task with formally improved models relative to GPT-3.5.

\textbf{Tic-Tac-Toe} \quad Here, GPT-4o stands out with a high win rate of 92\% at a temperature of 0.5, whereas GPT-3.5 and GPT-4 only achieve win rates of 53\% and 77\%, respectively, at their optimal temperatures (Figure~\ref{fig:battleship_Tic-Tac-best}). GPT-4o-mini performs slightly better than GPT-3.5, reaching a 61\% win rate at a temperature of 1.0. Nevertheless, none of the models reach the Minimax baseline win rate of 99.6\%, which corresponds to optimal play (cf. Supporting Information, Section~\ref{app:minimax}). All models exhibit deterministic play at the lowest temperature \( t = 0 \) (cf.\ Supporting Information, Section~\ref{app:heatmaps}). The random player rarely counters the central opening moves of the models (cf. Figures~\ref{fig:tictactoe_moves_3_5}-\ref{fig:tictactoe_moves_4o}). GPT-3.5 makes more errors compared to the other models, leading to increased losses due to illegal moves with increasing temperature (cf. Figure~\ref{fig:Tic-Tac-Toe_missed_wins_blocks}). GPT-4o and GPT-4o-mini demonstrate significantly fewer incorrect moves at all temperatures, even as randomness increases (cf. Figure~\ref{fig:tictactoe_moves_3_5}).

\begin{figure}[H]
    \centering
    \includegraphics[width=0.475\linewidth]{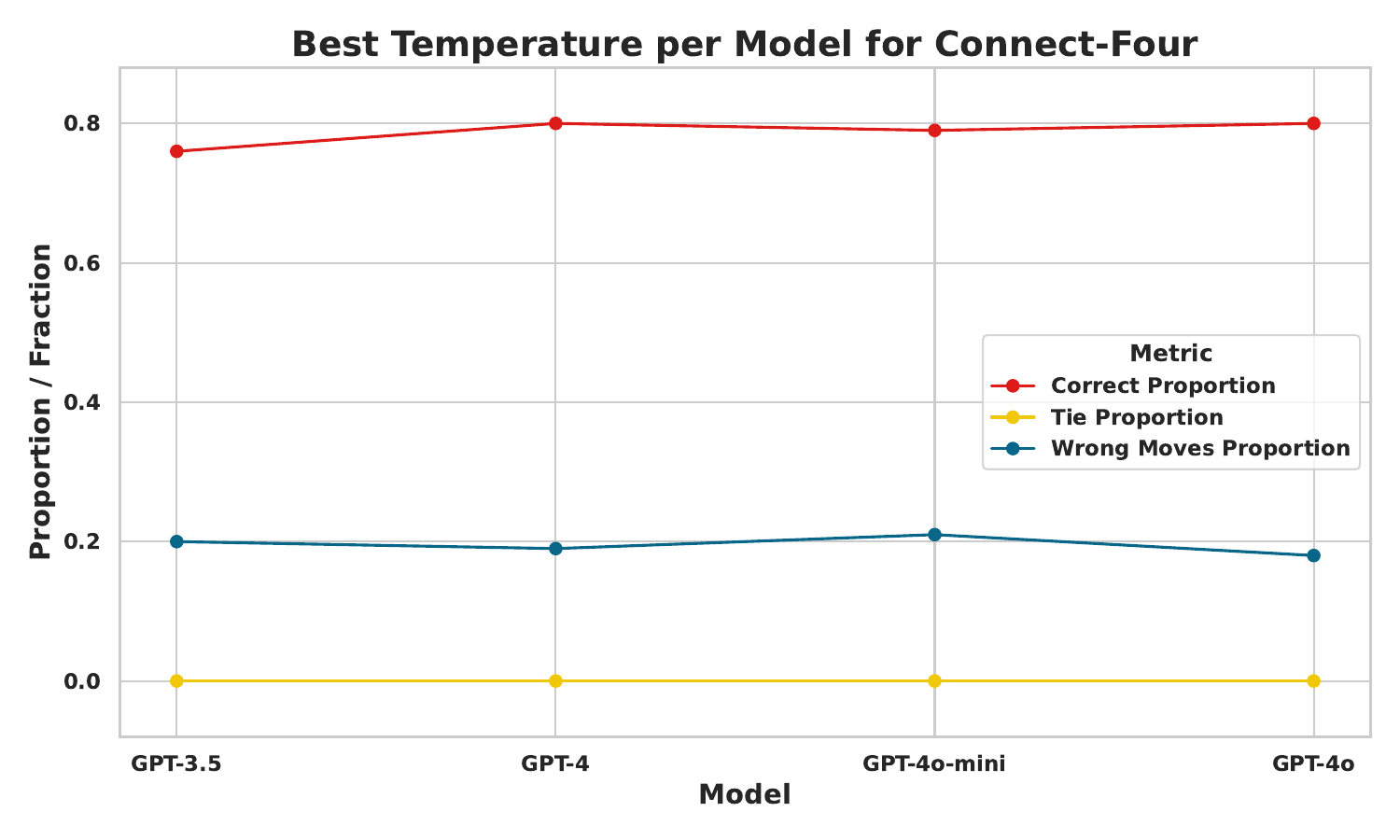} %
    \includegraphics[width=0.475\linewidth]{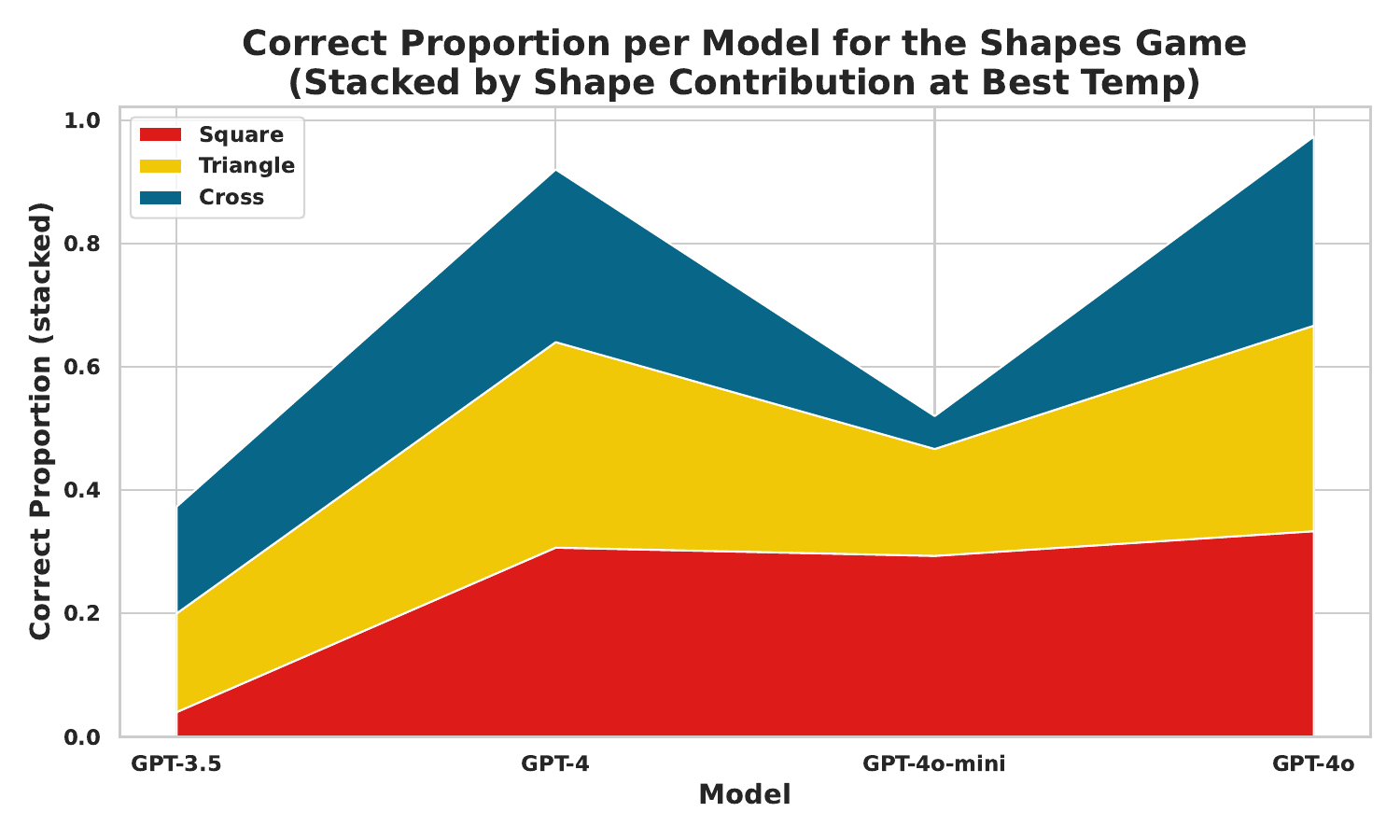}
    \caption{Comparison of best model performances in Connect-Four (left) and the Shape Recognition Game (right).}
    \label{fig:connect-four_shapes-best}
\end{figure}

\textbf{Connect-Four} \quad GPT-4 and GPT-4o exhibit the best performance, both reaching a win rate of 80\% at lower temperatures (Figure~\ref{fig:connect-four_shapes-best}). Nevertheless, both GPT-3.5 and GPT-4o-mini also perform competitively with win rates of 76\% and 79\%, respectively. All models experience a decline in performance as the temperature increases. Hence, all the tested LLMs show a comparable performance level in this benchmark task.

All models display deterministic play at the lowest temperature (cf. Supporting Information, Section~\ref{app:heatmap_connectfour}). The LLMs prefer the middle column, even when it is full. This increases the chance of winning against a random player, who only has a $1/7$ chance of blocking (Figure~\ref{fig:connectfour_moves_3_5}). While GPT-3.5 and GPT-4 show an increase in missed wins and blocks with higher temperatures, both GPT-4o and GPT-4o-mini keep these errors somewhat constant (cf. Figure~\ref{fig:connect_four_results}).

Across the three board games considered, no model consistently outperforms the others. While Tic-Tac-Toe performance increases with the expected model progression (GPT-3.5 $<$ GPT-4 $<$ GPT-4o), Connect-Four performance remains roughly constant and Battleship performance actually shows the opposite trend (GPT-3.5 $>$ GPT-4 $>$ GPT-4o). For the latter, GPT-3.5 remains the only model that is able to play a few rounds successfully.

% Performances vary across games and temperatures (cf. Table \ref{tab:win_rates}). In Battleship, GPT-3.5's wins drop from 10\% at \( t = 0.0 \) to 3\% at \( t = 1.0 \) and \( t = 1.5 \). GPT-4 performs poorly, with no wins at \( t = 0.0 \), \( t = 0.5 \), and \( t = 1.5 \), and only 4\% at \( t = 1.0 \). GPT-4o and GPT-4o-mini perform similarly to GPT-4 in Battleship, with zero wins across all temperatures.

% In Tic-Tac-Toe, GPT-3.5's win rate decreases from 53\% winrate at \( t = 0.0 \) to 19\% at \( t = 1.5 \). GPT-4 performs better, peaking at 77\% at \( t = 0.5 \) but dropping to 50\% at \( t = 1.5 \). GPT-4o surpasses both, achieving 92\% at \( t = 0.5 \) and 85\% at \( t = 1.0 \), though it drops to 11\% at \( t = 1.5 \). GPT-4o-mini exhibits a similar trend, reaching 59\% at both \( t = 0.5 \) and \( t = 1.5 \), with a low of 10\% at \( t = 0.0 \).

% In Connect-Four, GPT-3.5 improves from 64\% winrate at \( t = 0.0 \) to 76\% at \( t = 0.5 \) and maintains around 75\% at higher temperatures. GPT-4 starts with 80\% at \( t = 0.0 \) but falls significantly to 39\% at \( t = 1.5 \). GPT-4o and GPT-4o-mini maintain a higher winrate in Connect-Four, with GPT-4o achieving 81\% at \( t = 0.0 \) and remaining above 67\% at all temperatures, while GPT-4o-mini achieves 79\% at \( t = 0.5 \) and maintains a minimum of 69\% at \( t = 1.5 \).

\begin{table}[H]
\centering
\resizebox{\textwidth}{!}{
\begin{tabular}{lcccccccccc}
\toprule
\textbf{Model} & \textbf{Temp.} & \multicolumn{2}{c}{\textbf{Square}} & \multicolumn{2}{c}{\textbf{Triangle}} & \multicolumn{2}{c}{\textbf{Cross}} & \multicolumn{2}{c}{\textbf{Overall}} \\
\cmidrule(lr){3-4} \cmidrule(lr){5-6} \cmidrule(lr){7-8} \cmidrule(lr){9-10}
 & & \textbf{Correct (\%)} & \textbf{Incorrect (\%)} & \textbf{Correct (\%)} & \textbf{Incorrect (\%)} & \textbf{Correct (\%)} & \textbf{Incorrect (\%)} & \textbf{Correct (\%)} & \textbf{Incorrect (\%)} \\
\midrule
\textbf{GPT-3.5}   & 0.0 & 4.00  & 96.00  & 12.00 & 88.00  & \textbf{72.00} & 28.00  & 29.33 & 70.67 \\
\textbf{GPT-3.5}   & 0.5 & 0.00  & 100.00 & 24.00 & 76.00  & 56.00 & 44.00  & 26.67 & 73.33 \\
\textbf{GPT-3.5}   & 1.0 & 0.00  & 100.00 & 24.00 & 76.00  & 52.00 & 48.00  & 25.33 & 74.67 \\
\textbf{GPT-3.5}   & 1.5 & \textbf{12.00} & 88.00  & \textbf{48.00} & 52.00  & 52.00 & 48.00  & \textbf{37.33} & 62.67 \\
\midrule
\textbf{GPT-4}     & 0.0 & \textbf{92.00} & 8.00   & \textbf{100.00}& 0.00   & 84.00 & 16.00  & \textbf{92.00} & 8.00  \\
\textbf{GPT-4}     & 0.5 & 84.00 & 16.00  & \textbf{100.00}& 0.00   & 80.00 & 20.00  & 88.00 & 12.00 \\
\textbf{GPT-4}     & 1.0 & 88.00 & 12.00  & 92.00 & 8.00   & \textbf{92.00} & 8.00   & 90.67 & 9.33  \\
\textbf{GPT-4}     & 1.5 & 80.00 & 20.00  & 96.00 & 4.00   & 76.00 & 24.00  & 84.00 & 16.00 \\
\midrule
\textbf{GPT-4o}    & 0.0 & \textbf{100.00}& 0.00   & \textbf{100.00}& 0.00   & \textbf{92.00} & 8.00   & \textbf{97.33} & 2.67  \\
\textbf{GPT-4o}    & 0.5 & \textbf{100.00}& 0.00   & \textbf{100.00}& 0.00   & 76.00 & 24.00  & 92.00 & 8.00  \\
\textbf{GPT-4o}    & 1.0 & \textbf{100.00}& 0.00   & 96.00 & 4.00   & 84.00 & 16.00  & 93.33 & 6.67  \\
\textbf{GPT-4o}    & 1.5 & \textbf{100.00}& 0.00   & 96.00 & 4.00   & 72.00 & 28.00  & 89.33 & 10.67 \\
\midrule
\textbf{GPT-4o-mini}& 0.0 & 88.00 & 12.00  & \textbf{52.00} & 48.00  & \textbf{16.00} & 84.00  & \textbf{52.00} & 48.00 \\
\textbf{GPT-4o-mini}& 0.5 & \textbf{100.00}& 0.00   & 40.00 & 60.00  & 12.00 & 88.00  & 50.67 & 49.33 \\
\textbf{GPT-4o-mini}& 1.0 & 96.00 & 4.00   & 40.00 & 60.00  & \textbf{16.00} & 84.00  & 50.67 & 49.33 \\
\textbf{GPT-4o-mini}& 1.5 & \textbf{100.00}& 0.00   & 44.00 & 56.00  & 12.00 & 88.00  & \textbf{52.00} & 48.00 \\
\bottomrule
\end{tabular}
}
\caption{Correct and incorrect rates (\%) of the LLMs for each shape individually and overall across temperatures.}
\label{tab:shapes_comparison_overall}
\end{table}

\textbf{Shape Recognition Game} \quad The results for both the individual shapes and overall are provided in Table~\ref{tab:shapes_comparison_overall}. GPT-3.5 shows very poor performance on squares, which is worse than random choice. However, it seems to recognize both triangles and crosses more consistently. In contrast, both GPT-4 and GPT-4o identified most shapes, with GPT-4 excelling at recognizing triangles and GPT-4o at both squares and triangles. GPT-4o-mini performs significantly better than GPT-3.5 on both squares and triangles, but worse than random on crosses. The overall performance comparison is illustrated in Figure~\ref{fig:connect-four_shapes-best}.

\begin{figure}[H]
    \centering
    \includegraphics[width=0.475\linewidth]{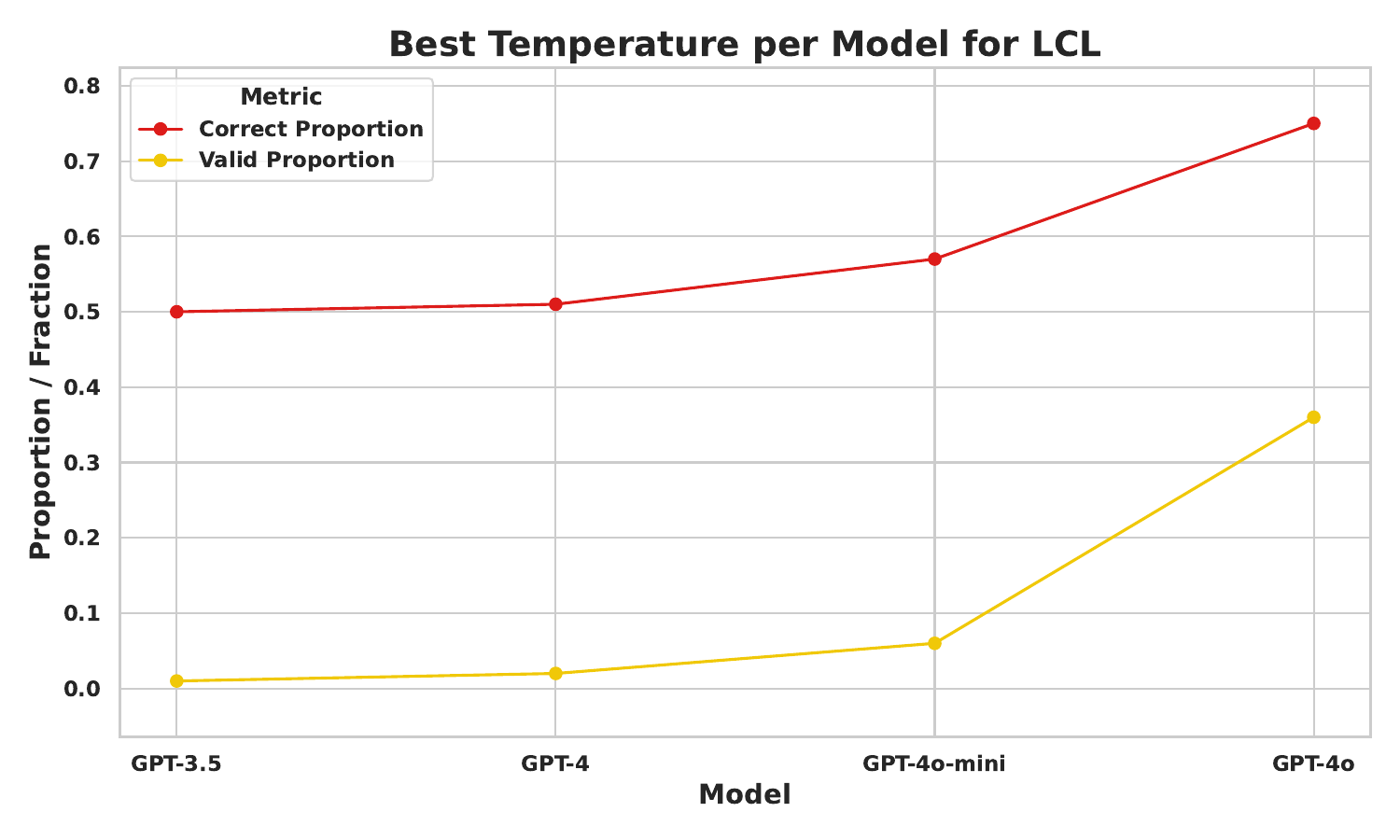} %
    \includegraphics[width=0.475\linewidth]{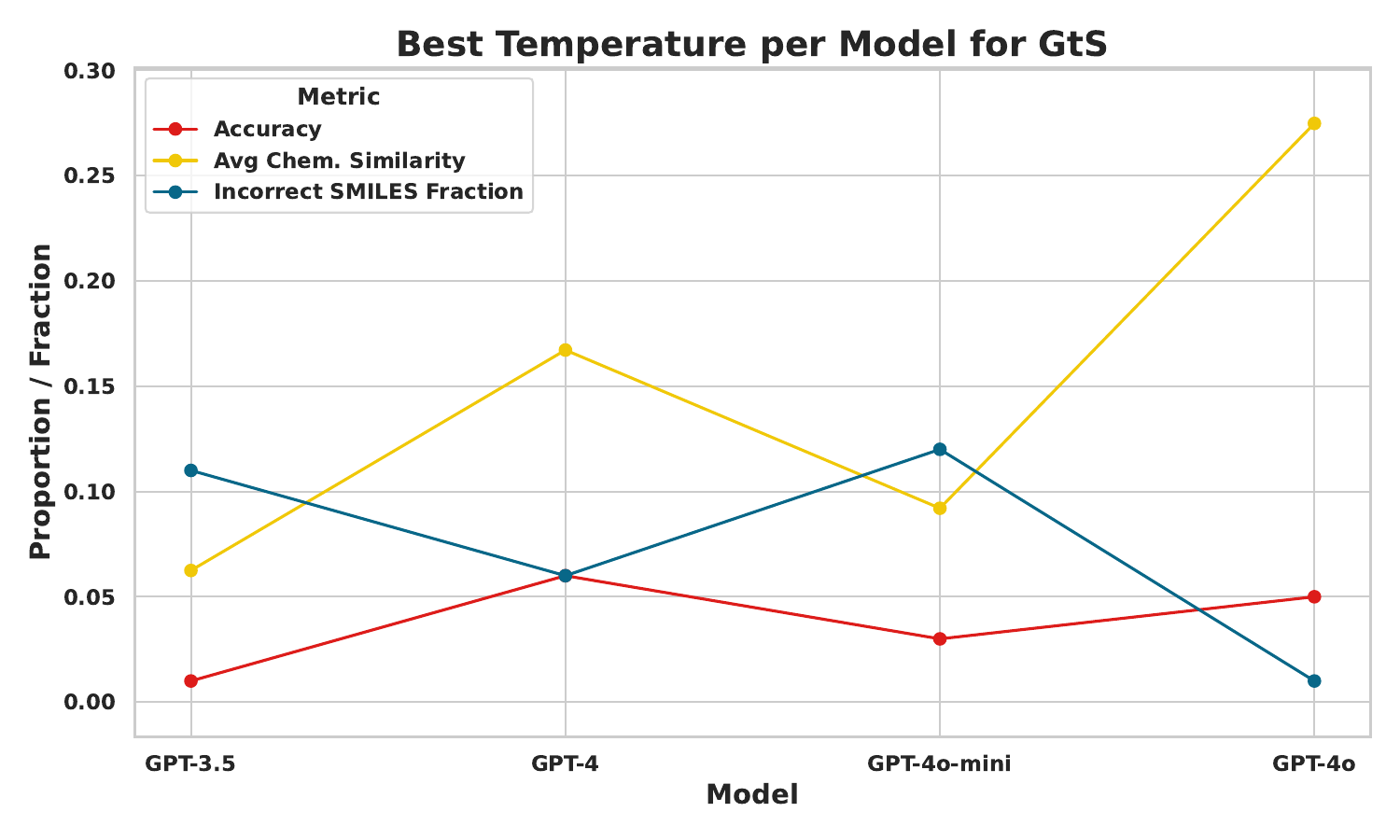}
    \caption{Comparison of best model performances in LCL (left) and GtS (right).}
    \label{fig:lcl-gts-best}
\end{figure}

\textbf{Lego Connect Language} \quad 
Table \ref{tab:model_correct_valid_proportions} summarizes the key performance metrics for the two LCL benchmark tasks, validity testing and construct generation, across temperatures. The corresponding best results are illustrated in Figure~\ref{fig:lcl-gts-best}. In validity testing, both GPT-3.5 and GPT-4 only achieved correct response rates that correspond to random answering (50\%). In contrast, both GPT-4o and GPT-4o-mini recorded somewhat higher validity rates, with GPT-4o peaking at 75\% and GPT-4o-mini reaching 57\%. In the construct generation task, all models struggled to produce valid suggestions. GPT-3.5, GPT-4, and GPT-4o-mini generated hardly any valid constructs. Surprisingly, GPT-4o-mini even performs slightly better with 6\% correct constructs. GPT-4o demonstrated a significant improvement reaching 36\% at its best temperature setting.

\begin{table}[H]
\centering
\tiny
\resizebox{0.75\textwidth}{!}{
\begin{tabular}{lccccc}
\toprule
\textbf{Model} & \textbf{Temp.} & \multicolumn{2}{c}{\textbf{Validity Responses}} & \multicolumn{2}{c}{\textbf{Construct Generation}} \\ 
\cmidrule(lr){3-4} \cmidrule(lr){5-6}
 & & \textbf{Proportion (\%)} & \textbf{SE (\%)} & \textbf{Proportion (\%)} & \textbf{SE (\%)} \\ 
\midrule
\textbf{GPT-3.5}      & 0.0 & \textbf{50.00} & 5.03 & 0.00  & 0.00 \\
\textbf{GPT-3.5}      & 0.5 & 48.00 & 5.02 & 0.00  & 0.00 \\
\textbf{GPT-3.5}      & 1.0 & 41.00 & 4.94 & \textbf{1.00}  & 1.00 \\
\textbf{GPT-3.5}      & 1.5 & 48.00 & 5.02 & \textbf{1.00}  & 1.00 \\ 
\cmidrule(lr){1-6}
\textbf{GPT-4}        & 0.0 & 50.00 & 5.03 & 0.00  & 0.00 \\
\textbf{GPT-4}        & 0.5 & 50.00 & 5.03 & 0.00  & 0.00 \\
\textbf{GPT-4}        & 1.0 & 50.00 & 5.03 & \textbf{2.00}  & 1.41 \\
\textbf{GPT-4}        & 1.5 & \textbf{51.00} & 5.02 & \textbf{2.00}  & 1.41 \\ 
\cmidrule(lr){1-6}
\textbf{GPT-4o}       & 0.0 & 67.00 & 4.73 & 0.00  & 0.00 \\
\textbf{GPT-4o}       & 0.5 & 64.00 & 4.82 & 18.00 & 3.86 \\
\textbf{GPT-4o}       & 1.0 & \textbf{75.00} & 4.35 & \textbf{36.00} & 4.82 \\
\textbf{GPT-4o}       & 1.5 & 56.00 & 4.99 & 25.00 & 4.35 \\ 
\cmidrule(lr){1-6}
\textbf{GPT-4o-mini}  & 0.0 & 50.00 & 5.03 & 0.00  & 0.00 \\
\textbf{GPT-4o-mini}  & 0.5 & 54.00 & 5.01 & 0.00  & 0.00 \\
\textbf{GPT-4o-mini}  & 1.0 & \textbf{57.00} & 4.98 & 2.00  & 1.41 \\
\textbf{GPT-4o-mini}  & 1.5 & \textbf{57.00} & 4.98 & \textbf{6.00}  & 2.39 \\ 
\bottomrule
\end{tabular}
}
\caption{Proportions of correct responses and the corresponding standard errors (SE) of the LLMs on the LCL benchmarks across temperatures ($\text{SE (\%)} = \frac{\sigma}{\sqrt{n}} \times 100\%$, with the sample standard deviation $\sigma$ and sample size $n$).}
\label{tab:model_correct_valid_proportions}
\end{table}

\textbf{Guess-the-SMILES} \quad Table \ref{tab:smiles_results} summarizes the LLM results based on 100 runs per model at each temperature setting. Notably, invalid SMILES are assigned a similarity score of -1. GPT-3.5 generated hardly any correct SMILES across all temperatures, with average chemical similarity scores around zero and many invalid SMILES. GPT-4 demonstrates marginally better performance than GPT-3.5 by generating more correct SMILES and achieving higher chemical similarity scores. GPT-4o offers a further improvement with respect to the chemical similarity scores and a reduced number of invalid SMILES. Despite these enhancements, the overall accuracy remains comparable to GPT-4. GPT-4o-mini performs slightly better than GPT-3.5.

\begin{table}[H]
\centering
\resizebox{0.75\textwidth}{!}{
\begin{tabular}{lcccccc}
\toprule
\textbf{Model} & \textbf{Temp.} & \textbf{Correct} & \textbf{Incorrect} & \textbf{Invalid} & \textbf{Similarity} & \textbf{Accuracy (\%)} \\ 
\midrule
\textbf{GPT-3.5}    & 0.0 & 1 & 88 & 11 & 0.062 & \textbf{1.1} \\
\textbf{GPT-3.5}    & 0.5 & 0 & 89 & 11 & 0.053 & 0.0 \\
\textbf{GPT-3.5}    & 1.0 & 1 & 78 & 21 & $-$0.070 & \textbf{1.3} \\
\textbf{GPT-3.5}    & 1.5 & 0 & 70 & 30 & $-$0.197 & 0.0 \\ 
\cmidrule(lr){1-7}
\textbf{GPT-4}      & 0.0 & 3 & 90 & 7  & 0.158 & 3.2 \\
\textbf{GPT-4}      & 0.5 & 6 & 88 & 6  & 0.167 & \textbf{6.4} \\
\textbf{GPT-4}      & 1.0 & 1 & 95 & 4  & 0.167 & 1.0 \\
\textbf{GPT-4}      & 1.5 & 2 & 88 & 10 & 0.089 & 2.2 \\ 
\cmidrule(lr){1-7}
\textbf{GPT-4o}     & 0.0 & 5 & 94 & 1  & 0.275 & 5.1 \\
\textbf{GPT-4o}     & 0.5 & 4 & 92 & 4  & 0.221 & 4.2 \\
\textbf{GPT-4o}     & 1.0 & 5 & 91 & 4  & 0.223 & 5.2 \\
\textbf{GPT-4o}     & 1.5 & 5 & 83 & 12 & 0.122 & \textbf{5.7} \\ 
\cmidrule(lr){1-7}
\textbf{GPT-4o-mini}& 0.0 & 3 & 85 & 12 & 0.092 & \textbf{3.4} \\
\textbf{GPT-4o-mini}& 0.5 & 1 & 88 & 11 & 0.051 & 1.1 \\
\textbf{GPT-4o-mini}& 1.0 & 2 & 81 & 17 & 0.026 & 2.4 \\
\textbf{GPT-4o-mini}& 1.5 & 1 & 78 & 21 & $-$0.042 & 1.3 \\ 
\bottomrule
\end{tabular}
}
\caption{LLM results on the GtS benchmark. `Invalid' counts invalid SMILES. `Similarity' is the average chemical similarity score. `Accuracy' is computed as $\frac{\text{Correct}}{\text{Correct}+\text{Incorrect}} \cdot 100\%$.}
\label{tab:smiles_results}
\end{table}

All models demonstrate significant limitations in reproducing the depicted molecule or
generating valid SMILES (Figure~\ref{fig:lcl-gts-best}). While GPT-4 and GPT-4o perform slightly better than both GPT-3.5 and GPT-4o-mini, their capabilities remain low.

\begin{figure}[H]
    \centering
    \includegraphics[width=0.6\linewidth]{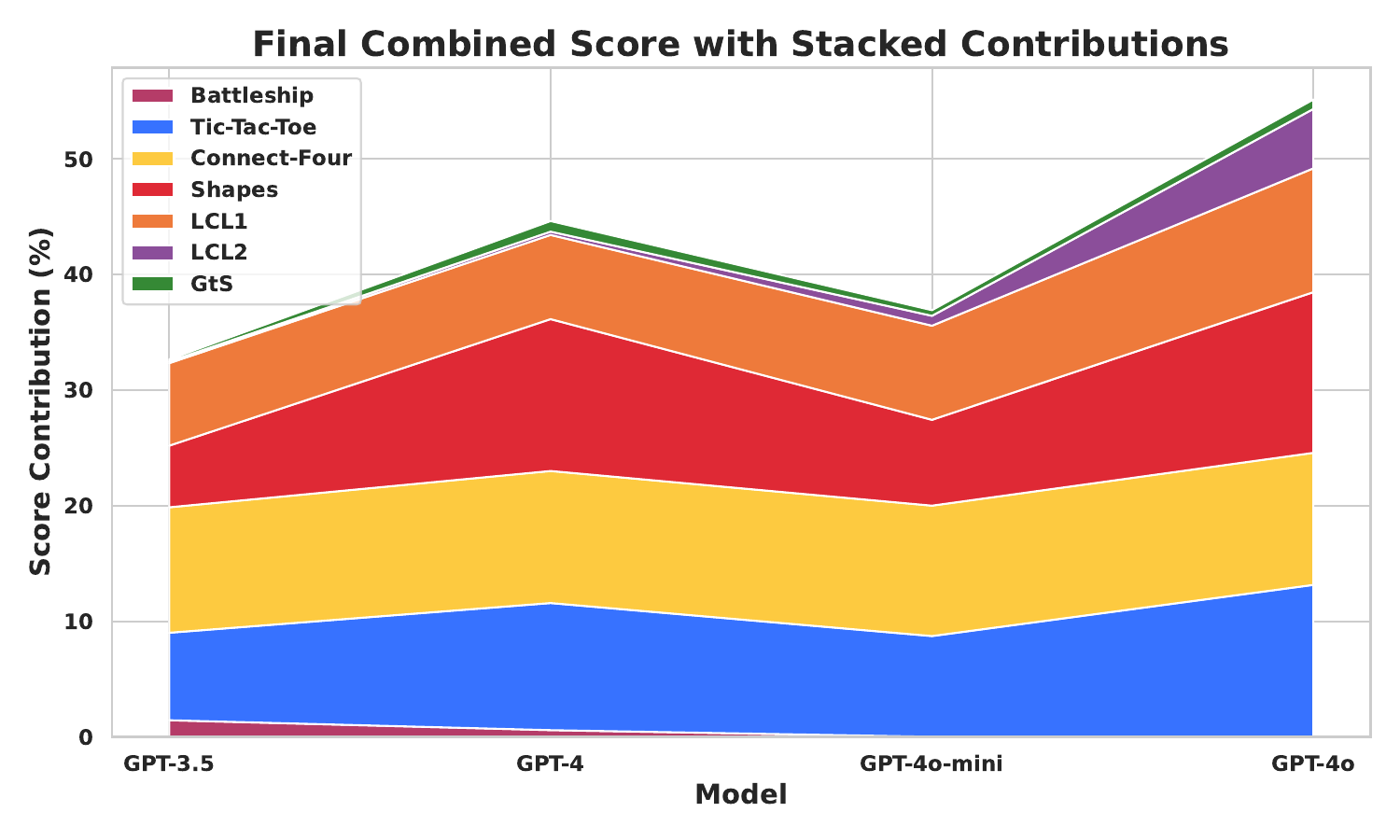}
    \caption{Comparison of average performance across all \texttt{ChildPlay} benchmark tasks. LCL1 corresponds to validity testing, LCL2 to construct generation.}
    \label{fig:avg_game_performance}
\end{figure}

\textbf{Overall Performance} \quad Finally, we derive an overall performance metric in the \texttt{ChildPlay} benchmarks to summarize the capabilities of the tested LLMs. The Combined Score in Figure~\ref{fig:avg_game_performance} is computed as the average of one performance metric per game, namely the win probabilities in Battleship, Tic-Tac-Toe, and Connect-Four, the proportion of correct responses in both the Shape Recognition Game and the two LCL tasks individually, and the accuracy from GtS. The corresponding numerical data is provided in Table~\ref{tab:overall_performance}. Overall, we observe that subsequent GPT model versions, on average, lead to better performance.

\begin{table}[H]
\centering
\tiny
\resizebox{1.0\textwidth}{!}{
\begin{tabular}{lcccccccc}
\toprule
\textbf{Model} & \textbf{Battleship} & \textbf{Tic-Tac-Toe} & \textbf{Connect-Four} & \textbf{Shapes} & \textbf{LCL1} & \textbf{LCL2} & \textbf{GtS} & \textbf{Overall} \\ 
\midrule
\textbf{GPT-3.5}      & \textbf{10.00} & 53.00 & 76.00 & 37.33 & 50.00 & 1.00 & 1.30 & 32.66 \\
\textbf{GPT-4}        & 4.00 & 77.00 & \textbf{80.00} & 92.00 & 51.00 & 2.00 & \textbf{6.40} & 44.63 \\
\textbf{GPT-4o-mini}  & 0.00 & 61.00 & 79.00 & 52.00 & 57.00 & 6.00 & 3.40 & 36.91 \\ 
\textbf{GPT-4o}       & 0.00 & \textbf{92.00} & \textbf{80.00} & \textbf{97.33} & \textbf{75.00} & \textbf{36.00} & 5.70 & \textbf{55.15} \\
\bottomrule
\end{tabular}
}
\caption{Summary of the best LLM performances on each of the benchmark tasks and overall \texttt{ChildPlay} performance. LCL1 corresponds to validity testing, LCL2 to construct generation.}
\label{tab:overall_performance}
\end{table}

We find GPT-4o to be the best model across all tasks, but not on each task individually. We also find a significant performance increase when going form GPT-3.5 to GPT-4 and when going from GPT-4 to GPT-4o. GPT-4o-mini performs slightly better than GPT-3.5 but significantly worse than GPT-4. Finally, with the highest combined score of around 55\%, there is still significant room for improved performance in the \texttt{ChildPlay} benchmark. The largest potential for improvement lies in the Battleship and GtS benchmark tasks.

\section*{Discussion}
The simple games of \texttt{ChildPlay} allow us to assess reasoning and problem-solving abilities of LLMs while minimizing the data contamination issues most other benchmarks suffer from. Our results show that the series of GPT models tested in this work can produce valid suggestions for simple games but often struggles to do so sensibly. In addition, they fail to apply optimal strategies such as blocking winning moves of the opponent or playing winning moves. This reveals severe limitations of these LLMs in strategic reasoning and problem solving, highlighting important areas for future improvement. %Mixed results were found in interpreting simple shapes from grids, where GPT-4 and GPT-4o performed well. The poor performances across the board in Battleship, LCL, and GtS suggest problems of the models with rule understanding and spatial reasoning in more complex tasks. 

Many current LLM benchmarks used to assess LLM performance permit using stochastic patterns instead of genuine reasoning. 
%Rather than being general, this might be a symptom of the transformer algorithm which finds the correct data manifold via input tokens as features to arrive at the most likely answer given the training data, allowing to bootstrap reasoning patterns. 
In contrast, the tasks in \texttt{ChildPlay} test reasoning without relying on language, forcing the models to play the game rather than describe it and thus avoid "parroting" \cite{Borji2023StochasticPO,Duan2023FlocksOS,stochasticParrots,Henrique2023Stochastic,Zecevic2023Causal,bender2021dangers}. Such non-language benchmark tasks challenge models to generalize across different encodings and probe out-of-training-distribution scenarios. %Testing LLMs like GPT-4 (according to OpenAI, a contender for AGI \cite{Bubeck2023SparksOA}) beyond text, we show that they are mediocre at best at simple tasks outside their training data, with only little improvements in more recent GPT iterations.

The \texttt{ChildPlay} benchmark is most closely related to the "non-language" category of BigBench, which includes ASCII recognition, chess, and Sudoku, but none like ours in terms of spatial reasoning involving games or construction tasks \cite{srivastava2023beyond}. Additionally, LCL is related to benchmark tasks developed by Bubeck et al. \cite{Bubeck2023SparksOA}, where GPT-4 was tasked to generate images using JavaScript or LaTeX. However, unlike these examples, which are prone to training data contamination, the BTC games, LCL, and GtS remain abstract as they rely on new "languages" or representations, challenging the model to operate outside its learned distributions. Hence, we believe \texttt{ChildPlay} is a valuable addition to current benchmarks.

Another related set of benchmarks is provided by the ongoing ARC-AGI competition \cite{arcprize2024}, which tests systems under a GI definition akin to the one in our introduction (i.e., equivalent to higher-order cognition enabling general problem solving beyond specific skills or knowledge domains). The ARC-AGI results highlight current deep learning limitations as, at the time of writing, open-source models only achieve limited performance, particularly in abstract reasoning. This aligns with our \texttt{ChildPlay} findings, where LLMs struggle with strategic reasoning and rule adherence, showing challenges in generalization beyond training data. %Like ARC-AGI, with the GtS game, we provide a hidden benchmark without source code that is only accessible via our API to avoid future models overfitting to the benchmark.

Our findings reveal a limitation in current LLM development. Progress in fluency and efficiency does not entail comparable progress in reasoning and strategic decision-making. Modern LLM benchmarks primarily adapt Turing's test \cite{turing} to assess language processing, potentially overlooking deeper complexities. While models like GPT-4 and GPT-4o exhibit what Humboldt described as the "infinite use of finite means" \cite{Merrill2023FormalLA} as popularized by Chomsky \cite{Chomsky1957SyntacticS}, generating contextually relevant sentences \cite{Sweet1989OnLT}, this does not imply reasoning proficiency. %It is potentially simply advanced pattern imitation.

As one might have expected, we observe a systematic improvement in performance across GPT versions. However, this systematic progression is only reflected in some of the benchmark tasks. For instance, both GPT-4 and GPT-4o perform well in simple shape recognition but fail completely in Battleship and show no improvement in Connect-Four. This inconsistency suggests that advancements in model architecture do not necessarily translate into improved generalization capabilities. A possible explanation for this lack of improvement is overfitting. Modern LLMs are increasingly trained on massive, structured datasets, which may enhance their performance on familiar tasks but limit their ability to generalize. This likely explains the strong performance of GPT-4 and GPT-4o in the Shape Recognition Game, which may overlap with training data. This is in strong contrast to their struggles in LCL, Battleship, and GtS. %As LLMs become more specialized in pattern recognition, they may lose flexibility in applying learned knowledge to unfamiliar scenarios.

%One key aspect to interpret the performance progression is the current push toward more economical models. Efficiency in inference time and cost has become a major focus, often at the expense of deeper reasoning capabilities. This trade-off is particularly noticeable in GPT-4o-mini, which performs slightly better than GPT-3.5 overall but underperforms significantly in most tasks compared to GPT-4. %The prioritization of efficiency over reasoning might explain why newer models do not always outperform their predecessors in complex problem-solving tasks.

Given that these LLMs are able to explain details of the SMILES representation and how they connect to molecular structures and the corresponding graphs, we were disappointed by the poor performance of all the tested LLMs in combining spatial reasoning with chemical logic in the GtS benchmark task. Both the low response accuracies and chemical similarity scores indicate that, on average, hardly any of the proposed SMILES is reasonably close to the ASCII depiction of its molecular graph. We believe that this indicates that the tested LLMs are unable to connect changes in molecular connectivity, as represented by the ASCII of the molecular graph, to changes in SMILES strings. However, general purpose LLMs like the GPT series of models have been used previously for predicting molecular properties\cite{jablonka2024leveraging}, conducting inverse design\cite{jablonka2024leveraging}, and guiding reaction optimization\cite{boiko2023autonomous}. Our results hint at training data contamination and overfitting being likely responsible for these results rather than genuine chemical reasoning.

%Recently, OpenAI has released their results on the ARC-AGI competition using a new model they call o3. This model has achieved ground-breaking results in two modalities: a high-performance and a low-performance setting, after pretraining on 75\% of the dataset. The results of the untrained o3 were not released \cite{openaio3}. One of the arguments that we attempt to build here is the importance of evaluating the potential for AGI on benchmark tasks outside the learned distribution. Hence, \texttt{ChildPlay} provides a complementary set of benchmarks to the ARC-AGI competition.

\subsection*{Limitations}
\quad Overall, we pursued benchmarking under zero-shot conditions, but future work should also test Chain-of-Thought (CoT) prompting, Tree-of-Thought (ToT) prompting, and Few-Shot learning \cite{DBLP:journals/corr/abs-2201-11903, yao2023treethoughtsdeliberateproblem, brown}. These methods reduce individual task complexity and, thus, may improve performance. 

Additionally, \texttt{ChildPlay} sometimes uses single character output, potentially leading to single character token \textit{myopia}. However, most LLMs like GPT have a multi-character tokenizer \cite{brown}. This is problematic when encoding state information as single characters. Furthermore, binary outcomes from our benchmark tasks can bias perceived capabilities. %Thus, we also assess non-binary performance metrics such as move choices, missed wins or blocks (see Supporting Information, Section~\ref{app:heatmaps} and Figures\ \ref{fig:Tic-Tac-Toe_missed_wins_blocks}, \ref{fig:connectfour_missed_wins_blocks}), and more fine-grained assessment metrics. 
This can misrepresent gradual improvements and cause sharp evaluation changes \cite{schaeffer2023mirage}. Accordingly, the shape recognition and LCL tasks might benefit from continuous metrics, distinguishing smoother performance differences to reflect reasoning abilities more accurately. %For continuous model outputs (e.g., shape vertex coordinates), metrics like the mean squared error can quantify differences between predicted and actual values. 
This is why we incorporated both continuous and discontinuous metrics in the GtS benchmark, where we observe that the continuous chemical similarity score captures more subtle model improvements compared to the response accuracy. This is evident when comparing the results from GPT-4 and GPT-4o. The latter shows significantly higher chemical similarity scores without any improvements in response accuracy.

Moreover, following Schaeffer et al., the \texttt{ChildPlay} games may not reflect true generalization or emergent abilities \cite{schaeffer2023mirage}. Benchmarks with nonlinear metrics can exaggerate LLM weaknesses or strengths. For instance, a sharp failure in Battleship might not indicate poor strategy universally but rather under specific conditions. We believe this not to be the case as our benchmarks are simple and fully observable games. We diminish this effect further by evaluating move heatmaps and by also counting illegal moves, missed blocks, and missed wins for more detailed performance evaluation (cf. Supporting Information). %In the Shape Recognition Game, we believe that performance differences are best rationalized by data contamination. We think it is likely that the training data of the GPT-4 models contained similar examples. This would explain the dramatic improvement at that game as it already knew the answers. This is supported by the models' poor performance in GtS, which is essentially an advanced shape recognition task that is very likely not included in the training data.

Future \texttt{ChildPlay} enhancements could involve feedback loops to reduce task complexity. 
%We translated a small dataset from the Lego corporation (cf. Supporting Information \ref{app:lego_kits}) for future reference and development of LCL. Additionally, LCL can be simulated in both 2D and 3D as a multimodal benchmark. 
Another important aspect is our ASCII representations, which might not convey the board states sufficiently well. More complex symbolic representations might enhance model performance. Furthermore, we plan to benchmark additional models more thoroughly, including open-source LLMs. Our preliminary analysis shows comparable or worse performance relative to GPT-3.5, GPT-4, or GPT-4o (cf.\ Supporting Information, Section~\ref{app:others}). %We also intend to explore deep reinforcement learning as in Schrittwieser et al., Kaiser et al., and Silver et al.\ \cite{Schrittwieser_2020, kaiser2020modelbased, silver2017mastering}.
%\newpage

% \noindent{The following Cell Press journals require most research articles to include their methods in a \textbf{METHODS} or \textbf{EXPERIMENTAL PROCEDURES} section, which should appear after the discussion: \textit{Cell Biomaterials}, \textit{Chem}, \textit{Chem Catalysis}, \textit{Cell Reports Physical Science}, \textit{Cell Reports Sustainability}, \textit{Device}, \textit{Joule}, \textit{Matter}, \textit{Newton}, \textit{One Earth}, and \textit{Patterns}. For all other Cell Press journals, please \textbf{delete} the contents of this page from the template and refer instead to \textbf{STAR Methods} on page 9.}

\subsection*{Conclusions}
%The hypothesis that LLMs possess broader cognitive functions cannot be empirically validated with established benchmarking paradigms, which conflate linguistic performance with signs of intelligence, such as spatial reasoning and strategic thinking, and suffer from training-test contamination. 
With \texttt{ChildPlay}, we developed a set of game-based benchmark tasks for LLMs that minimize training data contamination and allow us to critically assess reasoning and problem-solving abilities. Our results reveal that while GPT models are capable of handling many tasks, they fail to generalize consistently across a diverse range of non-linguistic reasoning challenges and often show disappointing problem-solving capabilities. While significant improvements can be observed over model iterations, these gains are not consistent across all tasks. Performance in strategic games, construct generation, and construct recognition remain moderate at best, suggesting difficulties with rule-based and spatial reasoning tasks. The disappointing performance of all the tested LLMs on connecting depictions of molecules to the underlying SMILES representation hints at a lack of understanding chemical structures and their encoding schemes, suggesting any apparent proficiency in that regard likely resulting from overfitting rather than chemical reasoning. The inconsistent performance progression across the various benchmark tasks suggests that improvements in LLMs may be more focused on efficiency and fluency rather than fundamental advancements in reasoning.

%Our results challenge the prevailing narrative that LLMs exhibit emergent reasoning or understanding in a cognitive sense. While these models excel at generating fluent, contextually appropriate text, their outputs are fundamentally rooted in statistical pattern recognition derived from training data. Our research unveils limitations of current LLMs with respect to tasks that require advanced reasoning, which is crucial for more complex applications. %This has broader implications for AI, particularly in developing models capable of true reasoning and adaptive thinking. Practically, if GPT or other LLMs cannot reason accurately on novel data, they should not be trusted to do so in standard applications. This is critical, especially in scientific contexts, as many users may be biased toward believing the outputs of LLMs because they appear credible. Our benchmark illustrates these pitfalls, showing that if LLMs even struggle with simple games, users should be more critical of their outputs.

The key takeaway of our work is not just the performance of the individual models in specific games. %Altogether, our results suggest that the tested models have limited ability to generalize and reason in novel contexts. Additionally, we do not observe a consistent improvement of model capabilities across benchmarks as formal model capabilities are increased, which is highlighted by the rather small difference in \texttt{ChildPlay} average performance between GPT-3.5, GPT-4, and GPT-4o. This has broader implications for AI in developing models capable of true reasoning and adaptive thinking. 
Our research highlights how LLMs handle tasks requiring higher-order reasoning, which is crucial for advanced applications such as scientific research. Practically, if LLMs cannot reason accurately on novel data, they should not be trusted to do so in standard applications. %This is critical, especially in scientific contexts, as many users may be biased toward believing LLM outputs because they appear credible. 
That is, if LLMs already struggle with simple games, users should be more critical of their outputs.

% We believe that \texttt{ChildPlay} will inspire a new wave of LLM benchmarks. While the widely used multiple-choice questions are evaluating abilities akin to crystallized intelligence, i.e., involving the use of knowledge that has been acquired through past learning and experience, we believe that games assess abilities akin to fluid intelligence, i.e., the ability to solve new problems and use logic in new situations.

% The results caution against claims of emergent intelligence and highlight the importance of benchmarks like \texttt{ChildPlay} that assess reasoning beyond text generation. Future research should focus on how to enhance reasoning capabilities in LLMs, potentially integrating symbolic reasoning or reinforcement learning methods to address these limitations.

%%%  The following components should appear after the 
%%%  experimental procedures (also known as methods). 
%%%  For journals using STAR Methods, these components 
%%%  should appear immediately after the discussion 
%%%  (after any "limitations" or "conclusions" subsection 
%%%  within the discussion).
%\newpage

\section*{Supporting Information Summary}
Detailed description of the games and their rules, introductory prompts, additional performance metrics and move analysis, examples of game states, preliminary results for additional LLMs.

%%%  The resource availability section is required 
%%%  for all research articles. This component 
%%%  has 3 subsections: "lead contact," "materials 
%%%  availability," and "data and code availability." 
%%%  All 3 subsections must be included, even if no 
%%%  unique materials were generated in the study. 
%%%  Do not edit or change the names of the subheadings. 
%%%  No other subheadings or text are allowed in the 
%%%  resource availability section.

%\subsection*{Lead contact}
%%%  Authors are required to designate a lead contact, 
%%%  who will be responsible for communication with 
%%%  the journal before and after publication and is 
%%%  the arbiter of disputes, including concerns 
%%%  related to reagents or resource sharing. Only 
%%%  one author can be named the lead contact, and 
%%%  only the lead contact’s information may be 
%%%  provided in this section.

%Requests for further information and resources should be directed to and will be fulfilled by the lead contact, Robert Pollice at \texttt{r.pollice@rug.nl}.

%\subsection*{Materials availability}

%%%  This subsection must include a statement describing 
%%%  the availability of newly generated materials 
%%%  associated with the paper, including any conditions 
%%%  for access. If there are no newly generated materials 
%%%  associated with the paper, the statement should 
%%%  state this, e.g.: This study did not generate new 
%%%  materials.

%This study did not generate new materials.

\section*{Author Contribution Statement}
All authors contributed equally.
%%%  This component is required for most research papers. 
%%%  Mention each individual author with a statement 
%%%  outlining the contribution of each author to the work.

\section*{Acknowledgments}
%We thank x for insightful discussions.
MarvinSketch (version 24.4.2) was used to prepare chemical structure schemes (Chemaxon, \href{https://www.chemaxon.com}{https://www.chemaxon.com}).

\subsection*{Data and Code Availability Statement}

%%%  All original research papers must include a 
%%%  comprehensive and accurate “data and code 
%%%  availability” statement within the “resource 
%%%  availability” component of the paper before it 
%%%  is accepted for publication. These statements 
%%%  are structured and consist of three bulleted 
%%%  components. Each component must be present.

All original code, except for the hidden task Guess-the-SMILES, has been deposited in the public GitHub repository: \href{https://github.com/BlueVelvetSackOfGoldPotatoes/child-play}{https://github.com/BlueVelvetSackOfGoldPotatoes/child-play} \\

\bibliography{main}

\bigskip

%%%  In your References, please include only articles 
%%%  that are published (online publication and 
%%%  preprint servers are OK). Unpublished data, 
%%%  submitted and/or accepted manuscripts, abstracts, 
%%%  and personal communications should be cited within 
%%%  the text only ("unpublished data," "data not 
%%%  shown," "Alice Smith, personal communication") 
%%%  and not included in the references list. Personal 
%%%  communication should be documented by a letter 
%%%  of permission. Whenever possible, please make 
%%%  sure your .bib file has the complete author lists 
%%%  for each item (at minimum, the first 11 authors 
%%%  listed).

\newpage

\renewcommand{\thefigure}{S\arabic{figure}}
\setcounter{figure}{0}
\renewcommand{\thetable}{S\arabic{table}}
\setcounter{table}{0}
\renewcommand{\thepage}{S\arabic{page}}
\setcounter{page}{1}

%\subsection*{Additional Resources}
\begin{center}
\begin{Huge}
\textbf{Supporting Information}
\end{Huge}
\end{center}
\appendix
\tableofcontents
\section{Algorithms}
\subsection{Generating Games} 
Our program enables simulation, testing, and analysis of game-play of the different benchmarks found in ChildPlay. 

The program begins by setting up variables for debug outputs and game configurations, initializes the game environment based on these configurations, and sets up players accordingly. It then simulates a series of games, saving results and logs in JSON format, and generates heatmaps to analyze gameplay strategies and player decisions. The script also handles results aggregation and visualization, including the generation of bar plots of the results.

Our program is structured to simulate games with various players through a base class and its derived classes, each representing different player types. This class, the PlayerBase class, serves as the foundational class for all player types. It initializes player properties such as ID, name, and a debug flag, and it also manages message storage for debugging or interaction logging. A method is available that can store or print board states and plays based on the debug flag, facilitating debugging and tracking game interactions.

The LLMPlayer class, derived from PlayerBase, represents a player that utilizes a large language model, in our case, this player is either GPT-3.5 or GPT-4. Its constructor initializes the player with an ID, name, model details, and the game being played. It inherits from PlayerBase and extends functionality for LLM interactions. We introduce a method to print messages related to API interactions, and another that formulates a prompt for the LLM, invokes it, and processes its response. Then, a different method interprets the LLM's response to determine the validity of a move within the game's context.

The TextPlayer class, also derived from PlayerBase, represents a player who inputs moves via text, through the console - this is a human player. Its constructor initializes player-specific details and a callback function for receiving input. There is a method that manages receiving and validating player inputs according to game rules.

The RandomPlayer class, another derived class from PlayerBase, simulates a player making random valid moves based on the current game state. We have designed a method in this class to generate random possible moves that always comply with the game's rules given the present game state.

We have two possible game loops, one designed to execute a series of games between two players, and another to simulate one single game. The former takes parameters including the game instance, the two players, the number of games to play, and a debug flag. This function runs a loop for the specified number of games, recording results, tracking messages, and logging game states and moves. It returns a summary of the game results and collected messages for further analysis. The second game loop simulates a single game between two players by initializing the game, managing turn-based gameplay, and logging detailed move information. It returns a tuple with game messages, counts of invalid moves, a log of moves, and the game outcome. 

Other functions, all helper functions, enable basic functionality, for example, in producing multiple random moves to generate a dataset, or loading data from a JSON file. We can also create a comprehensive dataset from multiple games and save it to a JSON file, load it, and print the board state for a specific record index. As stated before, we can also generate heatmaps showing the frequency of moves in the case of the board games, or of predicted versus actual shapes in the case of the Shape Recognition Game.

\subsection{Tic-Tac-Toe}
For the Tic-Tac-Toe task we first initialize the game with optional settings. If no specific options are provided, it defaults to a 3x3 board size. Customizable options include setting the board size and a debug flag. Upon initialization, a method is called to set up or reset the game state. Additionally, it establishes a game name and provides a prompt that explains the rules and how to input moves for an algorithmic explanation of the code used to run Tic-Tac-Toe games).

The \textit{reset\_game} method is responsible for resetting the game by creating a board of the specified size, defaulting to 3x3, filled with spaces that represent empty cells. It also initializes the current player as "P1", sets up a list to track moves made, and marks the game status as not over.

The \textit{get\_text\_state} method constructs a text-based or ASCII representation of the current game state. It displays column and row headers for the board and formats the board with dividers to clearly delineate rows and columns.

The guess method manages a player's move by first validating it to ensure it falls within the board boundaries and that the chosen position is not already occupied. If the move is valid, it updates the board with the player's symbol ('X' for player index 0 and 'O' for player index 1). After each valid move, the method checks for a win or a tie. If neither condition is met, it switches the turn to the next player. If a player makes a wrong move they lose the game.

The \textit{check\_win} method determines if the current player has won by checking all possible winning conditions. This includes vertical, horizontal, and diagonal lines. It returns True if any of these conditions are met with consecutive symbols matching the current player's.

The \textit{check\_tie} method checks for a tie by determining if all cells are filled without any player achieving a winning condition.

A \textit{switch\_player} method is available that alternates turns between the two players, switching control from "P1" to "P2" after each valid move.

\subsection{Connect-Four}
The Connect-Four constructor initializes the game with optional settings as well, such as row and column size and debug mode. If no specific options are provided, the game defaults to a 7x7 grid. Otherwise, it reads the values from the provided options dictionary. The constructor sets up basic game properties, including a reset board, a tuple to track the last move, the game over status, and the current player. It also includes a game instruction prompt.

The \textit{reset\_board} method resets the board to a 7x7 grid (or a custom size if specified) filled with dots (.) to represent empty cells.

The \textit{check\_tie} method determines if the game has ended in a tie. It checks if the top row of the board is completely filled without any winner being declared. This method ensures that the \textit{check\_win} method returns False before declaring a tie.

The \textit{check\_win} method assesses whether a winning condition has been met after a move. It examines the cell corresponding to the last move and checks horizontally, vertically, and diagonally (in both positive and negative directions) for four consecutive identical symbols (either 'X' or 'O').

The guess method handles a player's move attempt. It returns an invalid move if the attempt is made outside of valid columns or on a full column. For a valid move, it updates the board, checks for a win or a tie, and switches to the next player if the game is still ongoing.

The \textit{get\_text\_state} method generates a string representation of the current board state, displaying 'X' and 'O' for the two players.

The \textit{switch\_player} method alternates the active player between "P1" (Player 1) and "P2" (Player 2).

The \textit{board\_size} property returns the number of columns in the board, which is necessary for validation and display purposes.

\subsection{Battleship}
The Battleship constructor initializes, once again, the game with optional settings. If no options are provided, it defaults to a 5x5 board size. The constructor allows customization of the board size and a debug flag. It sets up four game boards for each player—two for ship placement and two for tracking guesses—using nested list comprehensions. Additionally, it initializes lists to keep track of the ship positions for both players, 2 lists each, 4 in total - these correspond to the players own boards which they can see both the ships and hits, and their adversaries, where they can see only their own hits and misses. A dictionary, \textit{ship\_types}, defines ship names and their sizes, which scale according to the board size. The \textit{place\_ships} method is then called to randomly place ships on each player's board. Initial game properties are also set, including the current player and a game prompt that explains the rules.

The board property dynamically returns the guess board of the current player, in case we want to display it or carry out logic checks.

The \textit{place\_ships} method is responsible for randomly placing ships on a specified board. It attempts to place each ship either vertically or horizontally without overlapping or touching another ship, including diagonally adjacent spaces.

The \textit{is\_space\_free} method checks if a specified space on the board is free to accommodate a ship of a given size and orientation. It ensures that there is no overlap with existing ships and that the ship is not placed directly adjacent to another ship.

The \textit{get\_text\_state} method provides a string representation of both the ship and guess boards for a specified player, formatted for easy reading. This functionality is useful for displaying the game state in a text-based interface.

The guess method processes a player's guess by checking if it hits a ship and updating the boards accordingly. It also switches the player after each guess and checks for a win condition. Hits are marked with 'X' and misses with 'O' on the guess and ship boards.

The \textit{check\_win} method determines if the current player has won by checking if all positions of the opponent’s ships have been hit.

The \textit{switch\_player} method alternates turns between the two players after each guess.

\subsection{Shape Recognition}
The Shape Recognition Game starts by defining two constants, \textit{empty\_character} and \textit{full\_character}, which are represented by "0" and "1" respectively. These constants denote empty and filled cells in the grid.

One of the primary functions in the script is \textit{bar\_plot\_shapes}, which generates bar plots illustrating the counts of correct and incorrect answers for each shape. This function takes several parameters, including \textit{base\_path} for saving plots, models which is a list of GPT-type models used in the game, temperatures which refers to different temperature settings in the GPT-model predictions, and shapes representing the various shapes used in the game. The function iterates over the models and temperatures, loads game logs, counts the correct and incorrect answers, and plots these counts in bar charts. The resulting plots are saved to specified directories.

The script also includes several shape-drawing functions. The \textit{create\_board} function initializes an empty board of specified dimensions filled with \textit{empty\_character}. The \textit{draw\_rectangle} function allows drawing a rectangle or square on the board by filling cells with \textit{full\_character}. The \textit{draw\_circle} function draws an approximate circle using the midpoint circle algorithm. The \textit{draw\_triangle} function draws an equilateral triangle on the board by filling cells within the triangle's bounds. The \textit{draw\_cross} function draws a cross on the board with a specified arm length centered at a given point.

A key component of the script is the \textit{Shapes} class, which manages the game's logic. This class initializes the game with optional settings such as board size and debug mode. It sets the game name, the shape to be drawn, and a prompt explaining the game rules. The \textit{reset\_game} method creates a new board and randomly draws the specified shape (square, triangle, or cross) on it. It shuffles possible answer options and resets game state variables. The \textit{get\_text\_state} method generates a text-based representation of the game board and possible answers for display. The \textit{guess} method processes a player's guess, checks if it matches the correct shape, and ends the game, returning the outcome (win or loss). Methods \textit{check\_win}, \textit{check\_tie}, and \textit{check\_loss} are used to determine the game's outcome.

The script supports both visualization and analysis of game outcomes. The \textit{bar\_plot\_shapes} function helps visualize the correctness of guesses. The shape-drawing functions facilitate the random generation of different shapes for each new session. The Shapes class encapsulates the game’s logic, managing the board state, processing guesses, and determining game outcomes, thus supporting a single-player interaction where the player guesses the shape drawn on the board from a list of options.

\subsection{Guess-the-SMILES (GtS)}\label{app:GtS}

The hidden experiment, GtS, is a combination of two distinct parts: a Flask web application that interacts with a molecule generation and evaluation system, and a benchmarking script that tests different AI models' performance on predicting SMILES (Simplified Molecular Input Line Entry System) strings from molecular representations.

The generated molecules can be visualized as an ASCII print or a PNG image using functions like print\_mol\_ascii and draw\_mol\_coordgen. These representations provide users or models with a visual way to infer the SMILES string. Users can submit a predicted SMILES string for a given molecule, which the application evaluates using both chemical similarity, via molecular fingerprints, and string similarity, via Levenshtein distance. The evaluate\_prediction endpoint compares the predicted SMILES with the original, returning whether it is correct, along with similarity metrics.

The Flask app serves an HTML page where users can interact with the system by generating molecules, viewing their ASCII representations, and submitting predictions. The benchmarking script is designed to evaluate the performance of various AI models in predicting the correct SMILES string from molecular ASCII representations. Models and temperature settings are defined and can be used as hyper-parameters to influence the AI's predictions.

The results of these experiments are saved to CSV files for further analysis.

The data in the final results Table (cf. Main Text) was gathered by evaluating the performance of GPT-3.5 and GPT-4 language models, under different temperature settings (t=0, t=0.5, t=1, t=1.5). The goal was to assess each model's ability to correctly predict the SMILES ASCII string representation of a molecule given an ASCII depiction of its structure.

The process was repeated 100 times for each model and temperature setting. The data includes counts of correct and incorrect predictions, specific counts of incorrect SMILES, as well as computed accuracy, average chemical similarity, and average string distance.

The script to reproduce the experiment is available through the \href{https://github.com/BlueVelvetSackOfGoldPotatoes/child-play}{ChildPlay GitHub Repository}.

\subsection{Lego Connect Language (LCL)}
In the LCL script, we enable validating, and visualizing Lego-like constructs with different types of players.

We start by defining LCLVisualizer, which handles the visualization of Lego constructs. Its methods include \textit{draw\_piece}, which draws a single Lego piece on a plot, and \textit{display\_construct}, which displays and saves a complete Lego construct.

The LCLGame class manages the logic for generating and validating Lego constructs. It contains attributes such as pieces, a list of tuples representing Lego pieces, \textit{valid\_colors}, a list of valid colors for these pieces, and metrics, a dictionary for tracking validation metrics and construct details. Its methods include \textit{can\_place\_piece}, which determines if a Lego piece can be placed at a given position, \textit{find\_all\_valid\_positions}, which identifies all valid positions for placing new pieces, and \textit{build\_random\_valid\_assembly}, which constructs a valid Lego assembly by placing pieces in valid positions. Additionally, \textit{generate\_random\_piece} and \textit{generate\_random\_construct} are methods for generating random Lego pieces and constructs, respectively. The \textit{is\_valid\_construct} method checks the validity of a given construct, while \textit{generate\_valid\_or\_invalid\_construct} creates either a valid or invalid construct based on a given parameter. The class also includes methods like \textit{create\_tower}, \textit{create\_bridge}, and \textit{create\_staircase }for generating specific types of constructs, and \textit{validate\_construct}, which validates a player's response regarding the construct's validity. Finally, \textit{save\_metrics} saves the validation metrics to a file.

The RandomPlayer class represents a player that generates random answers. Its primary method, generate\_random\_answer, returns a random validity answer, either "valid" or "invalid."

The LLMPlayer class is designed for the GPT models to generate answers. It has attributes like model, which specifies the GPT-model type used, and temperature, which sets the temperature for the model's responses. Its methods include \textit{generate\_llm\_answer\_validity}, which produces a validity game answer using the LLM, and \textit{generate\_llm\_answer\_list}, which generates a list of Lego pieces using the LLM.

The main function orchestrates the simulation of games, including the generation of constructs, validation, and saving of results. It initializes both the game and visualizer, creates directories for saving experiment results, and defines the number of experiments, models, and temperatures. The function conducts two main types of experiments:

Validity Testing: In this experiment, constructs are generated (either valid or invalid), and players (either random or LLM) are asked to validate them. The results and visualizations of these constructs are saved.

Construct Generation: Here, LLM players are prompted to generate valid constructs based on a given prompt. These constructs are then validated, and results and visualizations are recorded.

The results of these experiments are saved to CSV files for further analysis.

The product of these experiments differs slighlty per type of game. We collected results (see Table \ref{tab:results_table}) as well as game state data based on player moves (see Table \ref{tab:game_logs_table}). Single-player games are simplified by setting irrelevant values to 0 (e.g. Player, turn, etc), but the overall format is the same. In the case of LCL, we save the results directly to two dataframes. For the construct experiment we record the model temperature, the model type, the experiment number, if the output is valid or not, the LLM's answer, and if the response is true or false. In the case of the validity experiment, we record the temperature, the model, the experiment number, if the output is valid or not, if the LLM's response is correct or incorrect, and the actual LCL construct.

\subsection{Game Data}

\begin{table}[H]
\centering
\begin{tabular}{|c|c|}
\hline
\textbf{Variable} & \textbf{Value} \\ \hline
Player   & A     \\ \hline
Move     & [X, Y] \\ \hline
Turn     & B     \\ \hline
\end{tabular}
\caption{Game logs with the details of a specific game state. The 'Player' variable (A) indicates the player number, 'Move' ([X, Y]) represents the coordinates of the move made, and 'Turn' (B) signifies the turn number in the game.}
\label{tab:game_logs_table}
\end{table}

\begin{table}[H]
\centering
\begin{tabular}{|c|c|}
\hline
\textbf{Variable}            & \textbf{Value} \\ \hline
P1 Wins             & C     \\ \hline
P2 Wins             & D     \\ \hline
Ties                & E     \\ \hline
P1 Wrong Moves      & F     \\ \hline
P2 Wrong Moves      & G     \\ \hline
\end{tabular}
\caption{Results summarizes the outcomes of a series of games. 'P1 Wins' (C) and 'P2 Wins' (D) indicate the number of games won by Player 1 and Player 2, respectively. 'Ties' (E) shows the number of games that ended in a tie. 'P1 Wrong Moves' (F) and 'P2 Wrong Moves' (G) represent the number of invalid moves made by Player 1 and Player 2, respectively.}
\label{tab:results_table}
\end{table}

\section{BTC Board Games}

\subsection{Prompts}\label{app:prompts}

\begin{table}[H]
  \label{table:games}
  \centering
  \begin{tabular}{p{0.2\textwidth} p{0.75\textwidth}}
    \toprule
    Game & Introductory Prompt \\
    \midrule
    \textbf{Battleship} & \textit{"Battleship is a two-player guessing game where each player has a fleet of ships on a secret grid and then takes turns guessing the locations of the opponent's ships. The objective is to sink all of the opponent's ships by correctly guessing their locations. O's in a board mean that the player selected a square to attack and there was no ship there - it's a miss. Had there been a ship there, instead of an O you would see an X. In your board, an <S> signifies a ship position, and a <\~> signifies the sea. Your input is just two numbers with a space in between, one for the row (from 0 to <self.board\_size-1>) and one for the column (from 0 to <self.board\_size-1>), like: 0 0, nothing else. Do not output anything else but the row col values."} \\
    \midrule
    \textbf{Tic-Tac-Toe} & \textit{"Tic-Tac-Toe is a two-player game played on a 3x3 grid. Players take turns placing their mark, X or O, in an empty square. The first player to place three of their marks in a horizontal, vertical, or diagonal row wins the game. You will play as player 1, therefore you play with X while your adversary plays with the symbol O. Your input is then a number (from 0 to 2) for the row followed by a space and another number (from 0 to 2) for the column, nothing else. Do not output anything else but the row col values else you lose."} \\
    \midrule
    \textbf{Connect-Four} & \textit{"Connect-Four is a two-player game. The pieces fall straight down, occupying the next available space within a column. The objective of the game is to be the first to form a horizontal, vertical, or diagonal line of four of one's own discs. In a board, player 1, you, plays with symbol X, while player 2, your opponent, plays with symbol O. Your input is just a number from 0 to 6, nothing else.  Do not output anything else but the col value else you lose."} \\
    \bottomrule
  \end{tabular}
  \caption{The three introductory prompts used for the board games in the ChildPlay suite.}
  \label{table:boardgames_prompts}
\end{table}

\subsection{Example}
\begin{figure}[H]
    \centering
    \begin{subfigure}[t]{0.145\linewidth}
        \centering
        \fbox{\includegraphics[width=\linewidth]{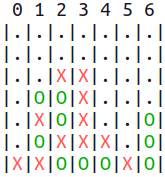}}
        \caption{Player X has won the Connect-Four game.}
        \label{fig:fig_board_3}
    \end{subfigure}
    \hspace{10pt} % Adds horizontal space between the subfigures
    \begin{subfigure}[t]{0.15\linewidth}
        \centering
        \fbox{\includegraphics[width=\linewidth]{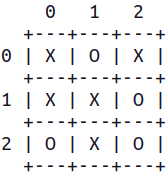}}
        \caption{Players tied at the Tic-Tac-Toe game.}
        \label{fig:fig_board_1}
    \end{subfigure}
    \hspace{10pt} % Adds horizontal space between the subfigures
    \begin{subfigure}[t]{0.35\linewidth}
        \centering
        \fbox{%
            \includegraphics[width=\linewidth]{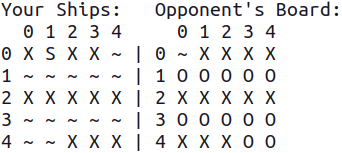}
        }
        \caption{A player has won the Battleship game.}
        \label{fig:fig_battleship_example}
    \end{subfigure}
    \caption{Examples of final board states in the three different board games.}
    \label{fig:Boargame_examples}
\end{figure}

Note that in the case of Connect-Four, a move consists of a singular scalar. A board state is shown after each play. Examples can be found in Figure~\ref{fig:Boargame_examples}. The move space in Tic-Tac-Toe corresponds to the full grid and the player selects a specific board square. In contrast, in Connect-Four it is a series of 7 columns. By definition, the player can only select a column and each piece piles on top of each other within it. Accordingly, this was our conceptual and algorithmic representation of the two games.

% \subsection[Move Mapping]{Move Mapping\footnote{See the left column for the model's moves, and the right column for the random player's moves.}}\label{app:heatmaps}

\subsection{Minimax versus Random Player}\label{app:minimax}

\begin{figure}[H]
    \centering
  \includegraphics[width=0.5\textwidth]{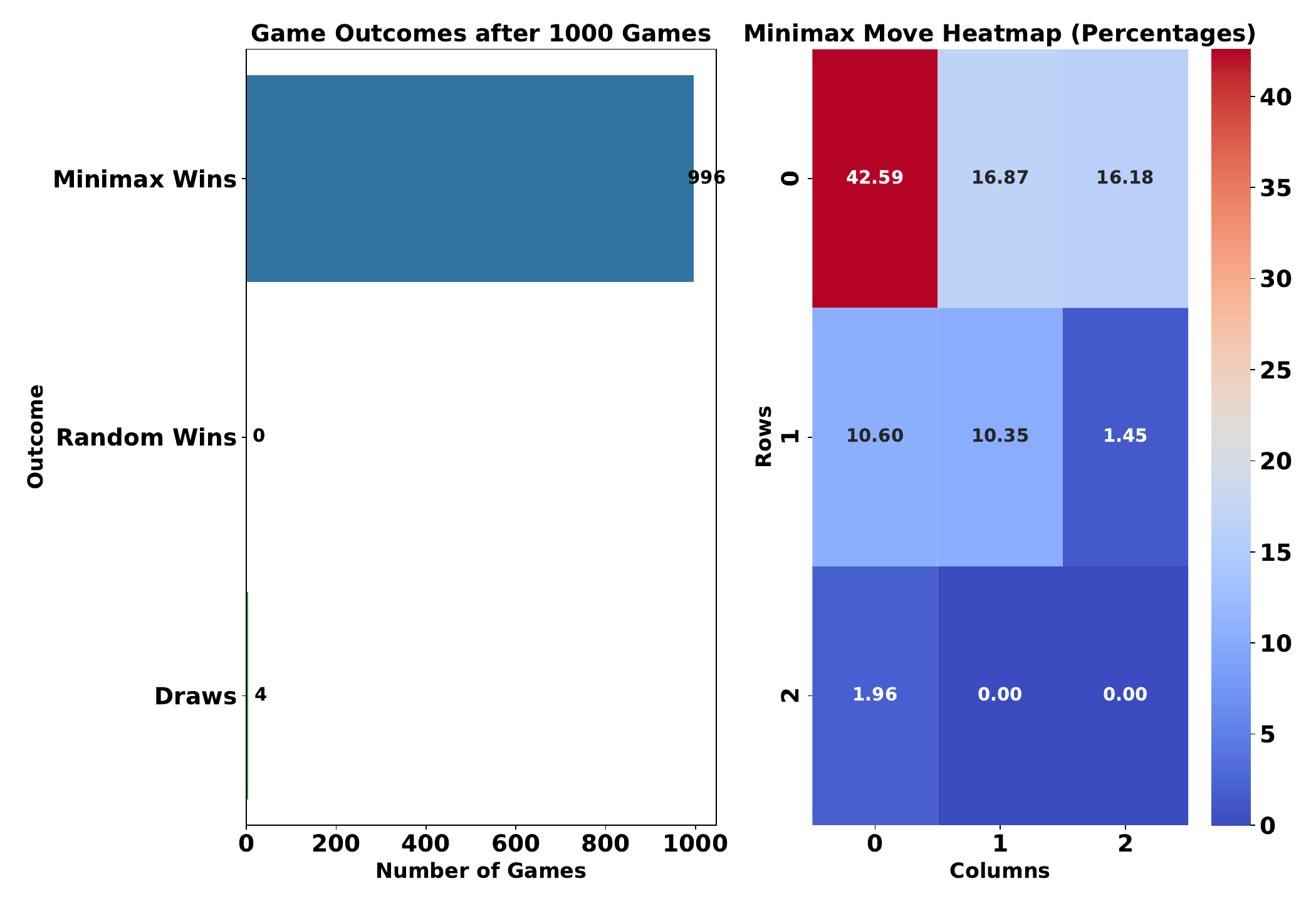}
  \caption{Minimax (agent) vs. random player - wins and losses on the left and a heatmap of move counts over the 1000 games for the minimax player on the right.}
  \label{fig:minimax}
\end{figure}

Figure~\ref{fig:minimax} presents the Tic-Tac-Toe results of the \textit{minimax} algorithm against the random player. The algorithm won 996 games, tied 4, and lost none. This replicates the results of previous research on Tic-Tac-Toe and we use that as a baseline for optimal play (see Section \ref{app:minimax})\cite{Swaminathan2020AnalysisOM, Alkaraz2020TicTacToeUT}.

\subsection{Game Outcomes}\label{app:btc_game_outcomes}

The Random Player's uniform move distribution seems to make it less likely to counter the LLMs, especially in Connect-Four. GPT-3.5 makes more mistakes than GPT-4 at Tic-Tac-Toe because it plays the middle row even when occupied, which produces an automatic loss. 

The GPT move biases in Battleship, Tic-Tac-Toe, and Connect-Four (BTC) can be observed in the heatmaps found in Section~\ref{app:heatmaps}. Errors, especially from GPT-3.5, generally increase with temperature, likely due to an increase in random move selection.

\subsubsection{Battleship}\label{app:battleship_outcomes}

\begin{figure}[H]
    \centering
    \includegraphics[width=0.85\linewidth]{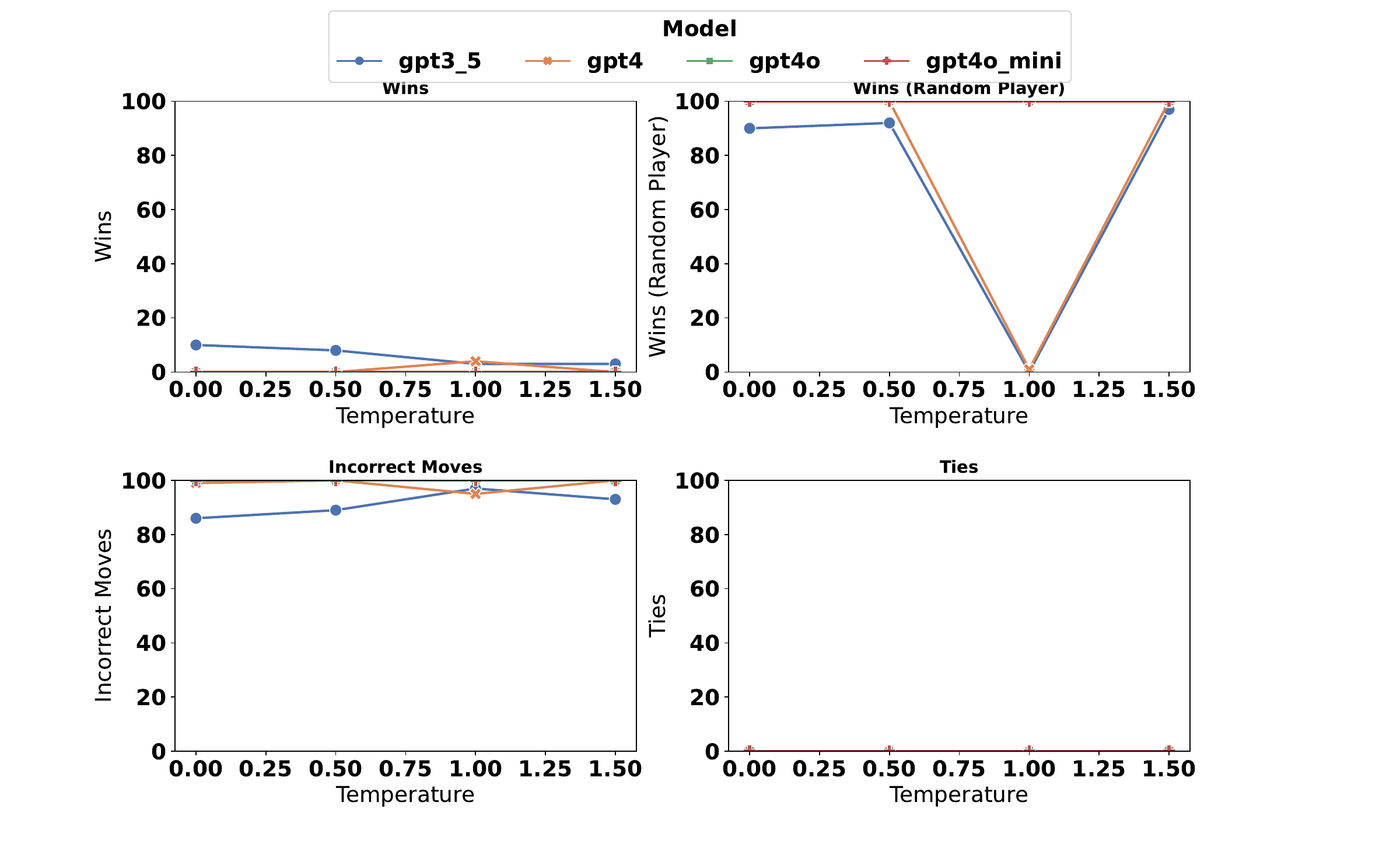}
    \caption{GPT models performance against a random player at Battleship.}
    \label{fig:Battleship_results}
\end{figure}

\subsubsection{Tic-Tac-Toe}\label{sec:tictactoe_outcomes}

% \begin{figure}[H]
%     \centering
%     \includegraphics[width=0.85\linewidth]{tictactoe_correct_incorrect_lines.pdf}
%     \caption{GPT models performance against a random player at the Tic-Tac-Toe game.}
%     \label{fig:tic_tac_toe_results}
% \end{figure}

\begin{figure}[H]
    \centering
    \includegraphics[width=1\linewidth]{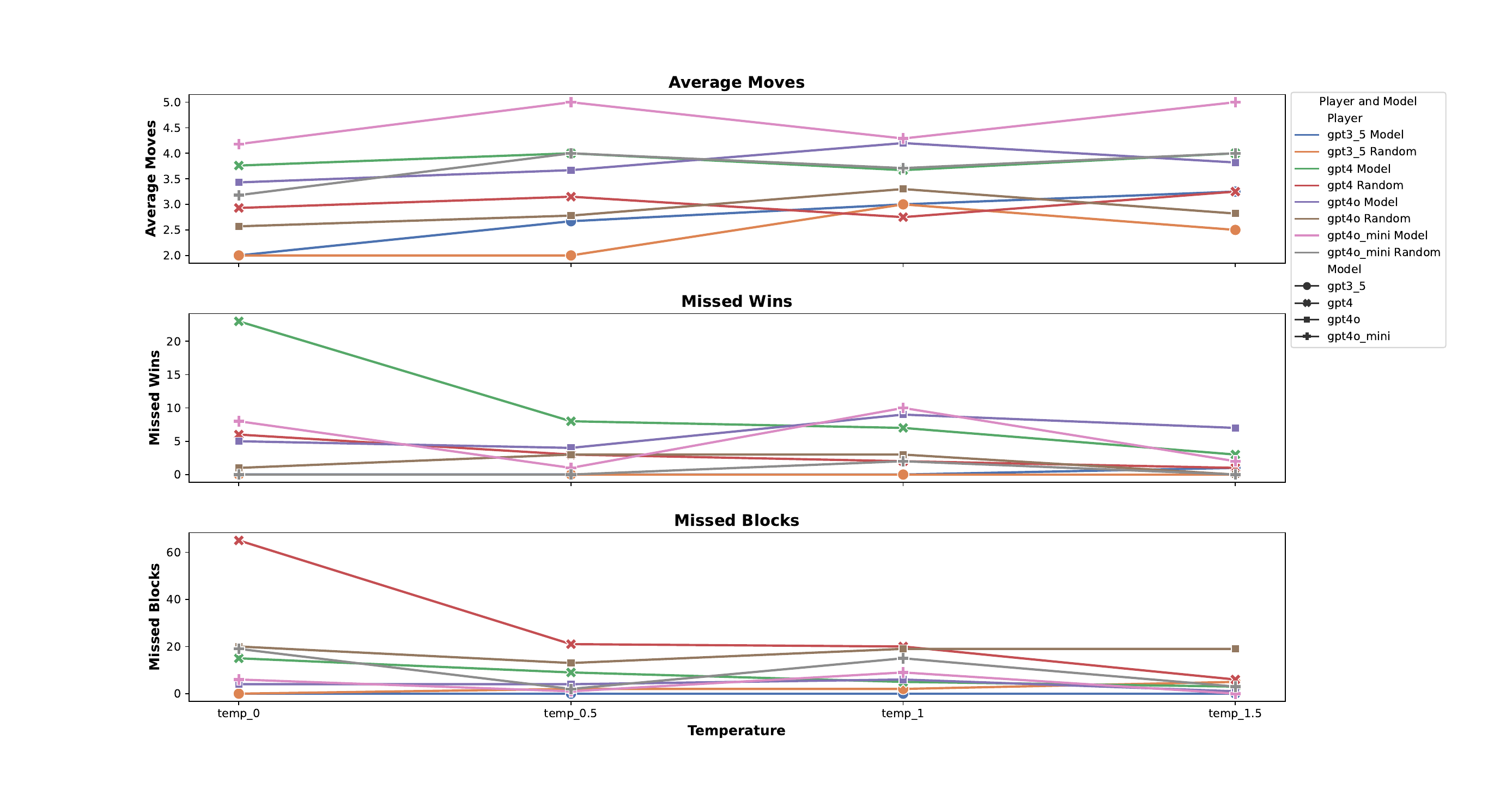}
    \caption{Tic-Tac-Toe: Missed wins, missed blocks, and average number of moves per game.}
    \label{fig:Tic-Tac-Toe_missed_wins_blocks}
\end{figure}

\subsubsection{Connect-Four}\label{app:connectfour_outcomes}

\begin{figure}[H]
    \centering
    \includegraphics[width=1\linewidth]{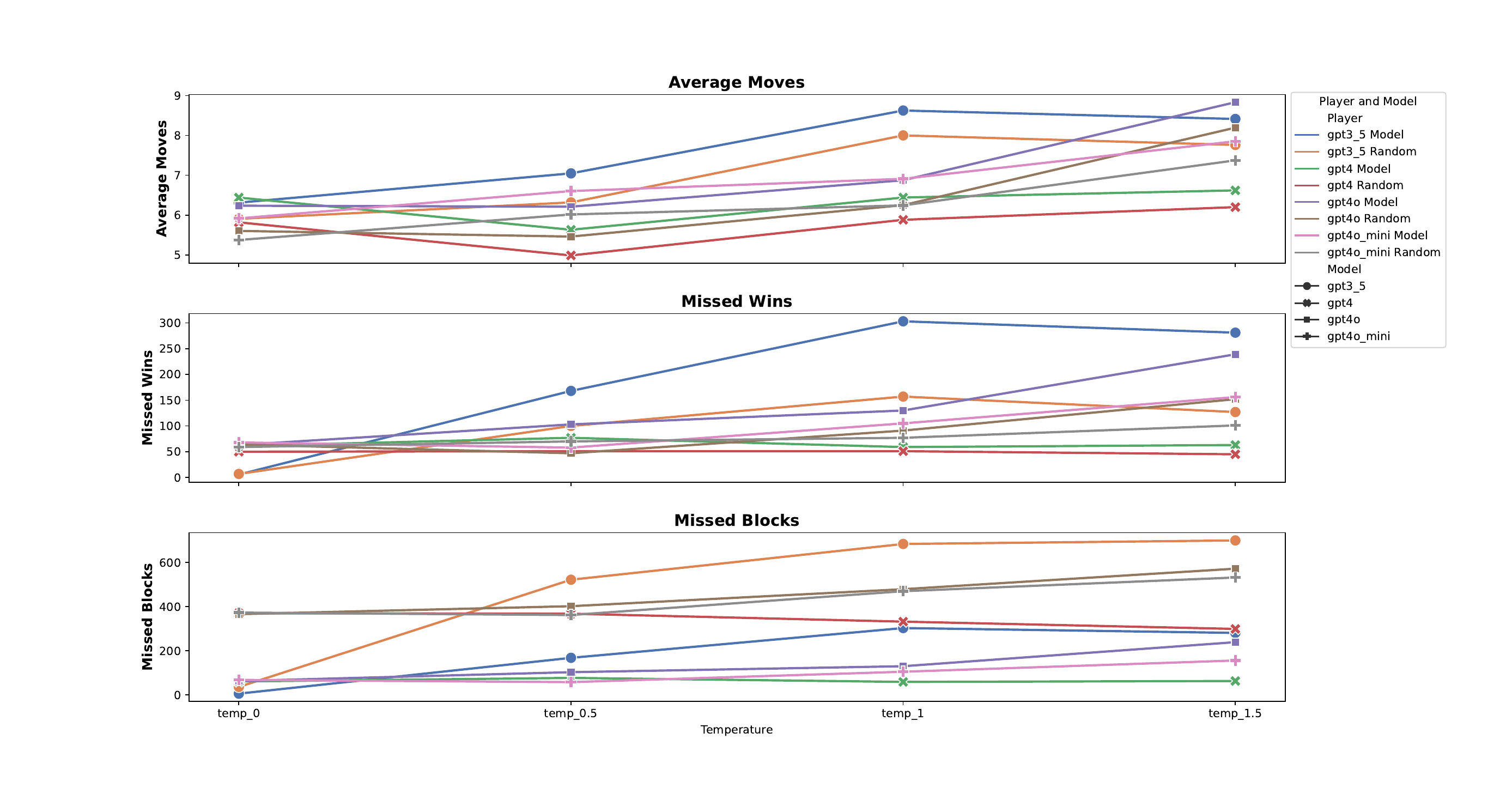}
    \caption{Connect-Four: Missed wins and missed blocks, as well as average number of moves per game.}
    \label{fig:connectfour_missed_wins_blocks}
\end{figure}

\begin{figure}[H]
    \centering
    \includegraphics[width=0.85\linewidth]{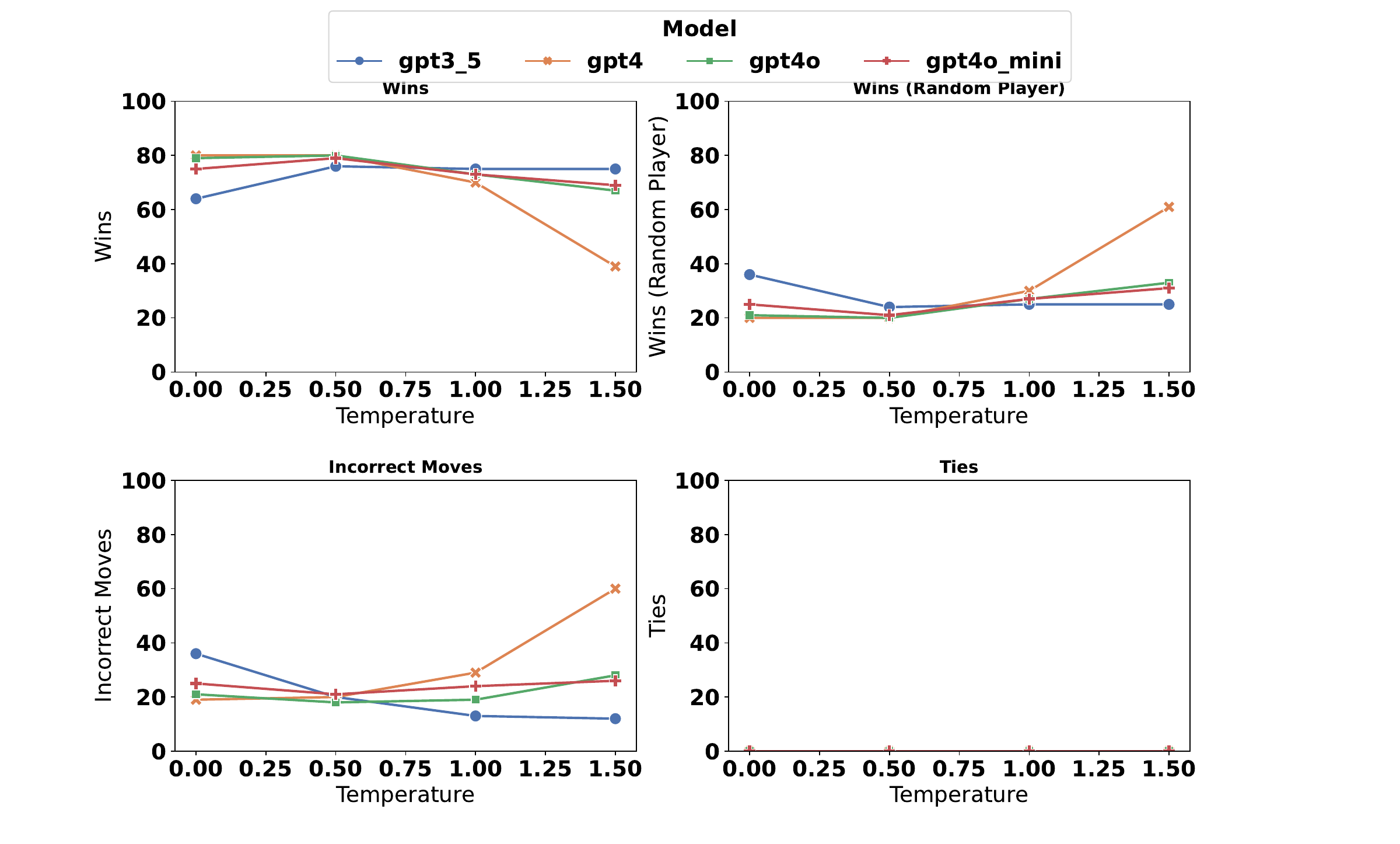}
    \caption{GPT models performance against a random player at the Connect-Four game.}
    \label{fig:connect_four_results}
\end{figure}

\subsection{Move Mapping}\label{app:heatmaps}
\subsubsection{Incorrect Move Probability}
The probability of incorrect moves $P_{\text{incorrect}}$ as a function of temperature is given by:

\[
P_{\text{incorrect}}(T) = \frac{\text{Incorrect Moves at } T}{\text{Total Moves}}
\]

For Tic-Tac-Toe, Connect Four, and Battleship, the results are as follows:

\[
P_{\text{incorrect, TTT}} = \left[ 4.7, 7.6, 7.6, 8.1 \right]
\]

\[
P_{\text{incorrect, CF}} = \left[ 1.8, 1.0, 0.65, 0.6 \right]
\]

\[
P_{\text{incorrect, BS}} = \left[ 3.583, 3.708, 4.042, 3.875 \right]
\]

These values indicate that the probability of incorrect moves varies across different games as the temperature changes. For Tic-Tac-Toe, the probability of incorrect moves increases with temperature, suggesting that higher temperatures may lead to less optimal decisions. In Connect Four, however, the probability of incorrect moves decreases as temperature increases, indicating a possible improvement in decision-making or adaptation to the conditions. In Battleship, the probability fluctuates but remains relatively stable.

\textbf{Average Number of Steps} \quad The average number of steps $\langle S \rangle$ as a function of temperature is calculated by averaging the steps observed at different temperatures:

\[
\langle S \rangle = \frac{1}{n} \sum_{i=1}^{n} \text{Steps at } T_i
\]

For Tic-Tac-Toe, Connect Four, and Battleship, the average number of steps are as follows:

\[
\langle S_{\text{TTT}} \rangle = 2.5
\]

\[
\langle S_{\text{CF}} \rangle = 5.0
\]

\[
\langle S_{\text{BS}} \rangle = 6.0
\]

These values show that the average number of steps taken in each game tends to stabilize regardless of temperature changes. This suggests that while the probability of making incorrect moves may fluctuate with temperature, the number of steps taken does not vary significantly.

\newpage

\subsubsection{Tic-Tac-Toe}\label{sec:heatmaps_tictactoe_all}
Below, see the left column for the model's moves, and the right column for the random player's moves.
\begin{figure}[H]
    \centering
    \begin{subfigure}[t]{1\linewidth}
        \centering
        \includegraphics[width=\linewidth]{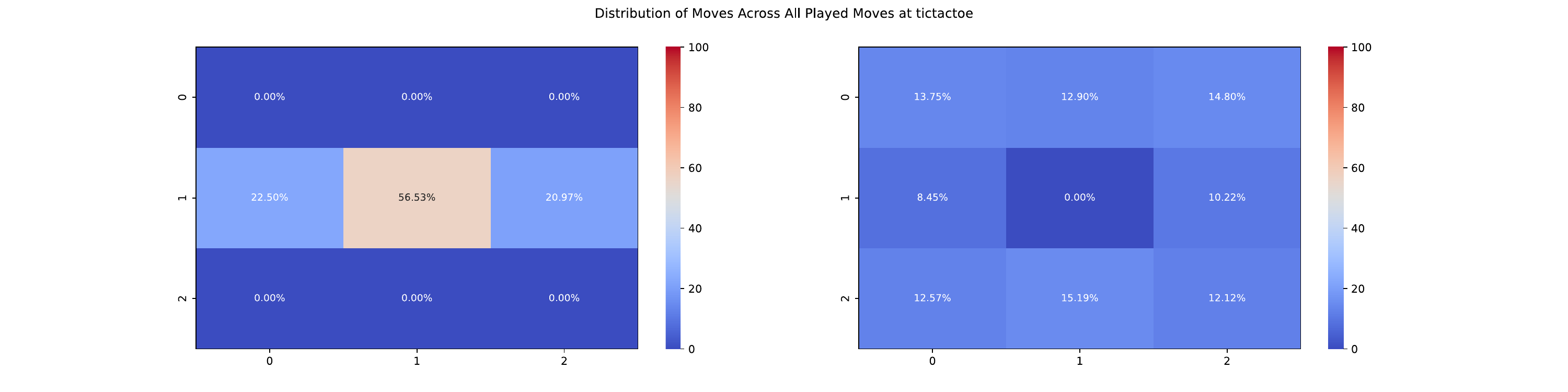}
        \caption{$t=0$.}
        \label{fig:tictactoe_1_3_5}
    \end{subfigure}
    \hspace{10pt} % Adds horizontal space between the subfigures
    \begin{subfigure}[t]{1\linewidth}
        \centering
        \includegraphics[width=\linewidth]{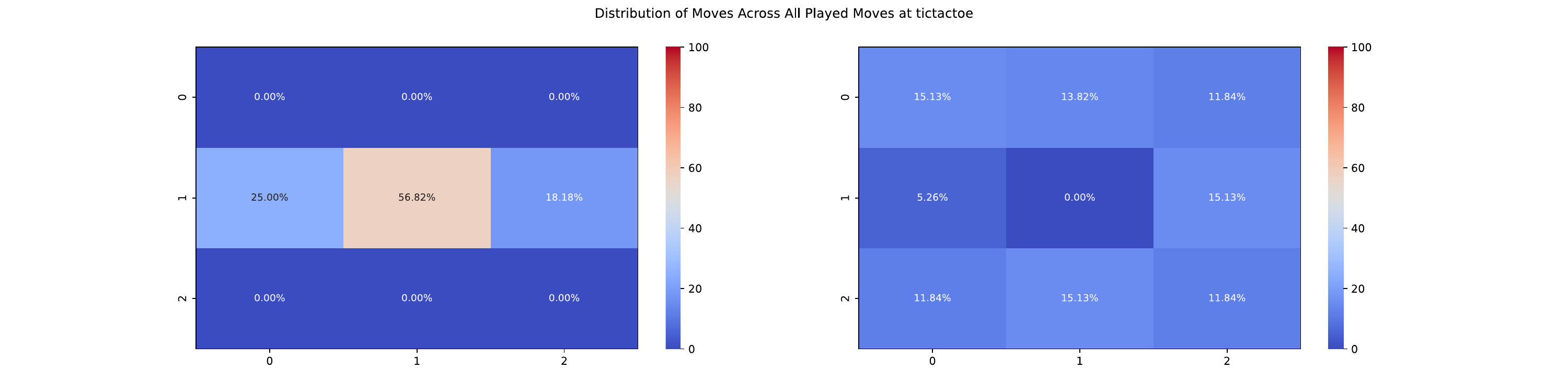}
        \caption{$t=0.5$.}
        \label{fig:tictactoe_2_3_5}
    \end{subfigure}
    \\ % Adds horizontal space between the subfigures
    \begin{subfigure}[t]{1\linewidth}
        \centering
        \includegraphics[width=\linewidth]{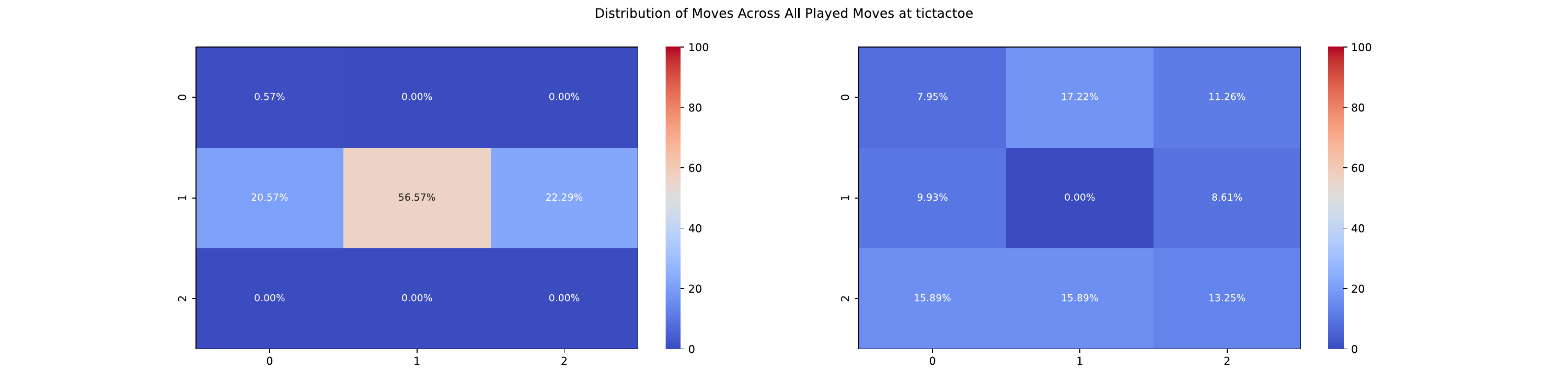}
        \caption{$t=1$.}
        \label{fig:tictactoe_3_3_5}
    \end{subfigure}
    \hspace{10pt} % Adds horizontal space between the subfigures
    \begin{subfigure}[t]{1\linewidth}
        \centering
        \includegraphics[width=\linewidth]{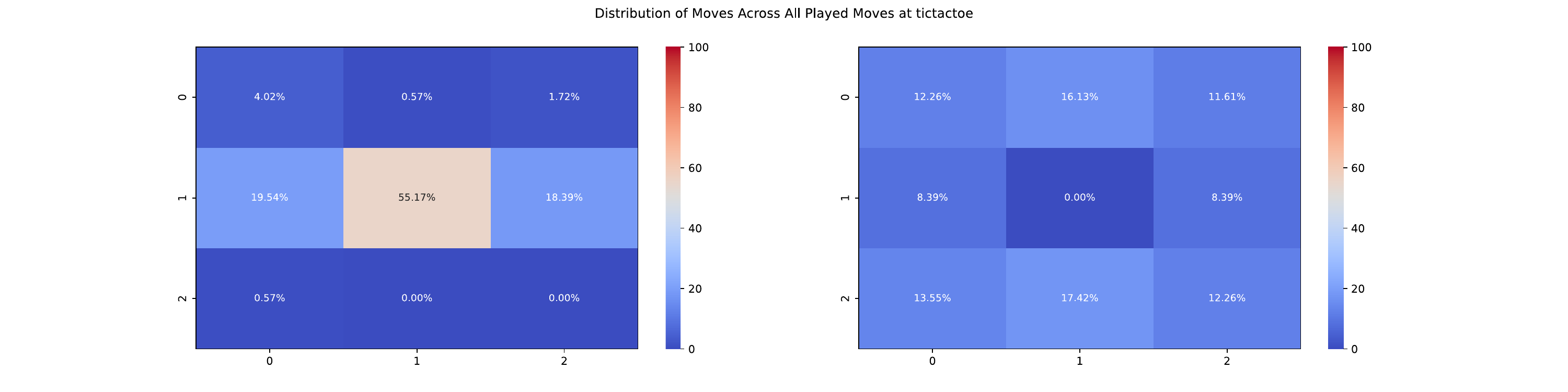}
        \caption{$t=1.5$.}
        \label{fig:tictactoe_4_3_5}
    \end{subfigure}
    \caption{Heatmap of model GPT-3.5's moves for the tic-tac-toe game, where 't' is temperature.}
    \label{fig:tictactoe_moves_3_5}
\end{figure}

\begin{figure}[H]
    \centering
    \begin{subfigure}[t]{1\linewidth}
        \centering
        \includegraphics[width=\linewidth]{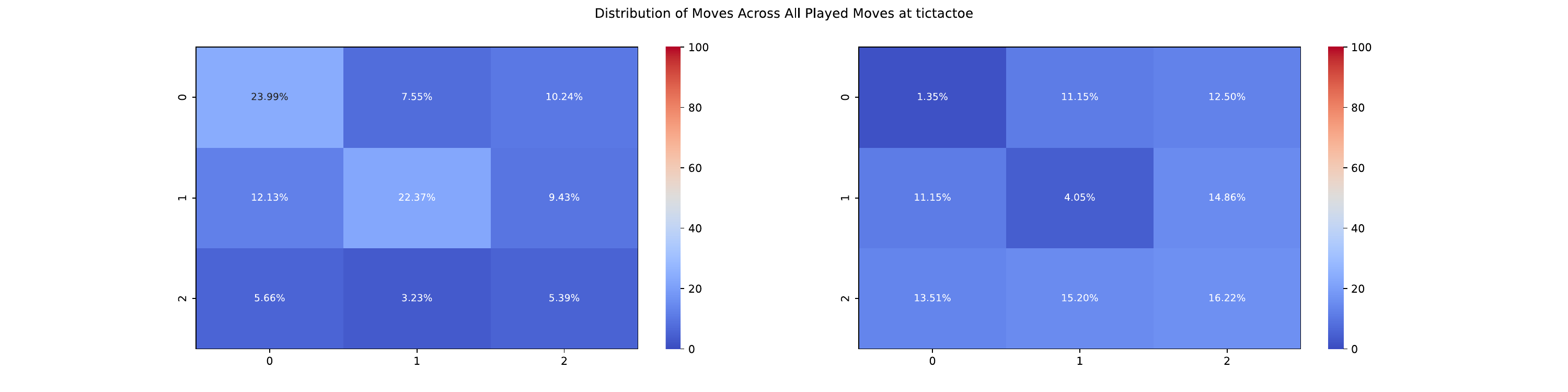}
        \caption{$t=0$.}
        \label{fig:tictactoe_1_4}
    \end{subfigure}
    \hspace{10pt} % Adds horizontal space between the subfigures
    \begin{subfigure}[t]{1\linewidth}
        \centering
        \includegraphics[width=\linewidth]{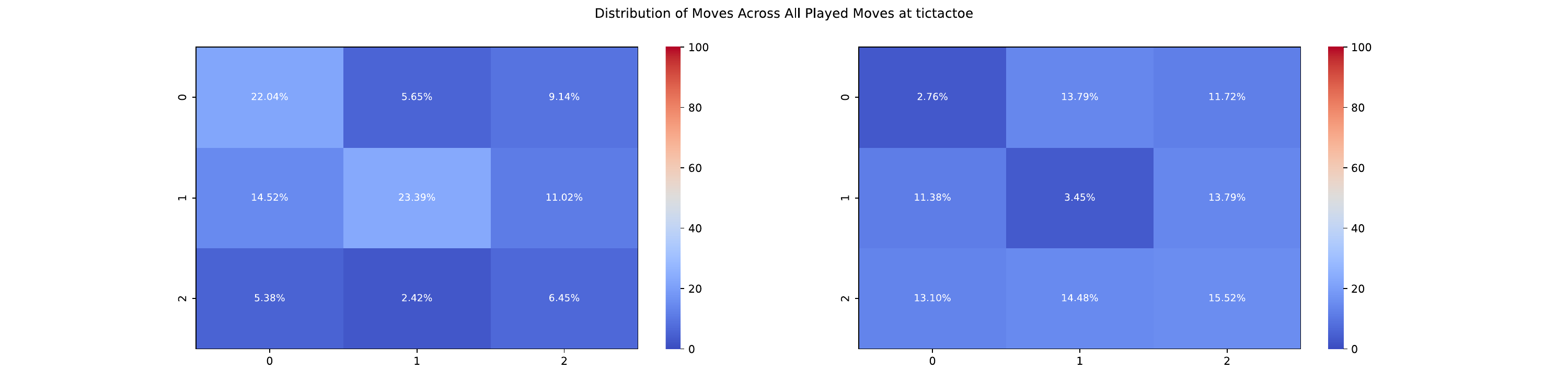}
        \caption{$t=0.5$.}
        \label{fig:tictactoe_2_4}
    \end{subfigure}
    \\ % Adds horizontal space between the subfigures
    \begin{subfigure}[t]{1\linewidth}
        \centering
        \includegraphics[width=\linewidth]{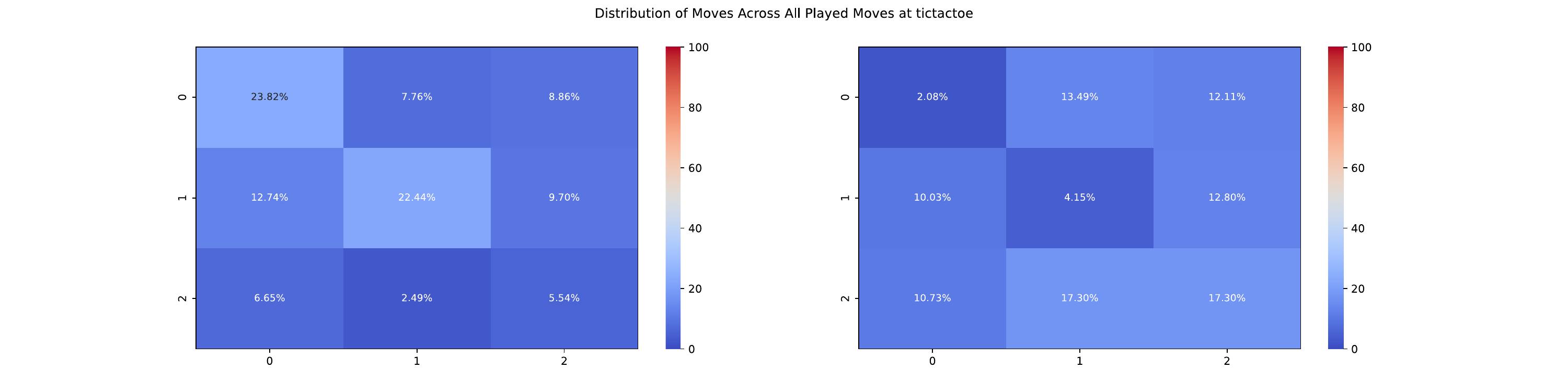}
        \caption{$t=1$.}
        \label{fig:tictactoe_3_4}
    \end{subfigure}
    \hspace{10pt} % Adds horizontal space between the subfigures
    \begin{subfigure}[t]{1\linewidth}
        \centering
        \includegraphics[width=\linewidth]{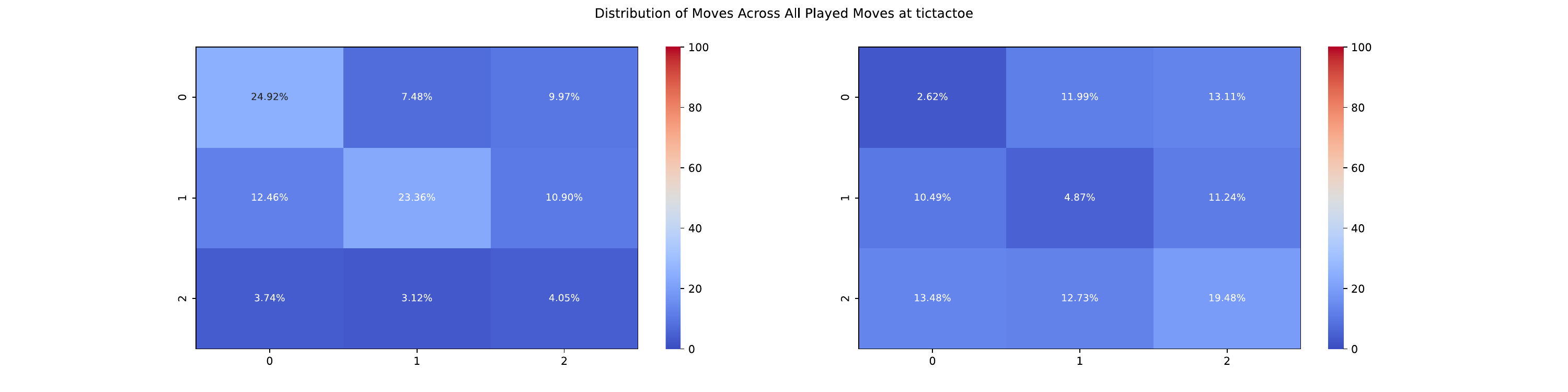}
        \caption{$t=1.5$.}
        \label{fig:tictactoe_4_4}
    \end{subfigure}
    \caption{Heatmap of model GPT-4's moves for the tic-tac-toe game, where 't' is temperature.}
    \label{fig:tictactoe_moves_4}
\end{figure}

% -----------------

\begin{figure}[H]
    \centering
    \begin{subfigure}[t]{1\linewidth}
        \centering
        \includegraphics[width=\linewidth]{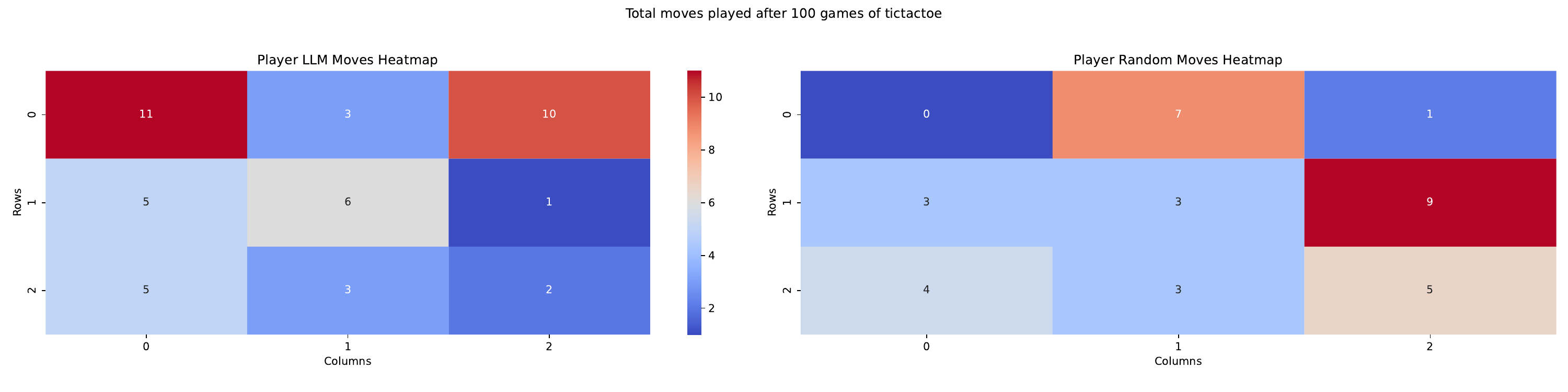}
        \caption{$t=0$.}
        \label{fig:tictactoe_4o_mini_0}
    \end{subfigure}
    \hspace{10pt} % Adds horizontal space between the subfigures
    \begin{subfigure}[t]{1\linewidth}
        \centering
        \includegraphics[width=\linewidth]{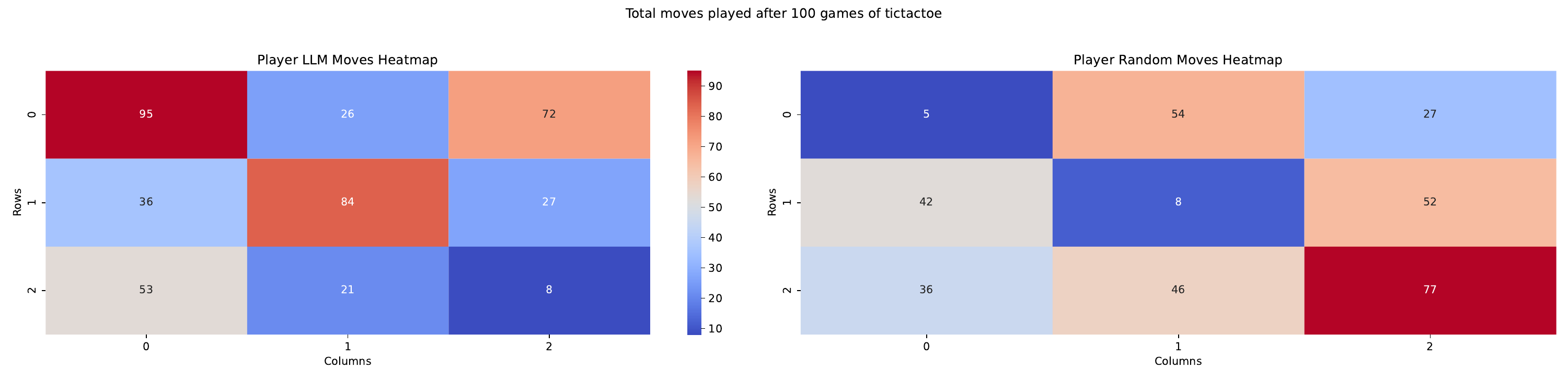}
        \caption{$t=0.5$.}
        \label{fig:tictactoe_4o_mini_0_5}
    \end{subfigure}
    \\ % Adds horizontal space between the subfigures
    \begin{subfigure}[t]{1\linewidth}
        \centering
        \includegraphics[width=\linewidth]{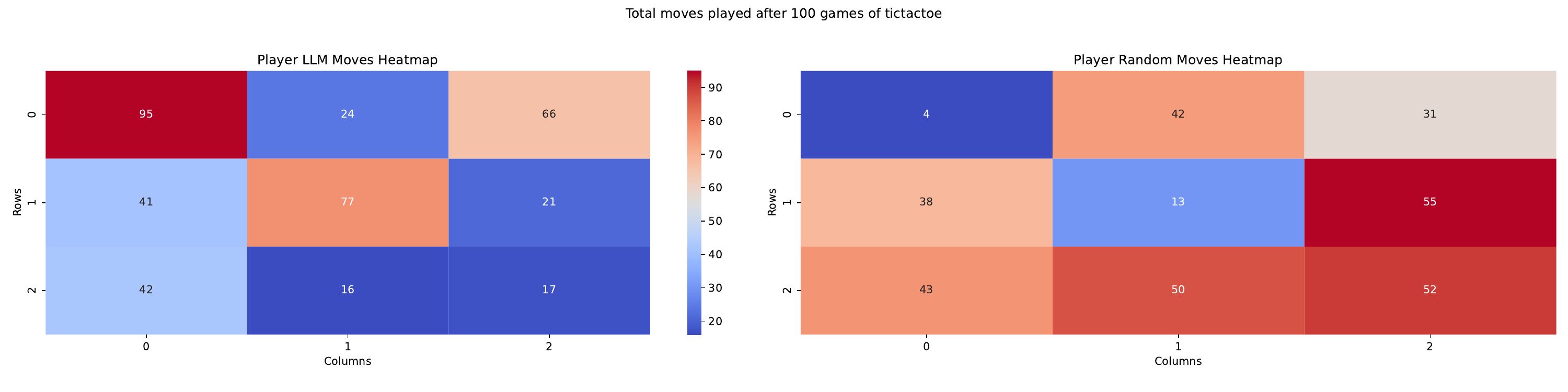}
        \caption{$t=1$.}
        \label{fig:tictactoe_4o_mini_1}
    \end{subfigure}
    \hspace{10pt} % Adds horizontal space between the subfigures
    \begin{subfigure}[t]{1\linewidth}
        \centering
        \includegraphics[width=\linewidth]{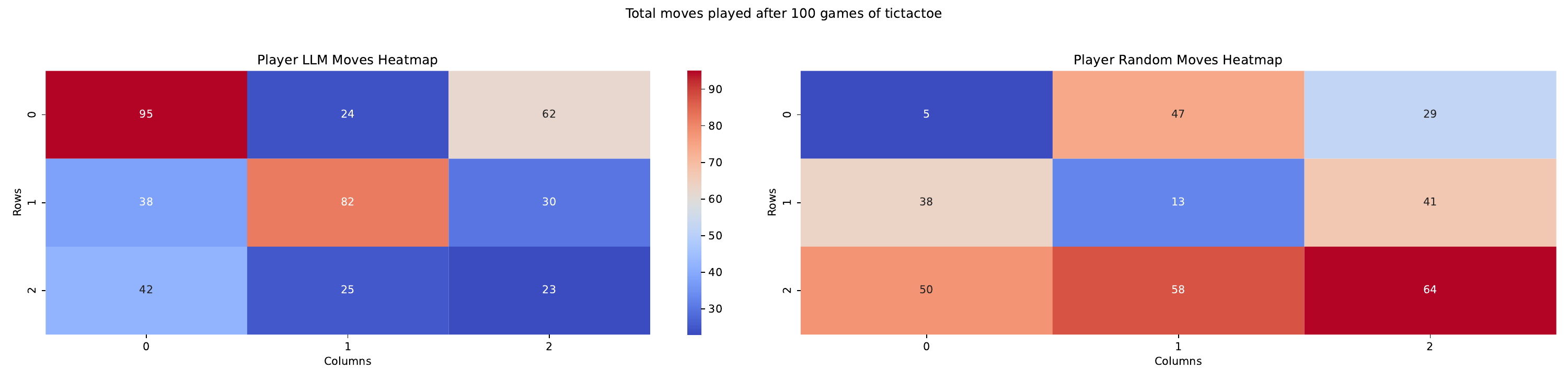}
        \caption{$t=1.5$.}
        \label{fig:tictactoe_4o_mini_1_5}
    \end{subfigure}
    \caption{Heatmap of model GPT-4o-mini's moves for the tic-tac-toe game, where 't' is temperature.}
    \label{fig:tictactoe_moves_4o_mini}
\end{figure}

\begin{figure}[H]
    \centering
    \begin{subfigure}[t]{1\linewidth}
        \centering
        \includegraphics[width=\linewidth]{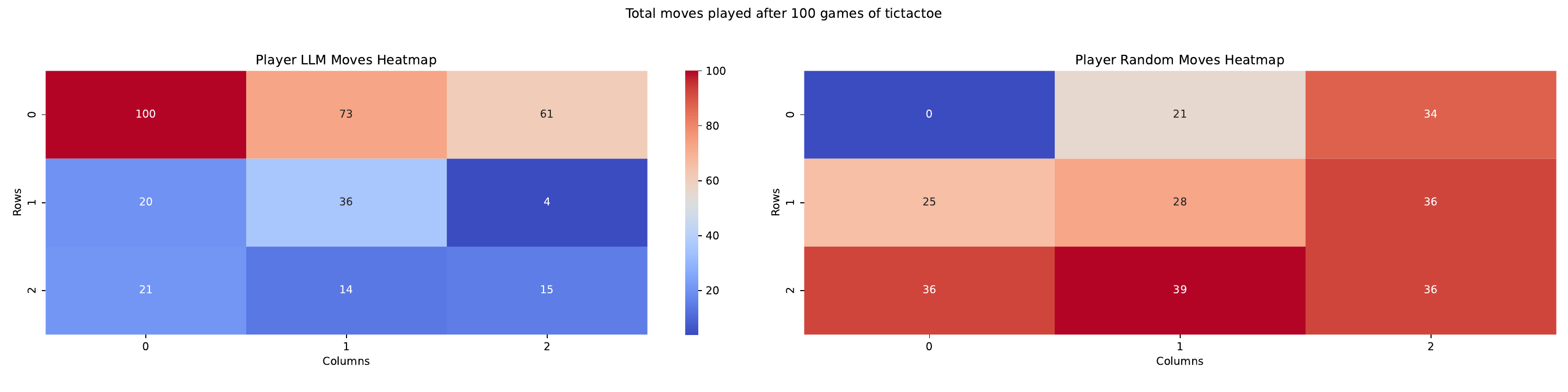}
        \caption{$t=0$.}
        \label{fig:tictactoe_4o_0}
    \end{subfigure}
    \hspace{10pt} % Adds horizontal space between the subfigures
    \begin{subfigure}[t]{1\linewidth}
        \centering
        \includegraphics[width=\linewidth]{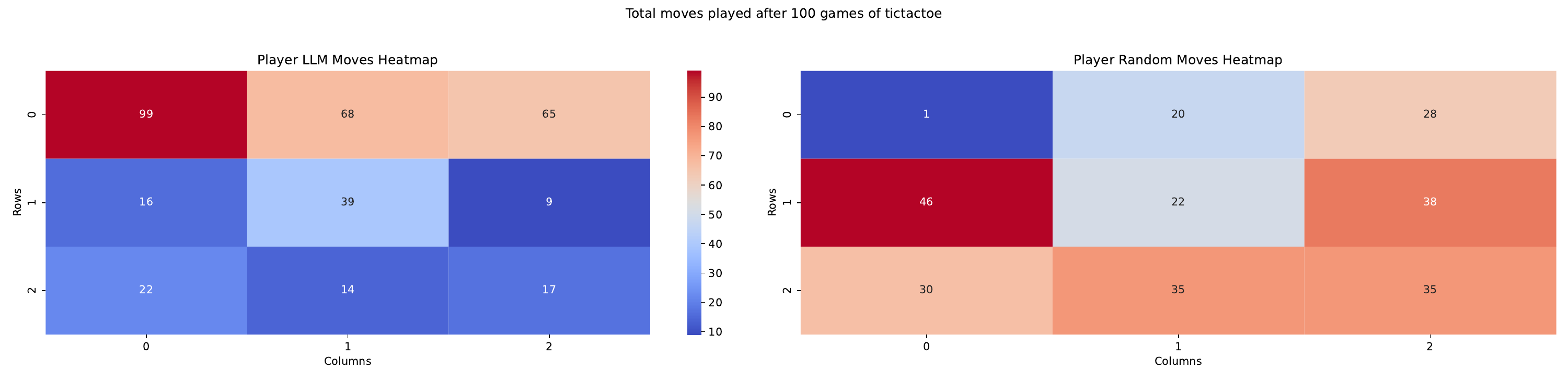}
        \caption{$t=0.5$.}
        \label{fig:tictactoe_4o_0_5}
    \end{subfigure}
    \\ % Adds horizontal space between the subfigures
    \begin{subfigure}[t]{1\linewidth}
        \centering
        \includegraphics[width=\linewidth]{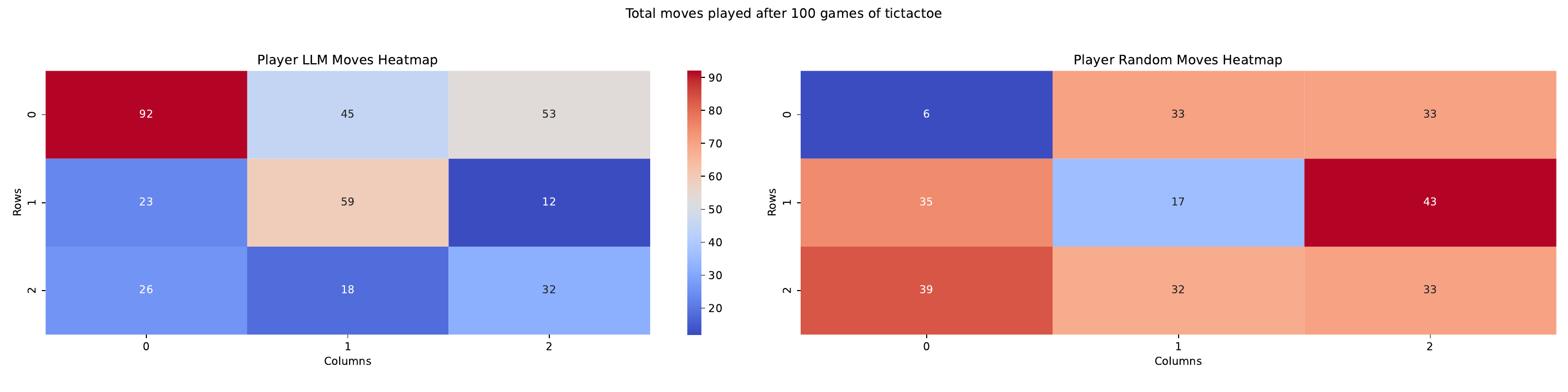}
        \caption{$t=1$.}
        \label{fig:tictactoe_4o_1}
    \end{subfigure}
    \hspace{10pt} % Adds horizontal space between the subfigures
    \begin{subfigure}[t]{1\linewidth}
        \centering
        \includegraphics[width=\linewidth]{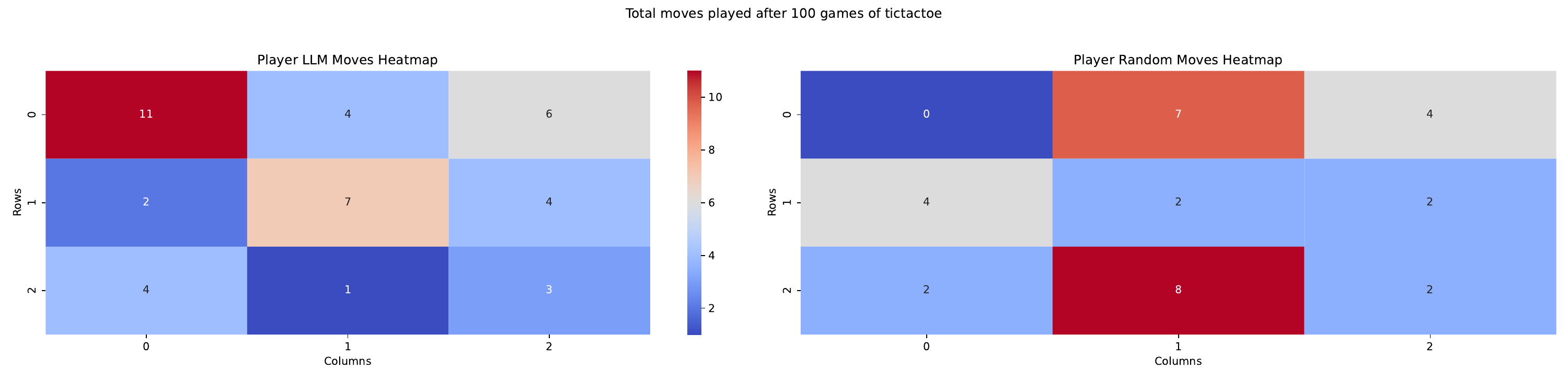}
        \caption{$t=1.5$.}
        \label{fig:tictactoe_4o_1_5}
    \end{subfigure}
    \caption{Heatmap of model GPT-4o's moves for the tic-tac-toe game, where 't' is temperature.}
    \label{fig:tictactoe_moves_4o}
\end{figure}

\newpage

\subsubsection{Connect-Four}\label{app:heatmap_connectfour}
Below, see the left column for the model's moves, and the right column for the random player's moves.

\begin{figure}[H]
    \centering
    \begin{subfigure}[t]{1\linewidth}
        \centering
        \includegraphics[width=\linewidth]{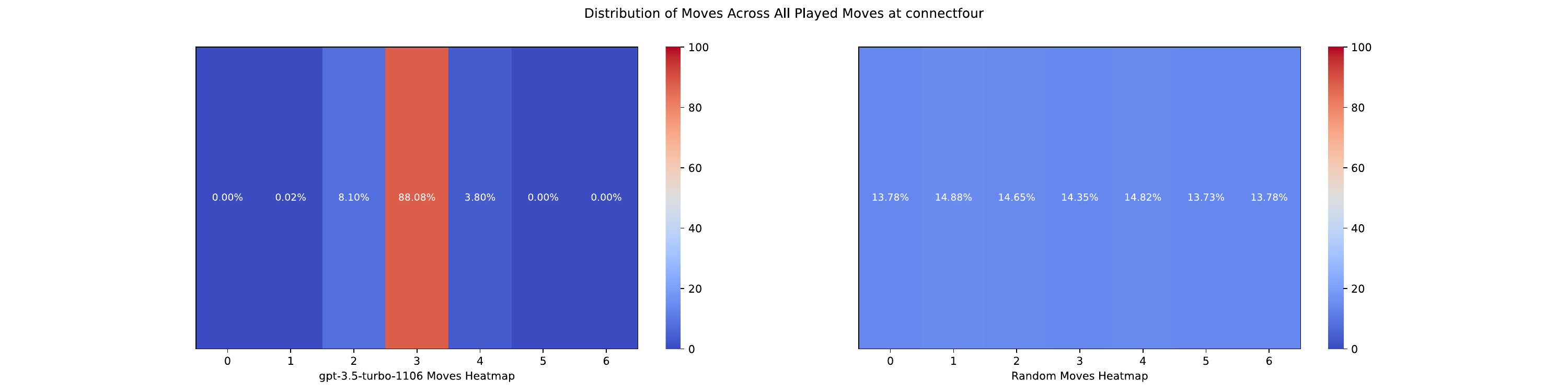}
        \caption{$t=0$.}
        \label{fig:connectfour_1_3_5}
    \end{subfigure}
    \hspace{10pt} % Adds horizontal space between the subfigures
    \begin{subfigure}[t]{1\linewidth}
        \centering
        \includegraphics[width=\linewidth]{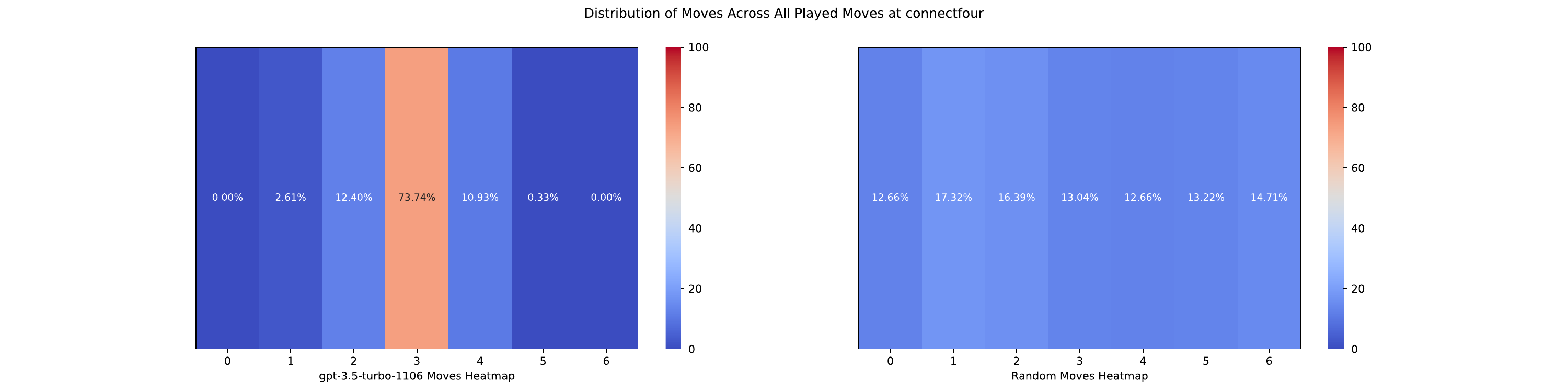}
        \caption{$t=0.5$.}
        \label{fig:connectfour_2_3_5}
    \end{subfigure}
    \\ % Adds horizontal space between the subfigures
    \begin{subfigure}[t]{1\linewidth}
        \centering
        \includegraphics[width=\linewidth]{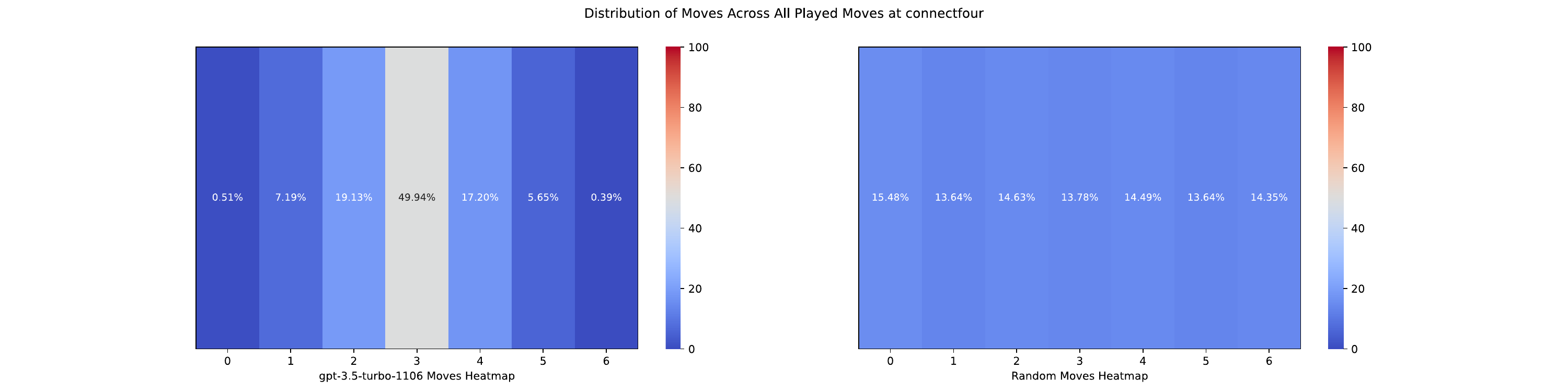}
        \caption{$t=1$.}
        \label{fig:connectfour_3_3_5}
    \end{subfigure}
    \hspace{10pt} % Adds horizontal space between the subfigures
    \begin{subfigure}[t]{1\linewidth}
        \centering
        \includegraphics[width=\linewidth]{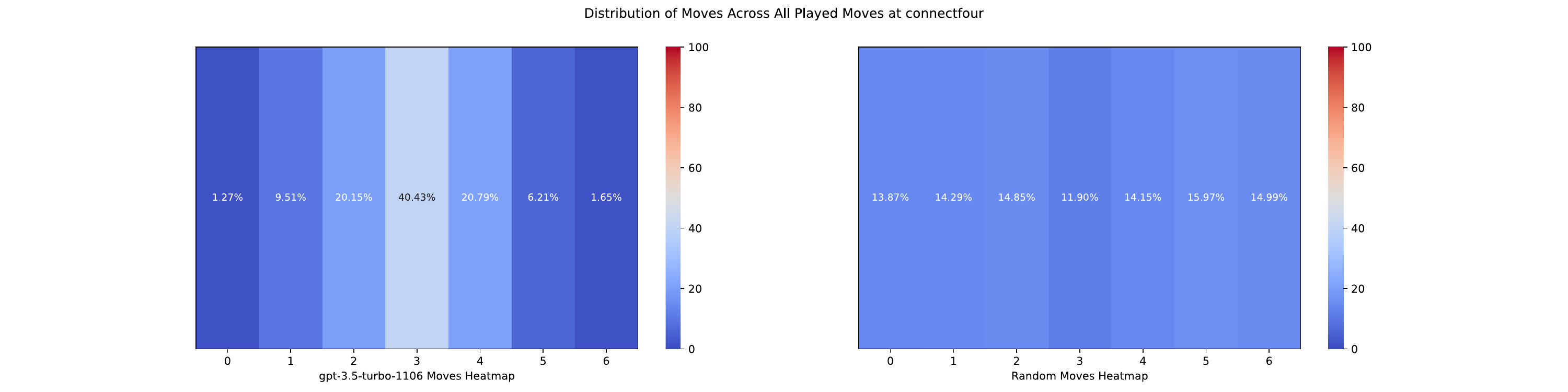}
        \caption{$t=1.5$.}
        \label{fig:connectfour_4_3_5}
    \end{subfigure}
    \caption{Heatmap of model GPT-3.5's moves for the connect-four game, where 't' is temperature.}
    \label{fig:connectfour_moves_3_5}
\end{figure}

\begin{figure}[H]
    \centering
    \begin{subfigure}[t]{1\linewidth}
        \centering
        \includegraphics[width=\linewidth]{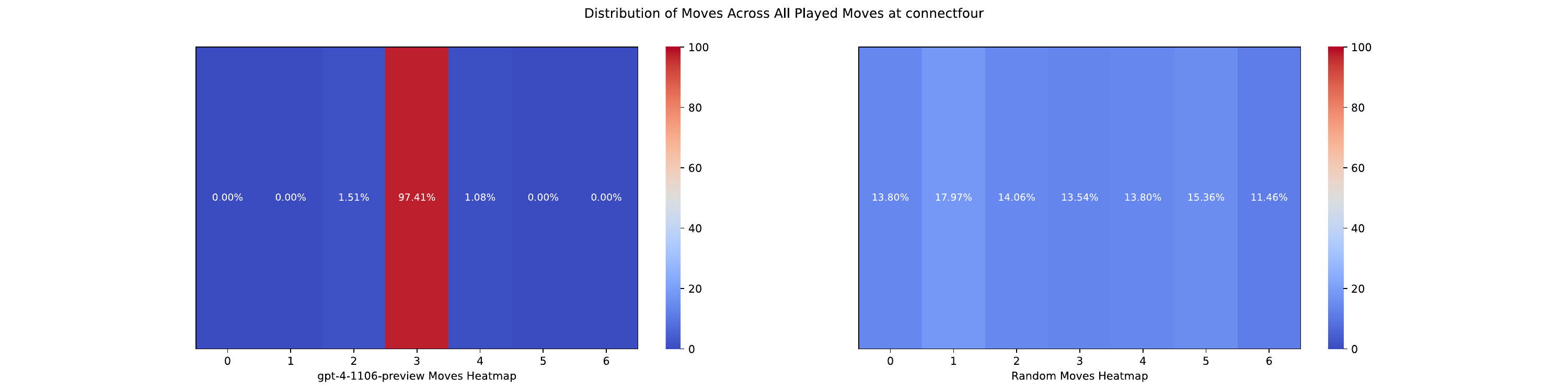}
        \caption{$t=0$.}
        \label{fig:connectfour_1_4}
    \end{subfigure}
    \hspace{10pt} % Adds horizontal space between the subfigures
    \begin{subfigure}[t]{1\linewidth}
        \centering
        \includegraphics[width=\linewidth]{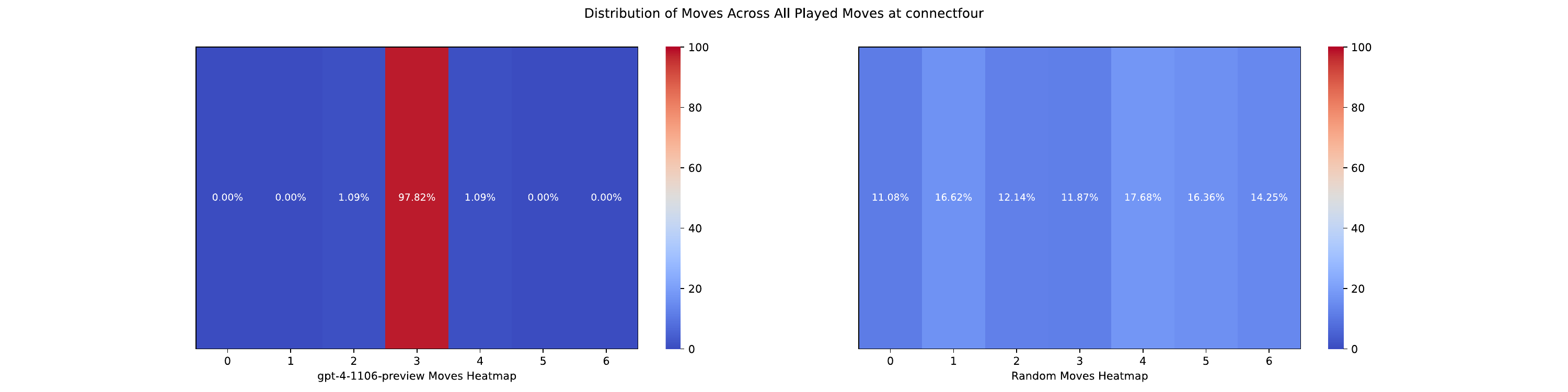}
        \caption{$t=0.5$.}
        \label{fig:connectfour_2_4}
    \end{subfigure}
    \\ % Adds horizontal space between the subfigures
    \begin{subfigure}[t]{1\linewidth}
        \centering
        \includegraphics[width=\linewidth]{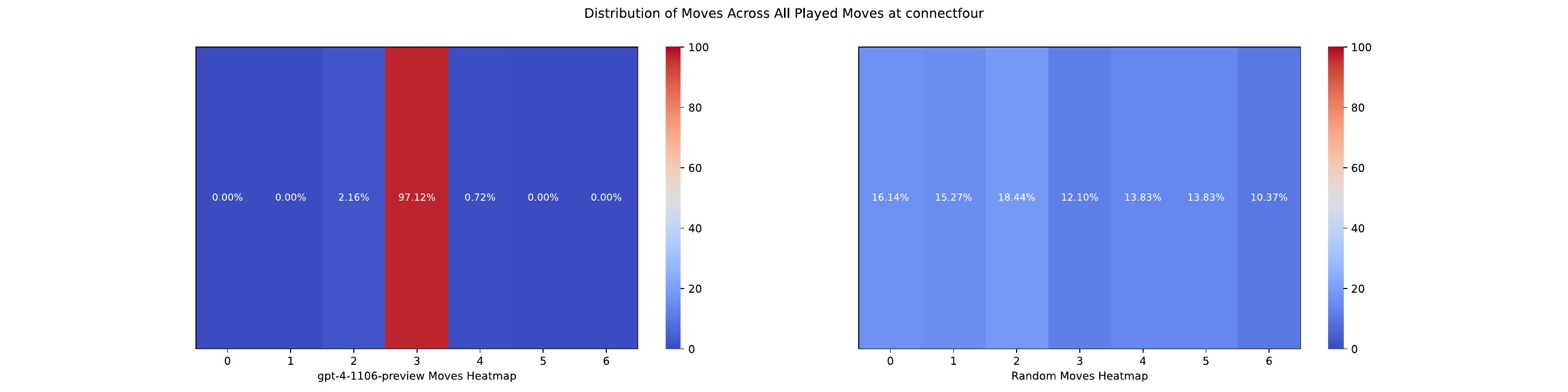}
        \caption{$t=1$.}
        \label{fig:connectfour_3_4}
    \end{subfigure}
    \hspace{10pt} % Adds horizontal space between the subfigures
    \begin{subfigure}[t]{1\linewidth}
        \centering
        \includegraphics[width=\linewidth]{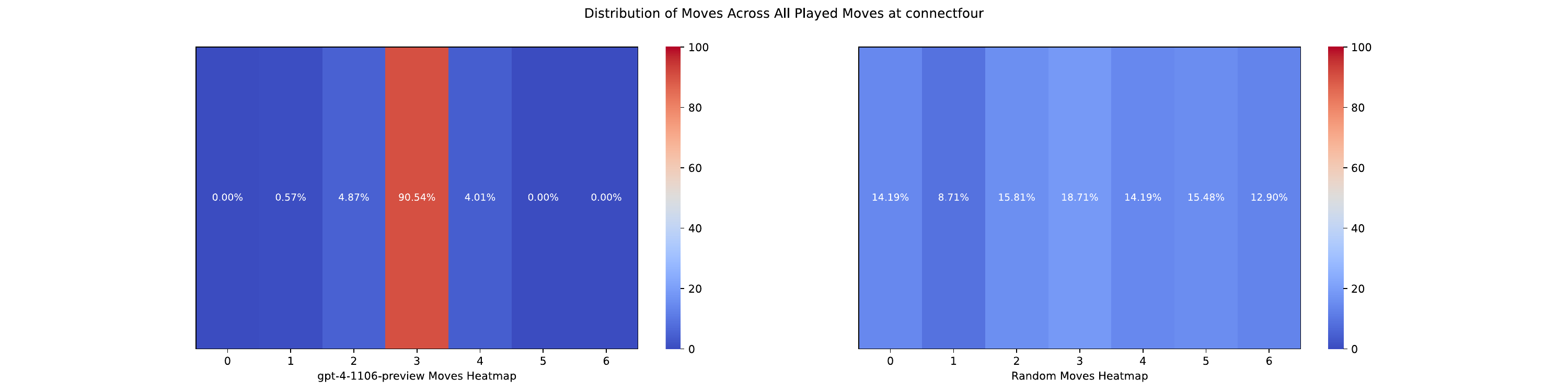}
        \caption{$t=1.5$.}
        \label{fig:connectfour_4_4}
    \end{subfigure}
    \caption{Heatmap of model GPT-4's moves for the connect-four game, where 't' is temperature.}
    \label{fig:connectfour_moves_4}
\end{figure}

\begin{figure}[H]
    \centering
    \begin{subfigure}[t]{1\linewidth}
        \centering
        \includegraphics[width=\linewidth]{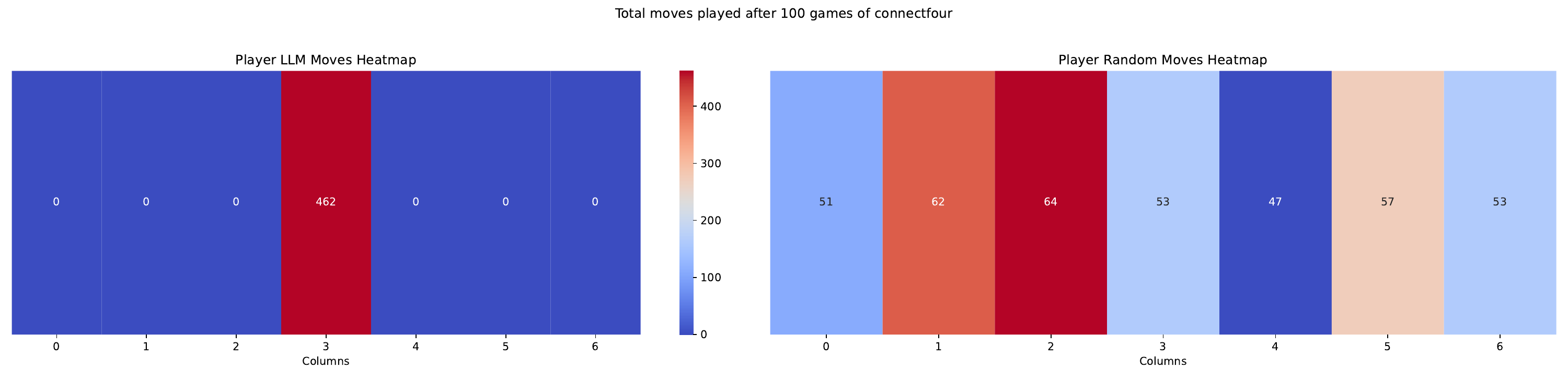}
        \caption{$t=0$.}
        \label{fig:connectfour_4o_mini_0}
    \end{subfigure}
    \hspace{10pt}
    \begin{subfigure}[t]{1\linewidth}
        \centering
        \includegraphics[width=\linewidth]{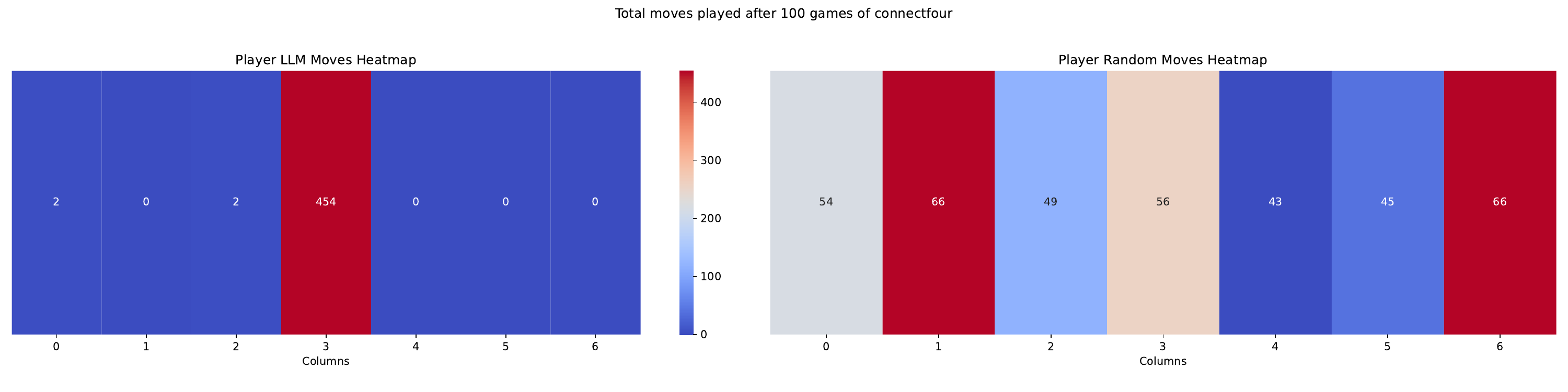}
        \caption{$t=0.5$.}
        \label{fig:connectfour_4o_mini_0_5}
    \end{subfigure}
    \\
    \begin{subfigure}[t]{1\linewidth}
        \centering
        \includegraphics[width=\linewidth]{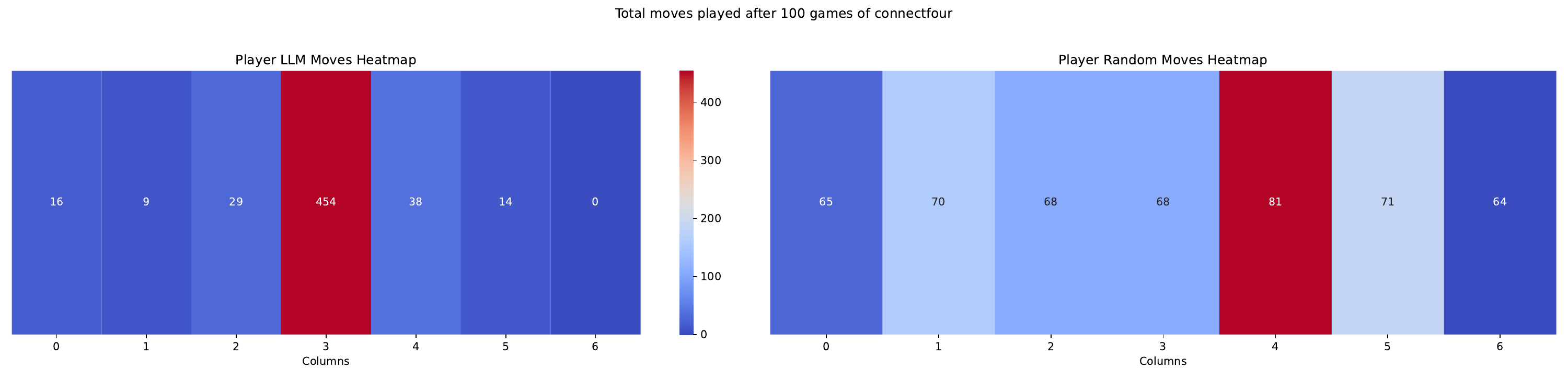}
        \caption{$t=1$.}
        \label{fig:connectfour_4o_mini_1}
    \end{subfigure}
    \hspace{10pt}
    \begin{subfigure}[t]{1\linewidth}
        \centering
        \includegraphics[width=\linewidth]{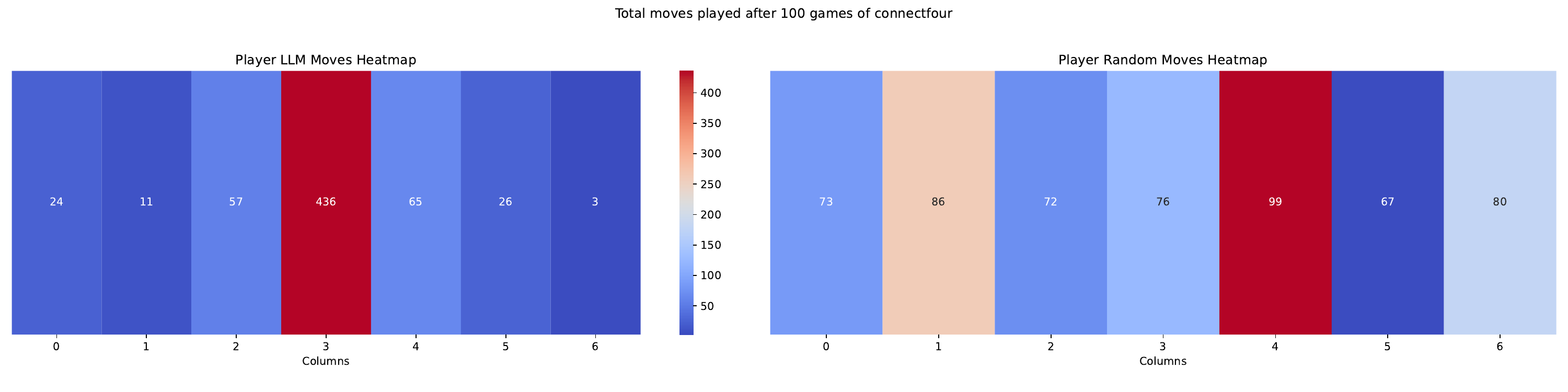}
        \caption{$t=1.5$.}
        \label{fig:connectfour_4o_mini_1_5}
    \end{subfigure}
    \caption{Heatmap of model GPT-4o-mini's moves for the Connect-Four game, where 't' is temperature.}
    \label{fig:connectfour_moves_4o_mini}
\end{figure}

\begin{figure}[H]
    \centering
    \begin{subfigure}[t]{1\linewidth}
        \centering
        \includegraphics[width=\linewidth]{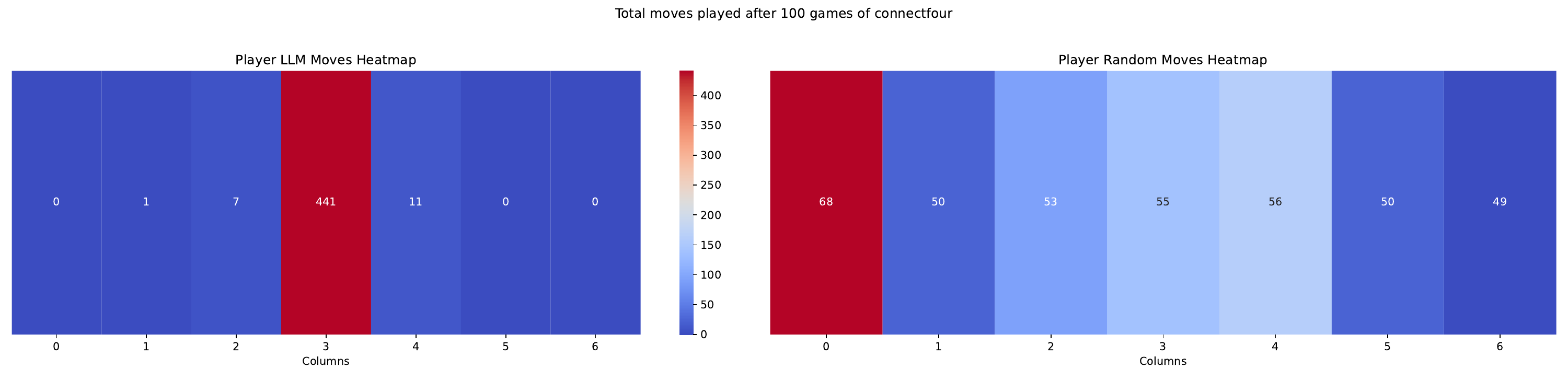}
        \caption{$t=0$.}
        \label{fig:connectfour_4o_0}
    \end{subfigure}
    \hspace{10pt}
    \begin{subfigure}[t]{1\linewidth}
        \centering
        \includegraphics[width=\linewidth]{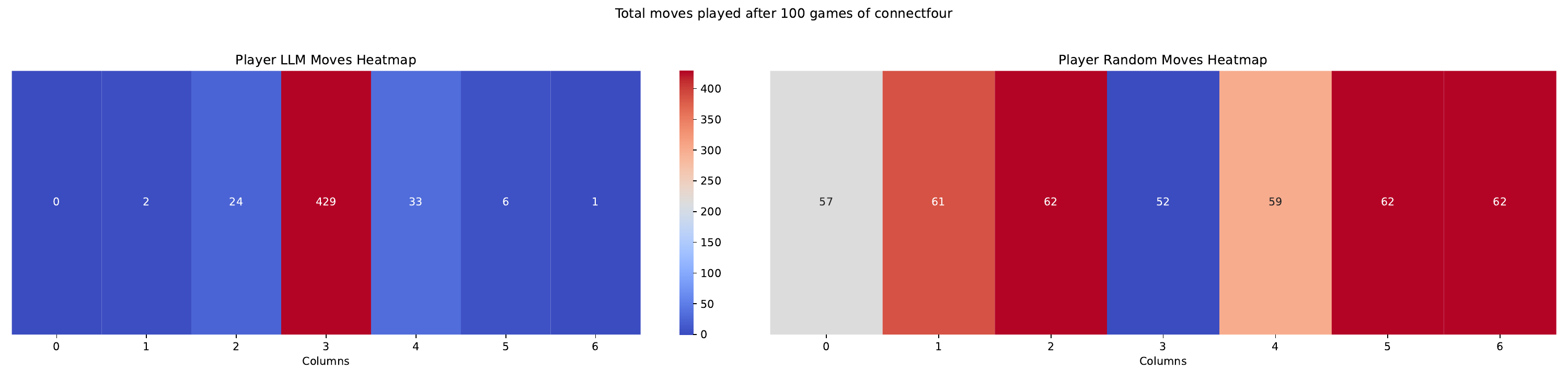}
        \caption{$t=0.5$.}
        \label{fig:connectfour_4o_0_5}
    \end{subfigure}
    \\
    \begin{subfigure}[t]{1\linewidth}
        \centering
        \includegraphics[width=\linewidth]{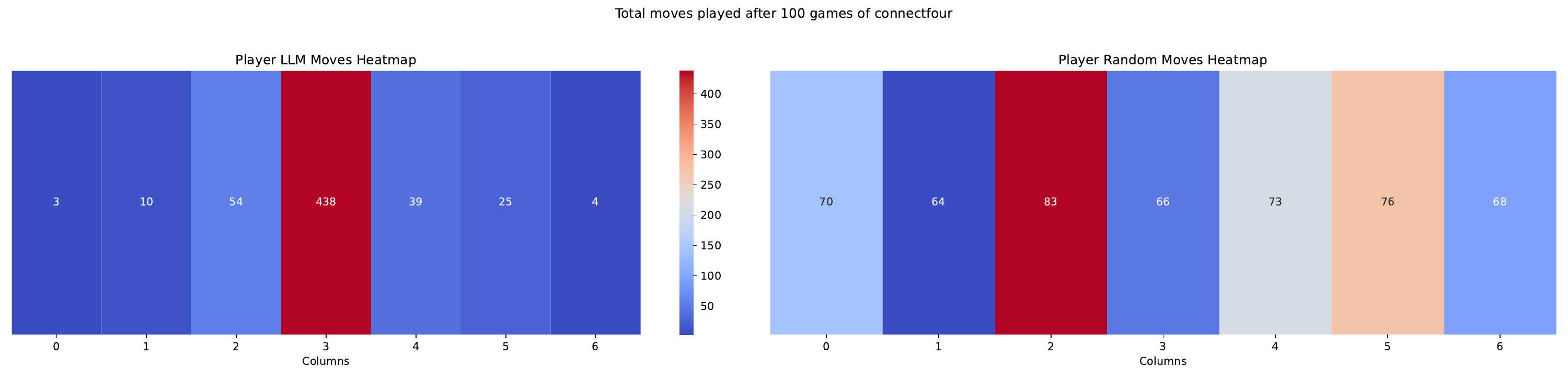}
        \caption{$t=1$.}
        \label{fig:connectfour_4o_1}
    \end{subfigure}
    \hspace{10pt}
    \begin{subfigure}[t]{1\linewidth}
        \centering
        \includegraphics[width=\linewidth]{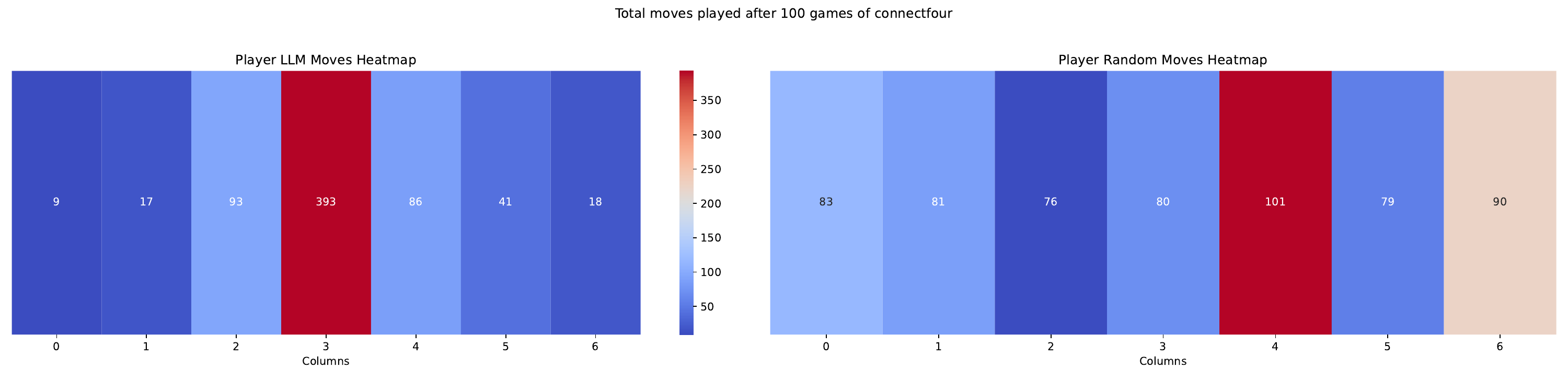}
        \caption{$t=1.5$.}
        \label{fig:connectfour_4o_1_5}
    \end{subfigure}
    \caption{Heatmap of model GPT-4o's moves for the Connect-Four game, where 't' is temperature.}
    \label{fig:connectfour_moves_4o}
\end{figure}

\newpage

\subsubsection{Battleship}\label{app:battleship_app}

Below, see the left column for the model's moves, and the right column for the random player's moves.

\begin{figure}[H]
    \centering
    \begin{subfigure}[t]{1\linewidth}
        \centering
        \includegraphics[width=\linewidth]{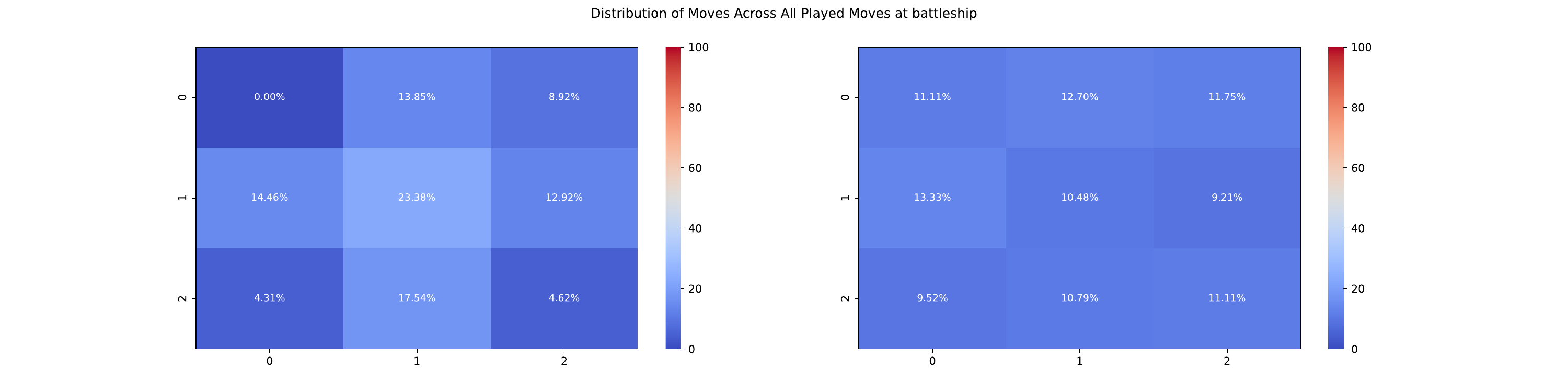}
        \caption{$t=0$.}
        \label{fig:battleship_1_3_5}
    \end{subfigure}
    \hspace{10pt} % Adds horizontal space between the subfigures
    \begin{subfigure}[t]{1\linewidth}
        \centering
        \includegraphics[width=\linewidth]{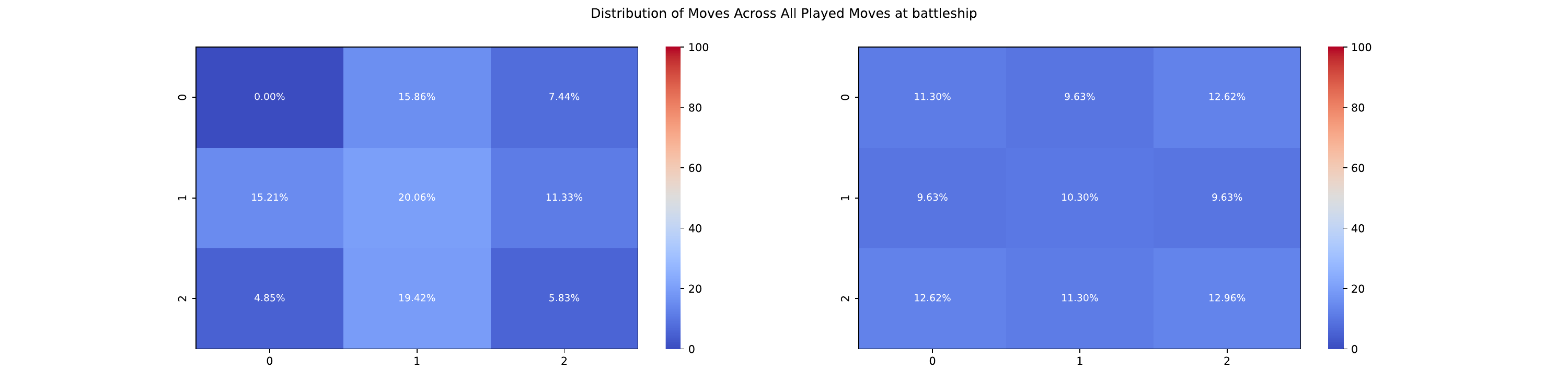}
        \caption{$t=0.5$.}
        \label{fig:battleship_2_3_5}
    \end{subfigure}
    \\ % Adds horizontal space between the subfigures
    \begin{subfigure}[t]{1\linewidth}
        \centering
        \includegraphics[width=\linewidth]{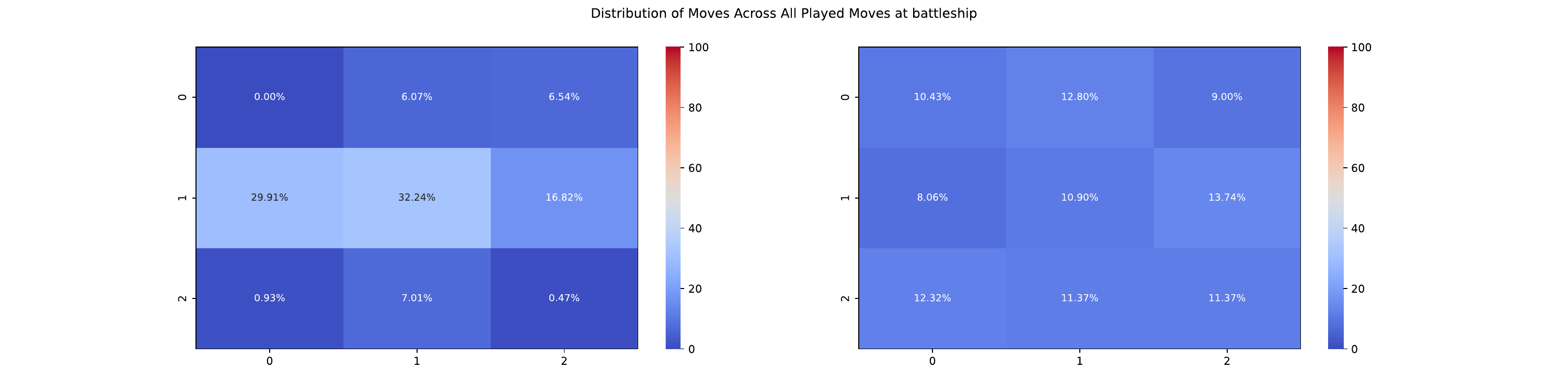}
        \caption{$t=1$.}
        \label{fig:battleship_3_3_5}
    \end{subfigure}
    \hspace{10pt} % Adds horizontal space between the subfigures
    \begin{subfigure}[t]{1\linewidth}
        \centering
        \includegraphics[width=\linewidth]{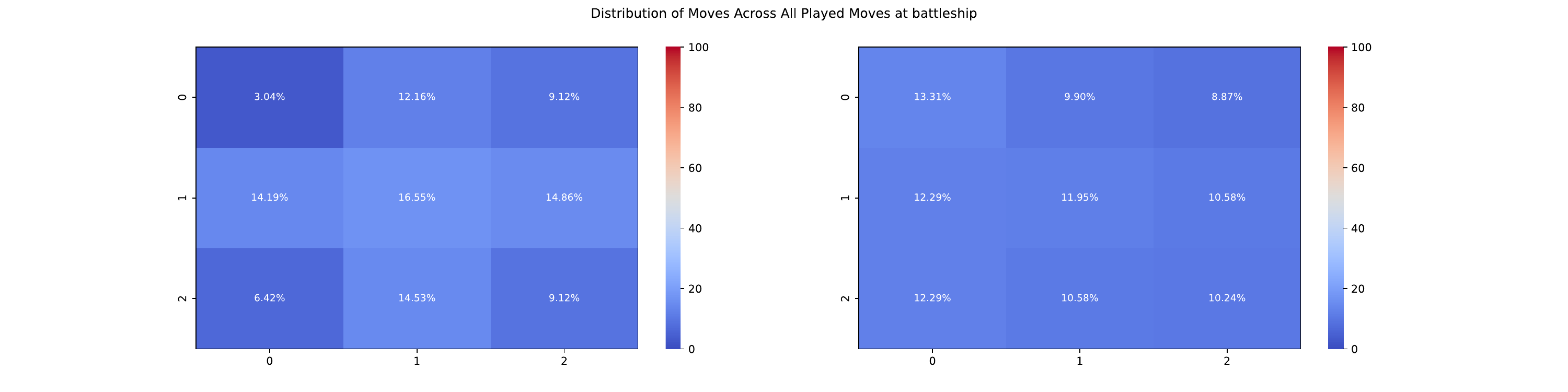}
        \caption{$t=1.5$.}
        \label{fig:battleship_4_3_5}
    \end{subfigure}
    \caption{Heatmap of model GPT-3.5's moves for the battleship game, where 't' is temperature.}
    \label{fig:battleship_moves_3_5}
\end{figure}

\begin{figure}[H]
    \centering
    \begin{subfigure}[t]{1\linewidth}
        \centering
        \includegraphics[width=\linewidth]{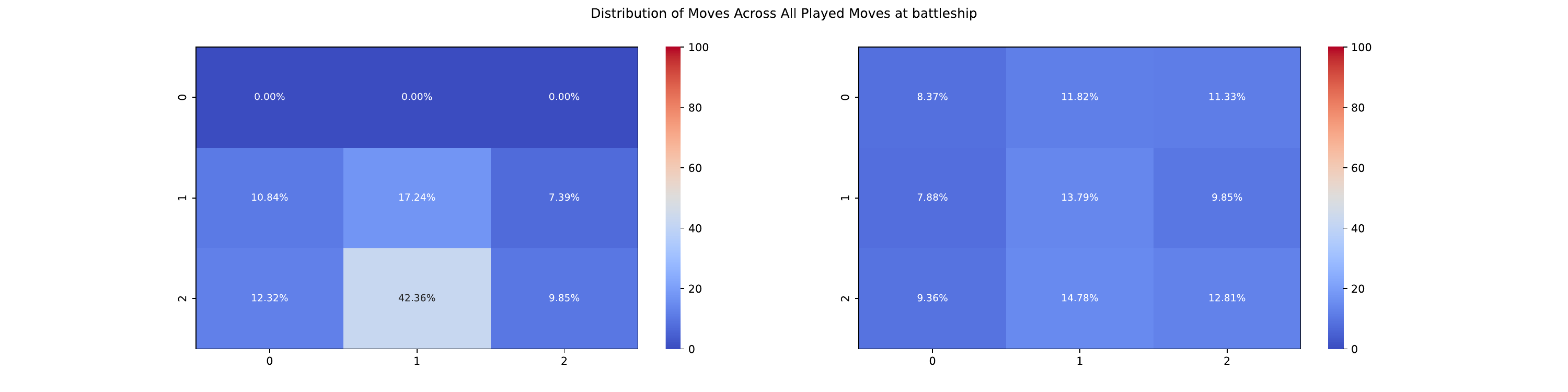}
        \caption{$t=0$.}
        \label{fig:battleship_1_4}
    \end{subfigure}
    \hspace{10pt} % Adds horizontal space between the subfigures
    \begin{subfigure}[t]{1\linewidth}
        \centering
        \includegraphics[width=\linewidth]{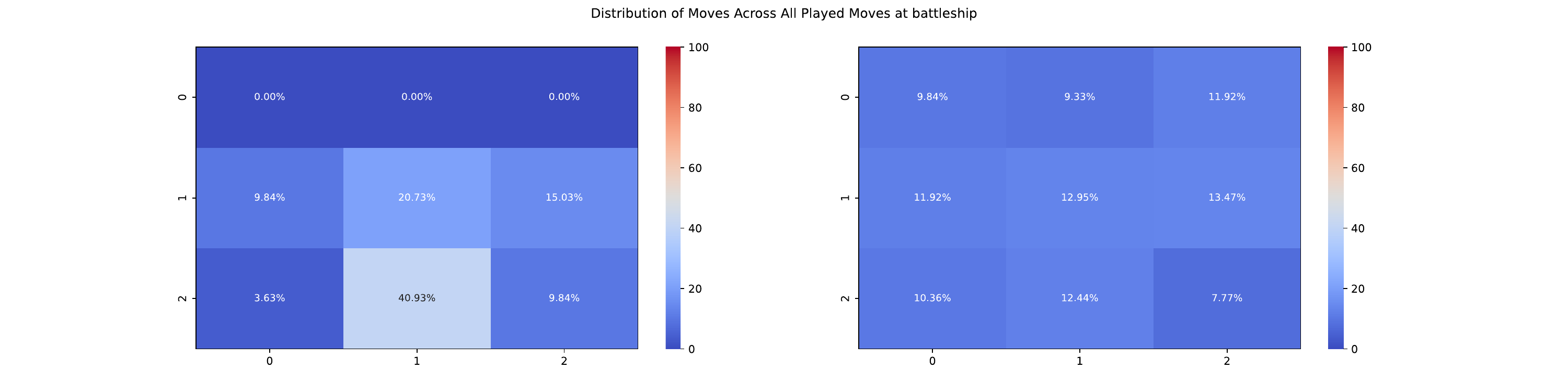}
        \caption{$t=0.5$.}
        \label{fig:battleship_2_4}
    \end{subfigure}
    \\ % Adds horizontal space between the subfigures
    \begin{subfigure}[t]{1\linewidth}
        \centering
        \includegraphics[width=\linewidth]{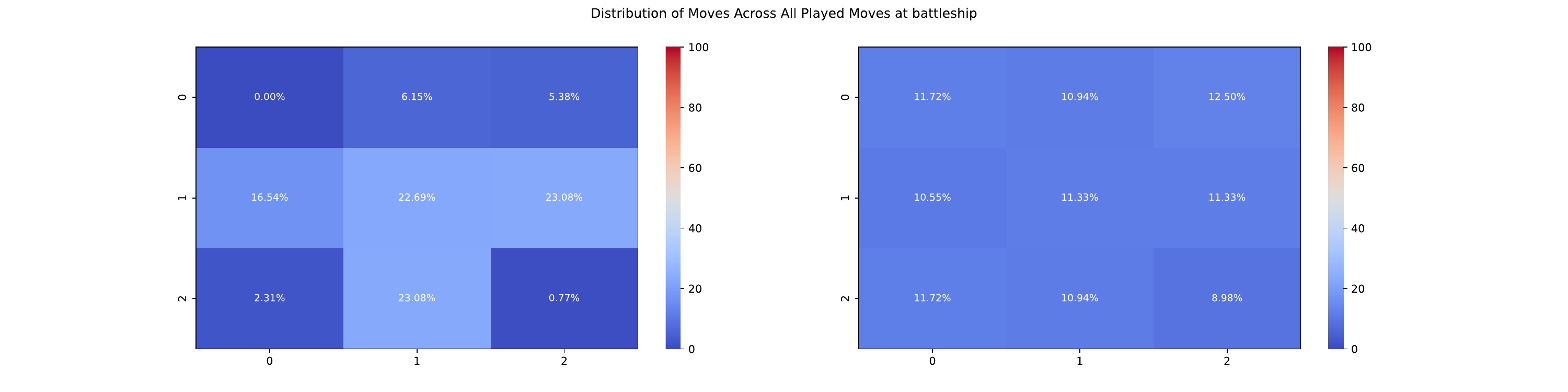}
        \caption{$t=1$.}
        \label{fig:battleship_3_4}
    \end{subfigure}
    \hspace{10pt} % Adds horizontal space between the subfigures
    \begin{subfigure}[t]{1\linewidth}
        \centering
        \includegraphics[width=\linewidth]{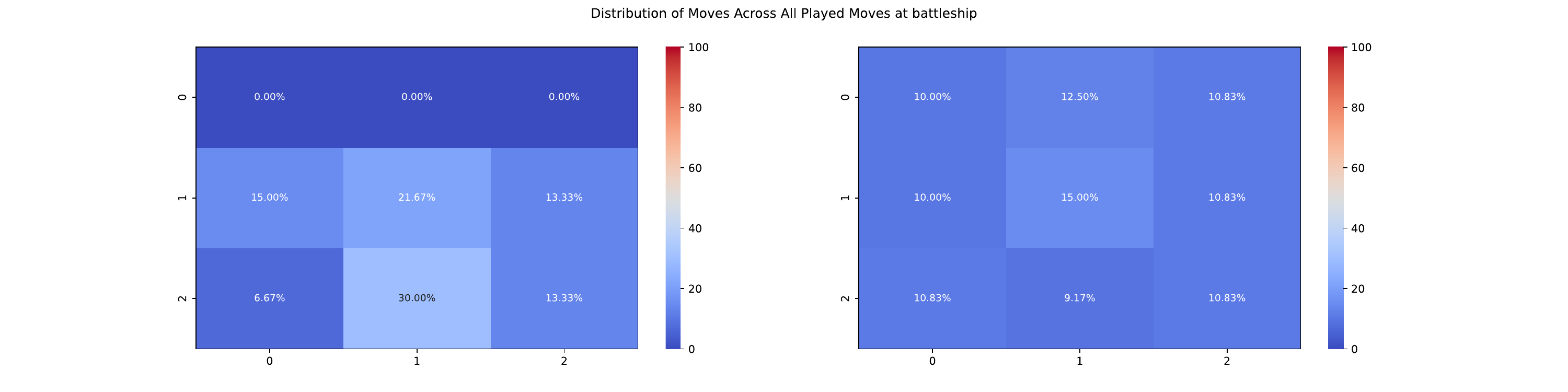}
        \caption{$t=1.5$.}
        \label{fig:battleship_4_4}
    \end{subfigure}
    \caption{Heatmap of model GPT-4's moves for the battleship game, where 't' is temperature.}
    \label{fig:battleship_moves_4}
\end{figure}

\begin{figure}[H]
    \centering
    \begin{subfigure}[t]{1\linewidth}
        \centering
        \includegraphics[width=\linewidth]{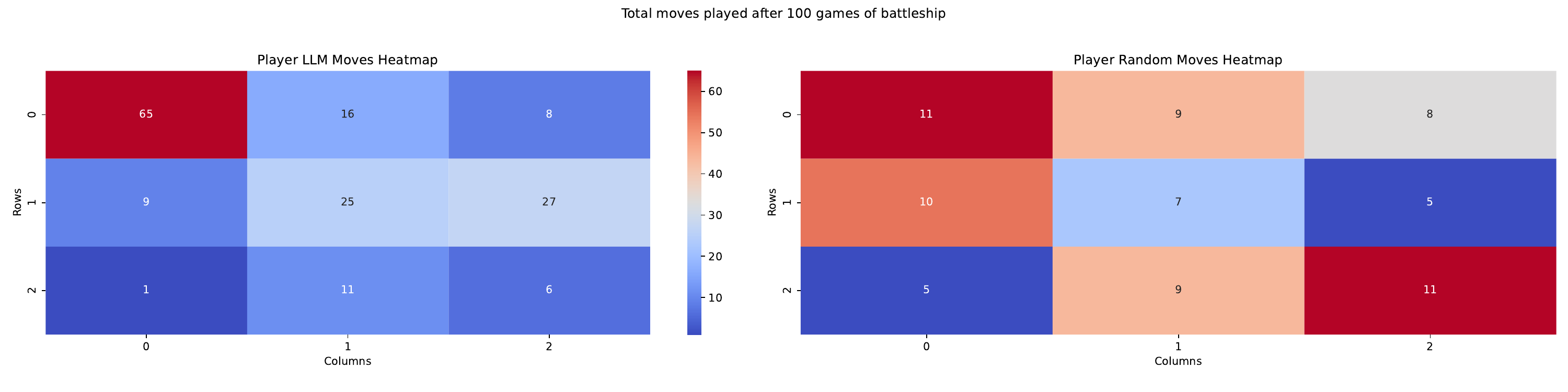}
        \caption{$t=0$.}
        \label{fig:battleship_4o_mini_0}
    \end{subfigure}
    \hspace{10pt}
    \begin{subfigure}[t]{1\linewidth}
        \centering
        \includegraphics[width=\linewidth]{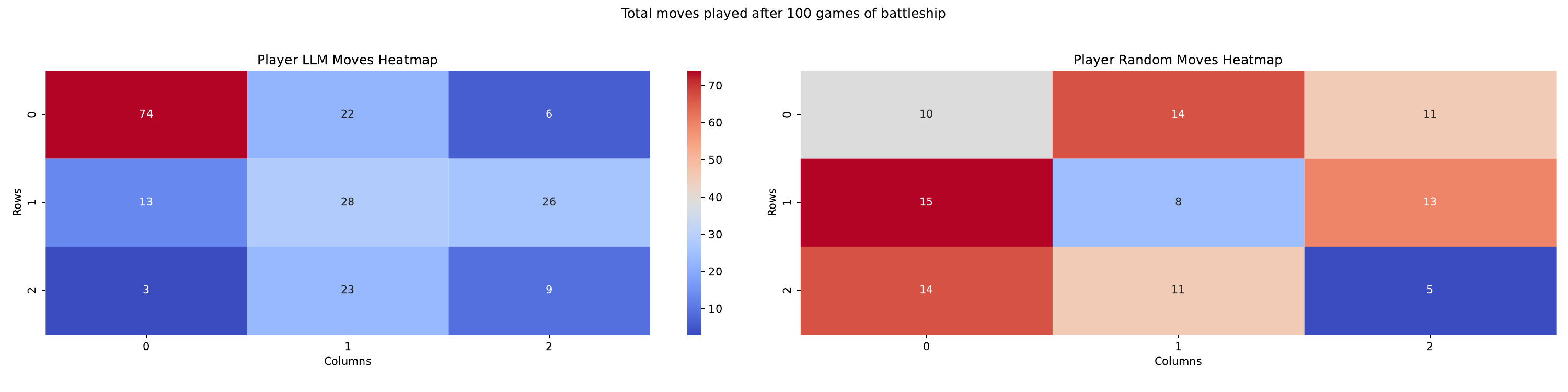}
        \caption{$t=0.5$.}
        \label{fig:battleship_4o_mini_0_5}
    \end{subfigure}
    \\
    \begin{subfigure}[t]{1\linewidth}
        \centering
        \includegraphics[width=\linewidth]{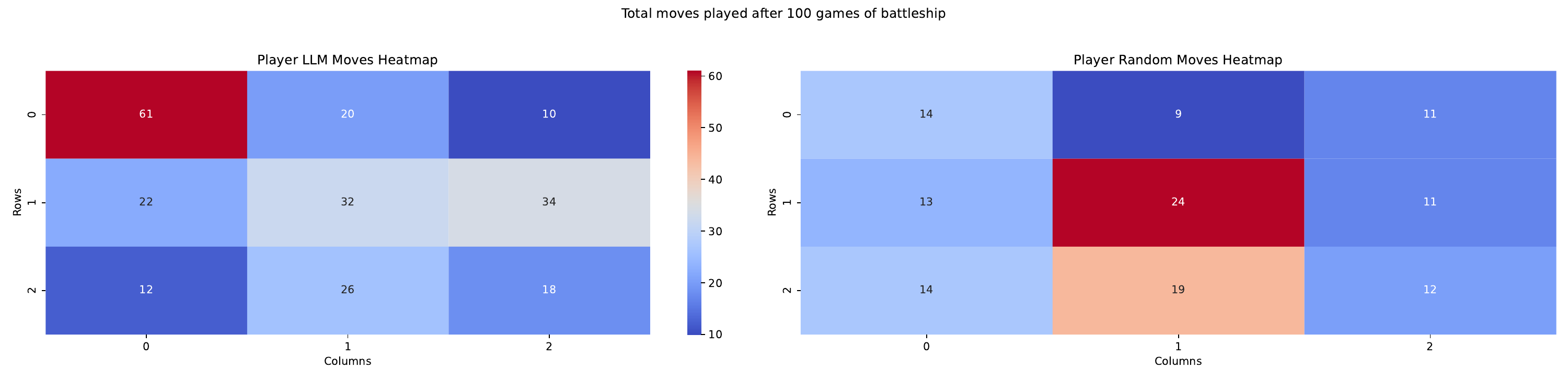}
        \caption{$t=1$.}
        \label{fig:battleship_4o_mini_1}
    \end{subfigure}
    \hspace{10pt}
    \begin{subfigure}[t]{1\linewidth}
        \centering
        \includegraphics[width=\linewidth]{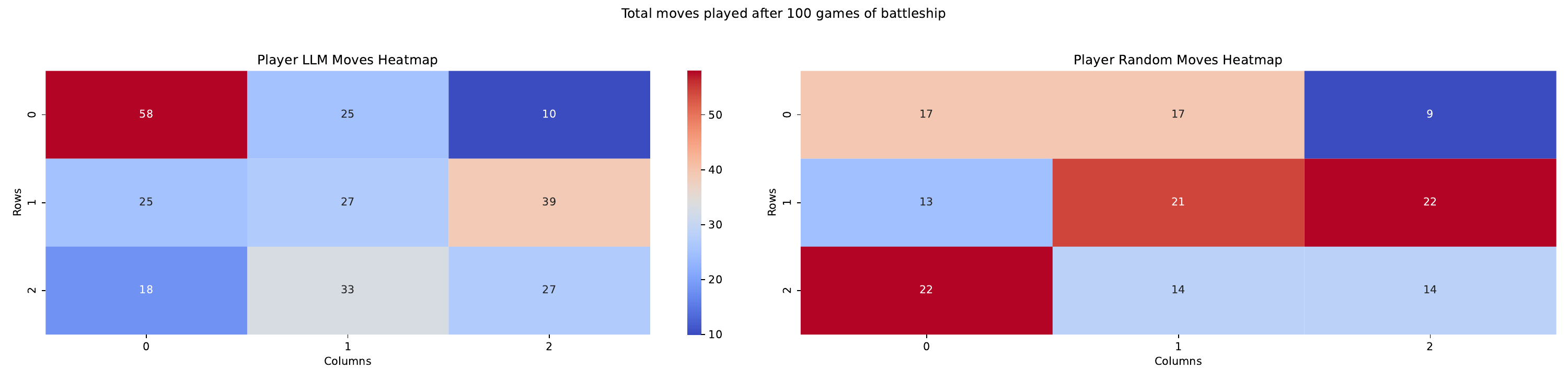}
        \caption{$t=1.5$.}
        \label{fig:battleship_4o_mini_1_5}
    \end{subfigure}
    \caption{Heatmap of model GPT-4o-mini's moves for the Battleship game, where 't' is temperature.}
    \label{fig:battleship_moves_4o_mini}
\end{figure}

\begin{figure}[H]
    \centering
    \begin{subfigure}[t]{1\linewidth}
        \centering
        \includegraphics[width=\linewidth]{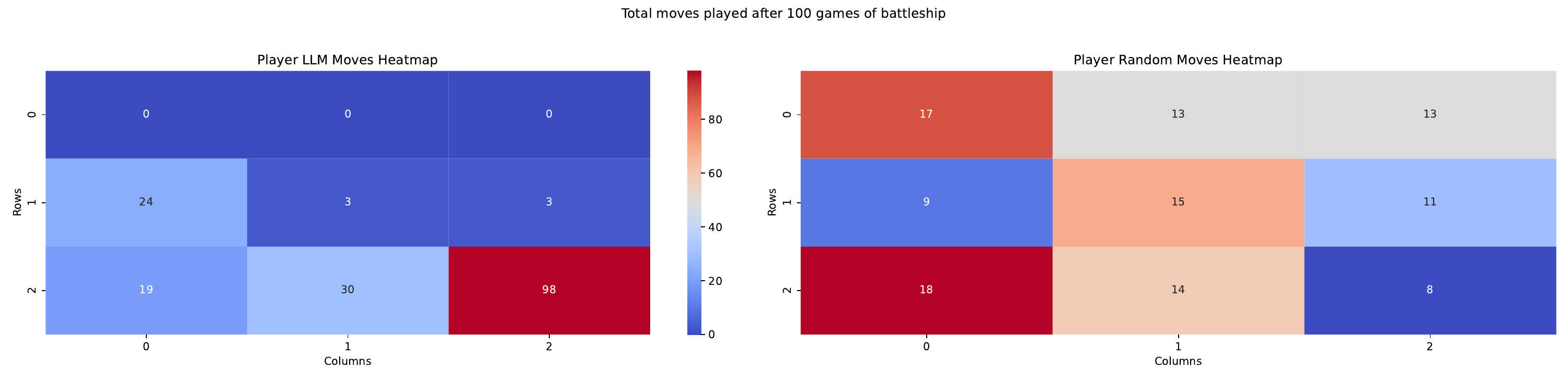}
        \caption{$t=0$.}
        \label{fig:battleship_4o_0}
    \end{subfigure}
    \hspace{10pt}
    \begin{subfigure}[t]{1\linewidth}
        \centering
        \includegraphics[width=\linewidth]{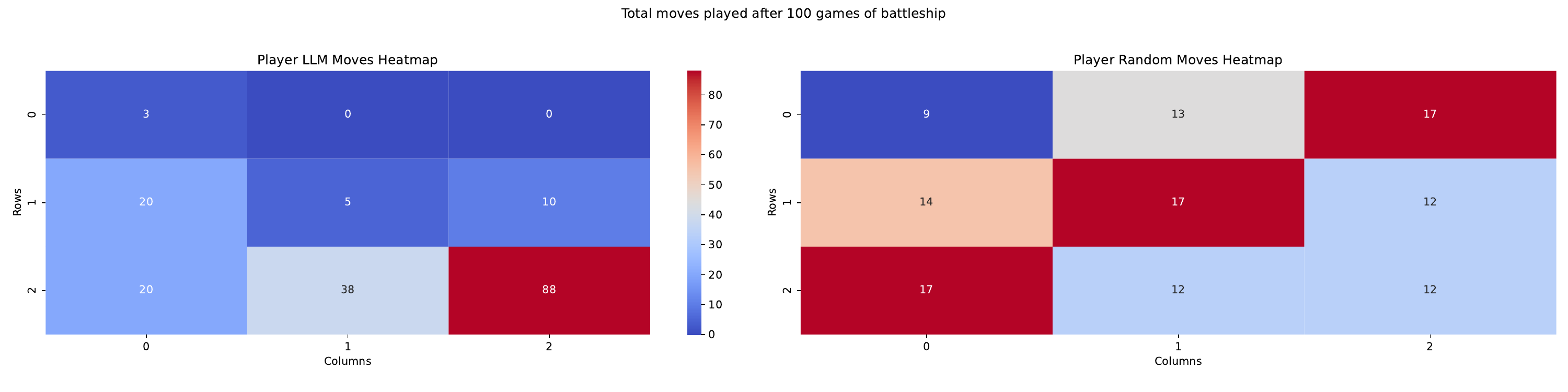}
        \caption{$t=0.5$.}
        \label{fig:battleship_4o_0_5}
    \end{subfigure}
    \\
    \begin{subfigure}[t]{1\linewidth}
        \centering
        \includegraphics[width=\linewidth]{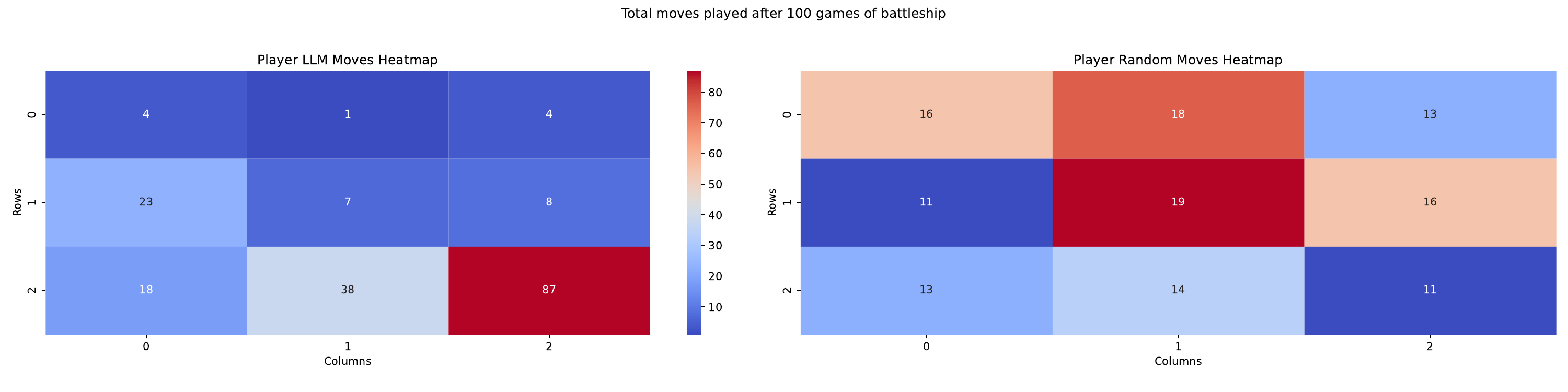}
        \caption{$t=1$.}
        \label{fig:battleship_4o_1}
    \end{subfigure}
    \hspace{10pt}
    \begin{subfigure}[t]{1\linewidth}
        \centering
        \includegraphics[width=\linewidth]{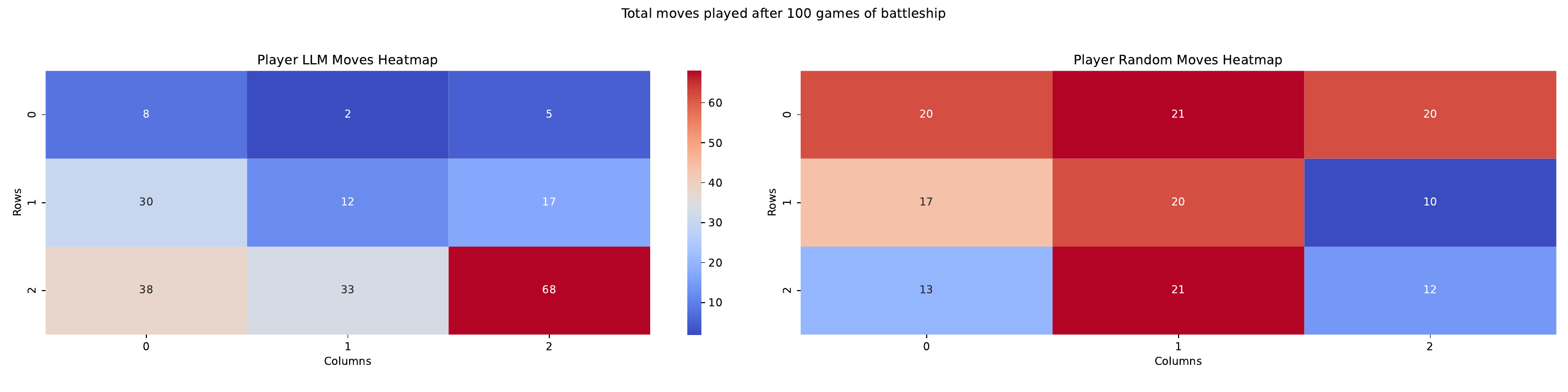}
        \caption{$t=1.5$.}
        \label{fig:battleship_4o_1_5}
    \end{subfigure}
    \caption{Heatmap of model GPT-4o's moves for the Battleship game, where 't' is temperature.}
    \label{fig:battleship_moves_4o}
\end{figure}

\newpage

\subsection{Optimal Play Prompting}\label{app:prompts_optimal}
\begin{table}[H]
\tiny
\centering
\begin{tabular}{|c|p{11.5cm}|}
\hline
\textbf{Game} & \textbf{Explanation} \\
\hline
Tic-Tac-Toe & Tic-Tac-Toe is a two-player game played on a 3x3 grid. Each player takes turns marking a square with their symbol (X or O), aiming to get three of their symbols in a row, column, or diagonal. To play optimally, prioritize securing the center square and blocking opponent's winning moves. \\
\hline
Battleship & Battleship is a two-player game where players hide ships on a grid and take turns guessing their opponent's ship locations. The goal is to sink all of the opponent's ships. To play optimally, start by targeting areas with higher probabilities of containing a ship and strategically target adjacent squares after a hit to maximize efficiency. \\
\hline
Connect Four & Connect Four is a two-player game played on a 6x7 grid. Players drop colored discs into columns, aiming to connect four of their own discs in a row, column, or diagonal. To play optimally, prioritize creating your own winning formations while blocking opponent's potential winning moves. \\
\hline
\end{tabular}
\caption{Optimal strategies for playing different games according to GPT-3.5.}
\label{tab:games_optimal_gpt3_5}
\end{table}

\begin{table}[H]
\tiny
\centering
\begin{tabular}{|c|p{11.5cm}|}
\hline
\textbf{Game} & \textbf{Explanation} \\
\hline
Tic-Tac-Toe & Play your first X in a corner to maximize opportunities. If the opponent plays in the center, play the opposite corner. Block your opponent's potential winning moves and always look to create a line of three. \\
\hline
Battleship & Randomize ship placements and start by targeting the center of the grid. Use a checkerboard pattern for efficient searching. Once a ship is hit, focus on the surrounding squares to determine its orientation and sink it. \\
\hline
Connect Four & Start in the center column to maximize opportunities in all directions. Build threats vertically, horizontally, and diagonally, and block the opponent's forming lines. Create multiple threats to force the opponent into a defensive position. \\
\hline
\end{tabular}
\caption{Optimal strategies for playing different games according to GPT-4.}
\label{tab:games_optimal_gpt4}
\end{table}

\begin{table}[H]
\tiny
\centering
\begin{tabular}{|c|p{11.5cm}|}
\hline
\textbf{Game} & \textbf{Explanation} \\
\hline
Tic-Tac-Toe & Tic-Tac-Toe is a two-player game played on a 3x3 grid. Each player takes turns marking a square with their symbol (X or O), aiming to get three of their symbols in a row, column, or diagonal. To play optimally, prioritize securing the center square, which maximizes the player's control over possible winning paths. In the absence of the center, playing in the corners is optimal. When a two-in-a-row pattern appears, immediately block your opponent. Building forks, or setups with two potential wins, forces the opponent into a defensive position, increasing chances of victory or draw if the game is played optimally. \\
\hline
Battleship & Battleship is a two-player game where players hide ships on a grid and take turns guessing their opponent's ship locations. The goal is to sink all of the opponent's ships by guessing correctly. For optimal play, begin by targeting squares in a checkerboard pattern to cover as much area with as few guesses as possible. Upon scoring a hit, concentrate shots on adjacent squares to locate and sink the entire ship. Targeting large open areas first can increase the probability of early hits and reduce the search space efficiently as ships are gradually located and sunk. \\
\hline
Connect Four & Connect Four is a two-player game played on a 6x7 grid. Players drop colored discs into columns, aiming to connect four of their own discs in a row, column, or diagonal. To play optimally, control the central columns, as they offer the most winning configurations. Building traps, where two potential winning moves are created simultaneously, can effectively force the opponent into a defensive stance. Playing ahead to block opponent formations while advancing your own connections maximizes control and improves winning opportunities, especially when creating two-way winning options in later moves. \\
\hline
\end{tabular}
\caption{Optimal strategies for playing different games according to GPT-4o.}
\label{tab:games_optimal_gpt4o}
\end{table}

%\label{app:prompts_optimal}
\begin{table}[H]
\tiny
\centering
\begin{tabular}{|c|p{11.5cm}|}
\hline
\textbf{Game} & \textbf{Explanation} \\
\hline
Tic-Tac-Toe & Tic-Tac-Toe is a two-player game played on a 3x3 grid. Each player takes turns marking a square with their symbol (X or O), aiming to get three of their symbols in a row, column, or diagonal. To play optimally, start by placing your symbol in the centre square if you go first. Always look for opportunities to create two potential winning paths while blocking your opponent from winning. If your opponent has two in a row, prioritize blocking them. The game will end in a draw if both players play optimally. \\
\hline
Battleship & Battleship is a two-player game where players hide ships on a grid and take turns guessing their opponent's ship locations. The goal is to sink all of the opponent's ships. To play optimally, start by placing your ships strategically to avoid clustering and target areas on the grid that have higher probabilities of containing ships. Use a systematic guessing strategy, such as targeting every other square initially, and narrow down your guesses based on hits. When you hit a ship, determine its orientation and focus your attacks on adjacent squares to sink it efficiently. \\
\hline
Connect Four & Connect Four is a two-player game played on a 6x7 grid. Players drop coloured discs into columns, aiming to connect four of their own discs in a row, column, or diagonal. To play optimally, always start by placing your disc in the centre column to maximize your winning potential. Create multiple threats by forming three in a row while blocking your opponent's attempts to connect four. Be vigilant about potential diagonal connections, and prioritize blocking any threats from your opponent immediately. Maintain a balance between offence and defence throughout the game. \\
\hline
\end{tabular}
\caption{Optimal strategies for playing different games according to GPT-4o-mini.}
\label{tab:games_optimal_gpt4o_mini}
\end{table}

\section{LCL}

\subsection{Prompts}
\begin{figure}[H]
\centering
\fbox{%
    \begin{minipage}{\dimexpr\textwidth-2\fboxsep-2\fboxrule\relax}
    \textbf{Validity Testing} prompt:\\
    \textit{"You will receive a description of a Lego structure, for instance, ((x1, y1, 'color1'), (x2, y2, 'color2')), which lists the coordinates and colors of two pieces. A construct is valid if all Lego pieces are connected but not overlapping. A Lego piece is connected through interlocking pegs, not by merely touching sides. Two Lego pieces overlap when they share the same y-coordinate and any part of their length has the same x-coordinate. If the following structure is valid then reply with valid, otherwise reply with invalid (do not justify your answer): <pieces>"}
     \end{minipage}}
     \caption{Validity testing prompt.}
     \label{fig:validity_prompt}
\end{figure}

\begin{figure}[H]
\centering
 \fbox{%
     \begin{minipage}{\dimexpr\textwidth-2\fboxsep-2\fboxrule\relax}
    \textbf{Construct Generation} prompt:\\
   \textit{ "A description of a Lego structure consists of a list of tuples, ((x1, y1, 'color1'), (x2, y2, 'color2')), where each tuple shows the coordinates and colors of a piece. Such a structure is valid if all Lego pieces are connected but not overlapping. A Lego piece is connected through interlocking pegs, not by merely touching sides. Two Lego pieces overlap when they share the same y-coordinate and any part of their length has the same x-coordinate. Produce a description of a valid structure using <n pieces> Lego pieces. Reply only with the Lego structure description following the format ((x1, y1, 'color1'), (x2, y2, 'color2'), ...), write nothing else but the structure."}
 \end{minipage}}
      \caption{Construct generation prompt.}
     \label{fig:generation_prompt}
\end{figure}

The prompts written in LaTeX from Figure~\ref{fig:validity_prompt} and Figure~\ref{fig:generation_prompt} were used both in the case of GPT-3.5 and GPT-4 in the main text. Notably, these tests are part of the ChildPlay suite. Further tests were conducted but not included in the ChildPlay suite and are illustrated herein. The reason why these tests have not been included in the suite is because they must be written as systematic benchmarks instead of experimental input-output segments. Currently, they stand as illustrative cases of spatial reasoning failure and success that supplement the benchmark but are not aimed at proving the model's capacity either way. They are simply an interesting addition.

%\section{LCL Syntax}
    
    \subsection{Definitions}\label{app:def_lcl}
    
    A piece $P$ is defined as a tuple $P = (l, w, (x, y), c, h)$ (see Table \ref{table:parameter_lcl})  where:
    \begin{enumerate}
    \item $l$ is the length of the piece, fixed at 4 units;
    \item $w$ is the width of the piece, fixed at 2 units;
    \item $x-axis$ corresponds to the position of the studs;
    \item $y-axis$ corresponds to layers - the first brick is at layer 0;
    \item $c$ is the color of the piece;
    \item $h$ is the height of the piece, fixed at 1 unit;
    \end{enumerate}
    
    For the sake of brevity, in most of the examples below we omit length ($l$), color ($c$), and height ($h$) since these are set as constants.

    \begin{table}[H]
  \centering
  \begin{tabular}{lll}
    \toprule
    Parameter & Description & Value \\
    \midrule
    \( l \) & Length of the piece & 4 units \\
    \( w \) & Width of the piece & 2 units \\
    \( (x, y) \) & Position of the studs (x-axis), layers (y-axis) & Var \\
    \( c \) & Colour of the piece & Var \\
    \( h \) & Height of the piece & 1 unit \\
    \bottomrule
  \end{tabular}
  \caption{Definition of a Piece \( P \)}
  \label{table:parameter_lcl}
\end{table}
    
    A construction, $M$, is then a valid construction in $LCL_2$ if and only if it follows the rules:
    \begin{enumerate}
    \item $P = (4, 2, (x, y), c, 1)$
    \item $M$ is composed entirely by $P$ pieces ($\Phi = {P}$);
    \item Every piece P must be connected to at least one other piece P;
    \item $M$ is symmetric along the line crossing the 2 by 4 pieces, between its pegs, along the piece's longest side;
    \item Pieces in the construct can only be manipulated horizontally in $n * pi$ rotations, with $n \in \mathbb{Z}$ (note that this makes width irrelevant);
    \item The position of a piece is defined by its left-most pair of studs;
    \item $M$ begins with a piece P at coordinates (0,0);
    \item All pieces placed in layer $n$ must be placed before any piece is placed in layer $n + 1$;
    \end{enumerate}

\begin{figure}[H]
    \centering
    \fbox{%
        \includegraphics[width=0.5\linewidth]{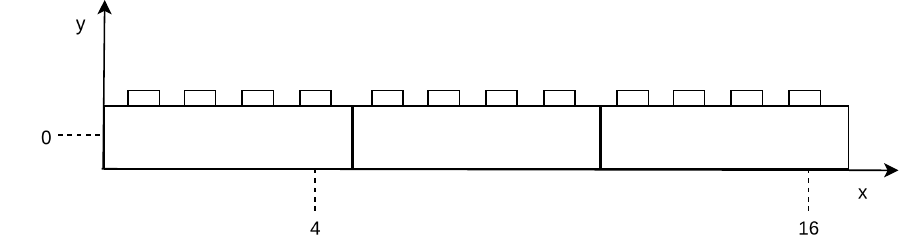}
    }
    \caption{A disconnected line of bricks is not a valid construct in $LCL_2$: \{$(0, 0), (4, 0), (8, 0)$\}.}
    \label{fig:fig_1}
\end{figure}

    Consider constructing a line using three bricks (we omit height $h$ since it is a constant, with value equal to 1). This is counter-intuitive, but note that a line cannot be represented as in Fig~\ref{fig:fig_1}, because the pieces are disconnected.
    
    $LCL_2$: $((0, 0), (4, 0), (8, 0))$ is then an example of what one expects to see as representing a line, but it is not valid in LCL. Because the pieces are disconnected from eachother, they just lay next to eachother, one after another in a row. Instead, $((0, 0), (4, 0), (2, 1))$, or $((0, 0), (-2, 1), (2, 1))$, or even $((0, 0), (-2, 1), (4, 1))$ would be valid constructs. 

Subsequently, both models were prompted with several additional requests that have not been integrated in the suite yet (see Table \ref{table:lcl_prompts}). 

For these experiments, the definition of LCL was provided to the model and it was accompanied by the prompt in Figure~\ref{fig:prompt_more}. 
\begin{figure}[H]
\centering
    \fbox{%
    \begin{minipage}{\dimexpr\textwidth-2\fboxsep-2\fboxrule\relax}

    \textbf{Prompt:} \textit{"I will give you a number of pieces, I will ask you for a shape and you'll output the coordinates per piece to form such a shape. It must be valid in LCL."}
 \end{minipage}}
      \caption{Extra testing prompts not in the suite.}
     \label{fig:prompt_more}
\end{figure}
 
\begin{table}[H]
  \centering
  \begin{tabular}{p{0.2\textwidth} p{0.75\textwidth}}
    \toprule
    Task & Description \\
    \midrule
    Triangle Construction & \textit{"Make a triangle with 5 bricks."} \\
    \midrule
    Humanoid Figure & \textit{"6 pieces. Build a humanoid figure."}\\
    \midrule
    Bart Simpson-Like Figure & \textit{"Let me help you. Imagine it's Bart Simpson. You have three yellow pieces, one for the head, two for the arms, one red for the torso, and two blue pieces for the legs."} \\
    \midrule
    Tower Construction & \textit{"Produce now a tower with 3 bricks."} \\
    \bottomrule
  \end{tabular}
  \caption{Sequence of building prompts.}
  \label{table:lcl_prompts}
\end{table}

\subsection{Construct Example}

\begin{figure}[H]
    \centering
    \fbox{\includegraphics[width=0.35\linewidth]{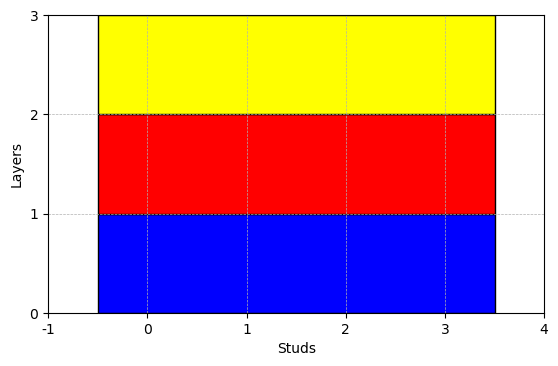}}
    \caption{A \textit{valid} tower representation in $LCL_2$.}
    \label{fig:tower}
\end{figure}

A simple example is found in Fig \ref{fig:tower}. This is a tower constructed from 3 bricks and is a valid $LCL_2$ construct.

This sequence forms the construction of a 3-brick line, each brick having a width of 4 units. But since this construction is composed of three columns, one piece $P$ each, it can be broken apart and is not a topological object (each piece can be moved individually). The \textbf{correct} construct with three bricks has many possible solutions. For a centre piece with two pieces on the bottom or two pieces on the top, we find $24$ possible solutions. In eq.~\ref{eq:1} is the general formula with $s$ being the amount of studs:

\begin{equation}
    \begin{split}
        f(0)=0 \\
        f(s)=4*(s-1)+f(s-1)
    \end{split}
\label{eq:1}
\end{equation}

And its non-recursive form:

\begin{equation}
    \begin{split}
        f(0)=0 \\
        f(s) = 2(s - 1)s
    \end{split}
\label{eq:2}
\end{equation}

We show two more simple examples:

\begin{figure}[H]
    \centering
    \fbox{%
        \includegraphics[width=0.5\linewidth]{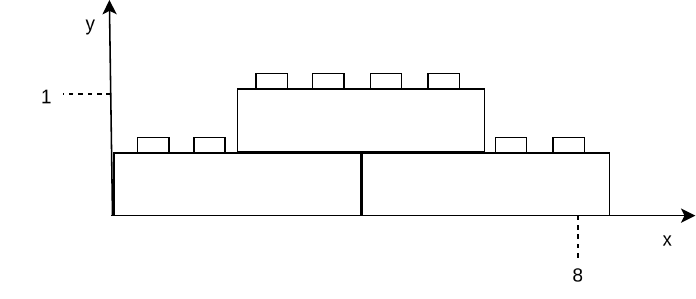}
    }
    \caption{A possible representation of the requested "line" as a valid construct in $LCL_2$: \{$(0, 0), (4, 0), (2, 1)$\}.}
    \label{fig:fig_2}
\end{figure}

, and:

\begin{figure}[H]
    \centering
    \fbox{%
        \includegraphics[width=0.5\linewidth]{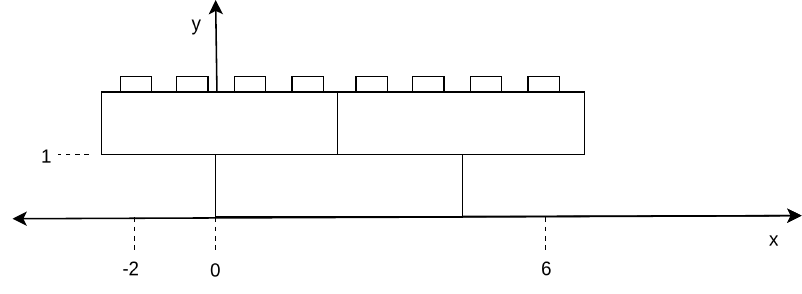}
    }
    \caption{Another possible representation of the requested "line" as a valid construct in $LCL_2$: \{$(0, 0), (-2, 1), (2, 1)$\}.}
    \label{fig:fig_3}
\end{figure}

The "three-in-a-line" can only be loosely interpreted in $LCL_2$, due to rule (2) - that pieces cannot be moved independently from the rest of the model. For this reason, one can imagine many more structures that loosely fall under the definition of a "line" or "wall", for example: 

\begin{figure}[H]
    \centering
    \fbox{%
        \includegraphics[width=0.5\linewidth]{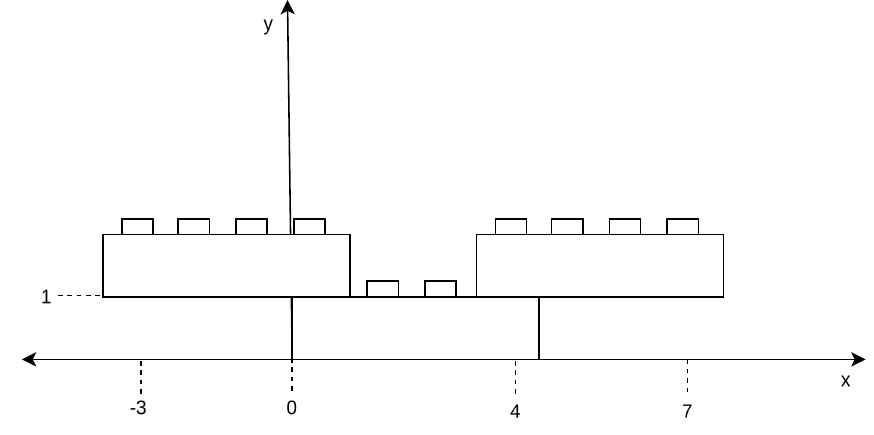}
    }
    \caption{Another possible construct for the requested "line". This is also a valid construct in $LCL_2$: \{$(0, 0), (-2, 1), (4, 1)$\}.}
    \label{fig:fig_4}
\end{figure}

Or even a stair-like structure:

\begin{figure}[ht]
    \centering
    \fbox{%
        \includegraphics[width=0.5\linewidth]{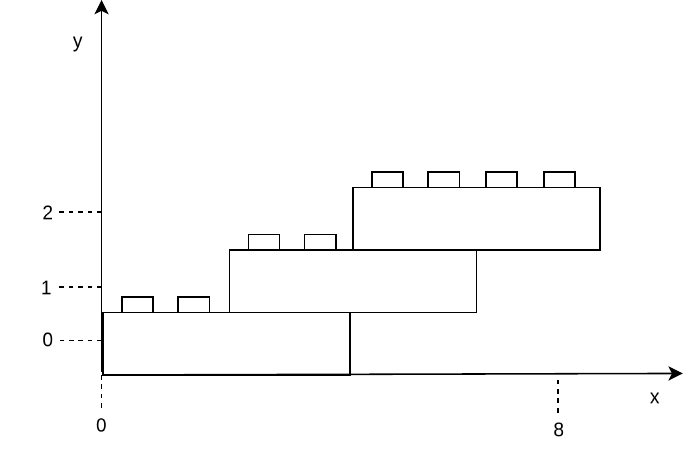}
    }
    \caption{Stair-like construct for the requested "line". This is also a valid construct in $LCL_2$: \{$(0, 0), (2, 1), (4, 2)$\}.}
    \label{fig:fig_5}
\end{figure}

A humanoid could also be easily represented in $LCL_2$ as:

\begin{figure}[h]
    \centering
    \fbox{%
        \includegraphics[width=0.5\linewidth]{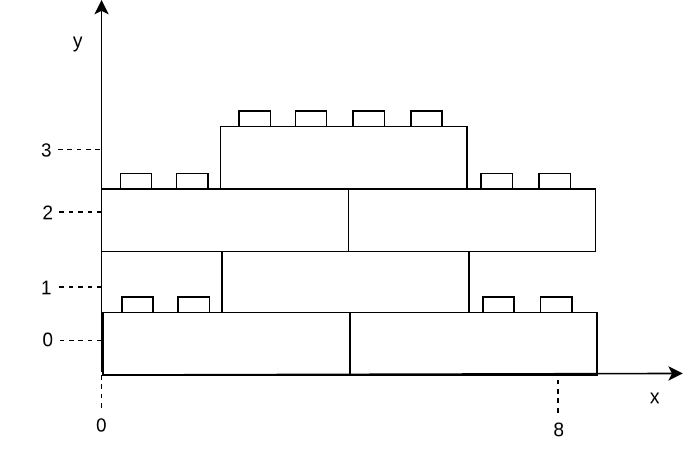}
    }
    \caption{A possible representation of a humanoid as a valid construct in $LCL_2$: \{$(0, 0), (4, 0), (2, 1), (0, 2), (4, 2), (2, 3)$\}.}
    \label{fig:fig_6}
\end{figure}

% \section{Decoding and Verification}
% Developing a decoder involves evaluating all possible arrangements to ensure the structure represents a valid construct of the intended object. A simplified version could check instructions step-by-step, comparing each with the provided key instruction. However, this may not detect alternative valid representations. Therefore, we propose a program that translates LCL instructions into a 2D representation, allowing visual evaluation. As it stands, it only encodes the typical 2 by 4 brick, but it can be enhanced to include all pieces.

% \begin{lstlisting}[language=Python, label={script}]

% \end{lstlisting}

\subsubsection{Soft Experiments}\label{app:soft_lcl}
\begin{figure}[H]
    \centering
    \begin{figure}[H]
        \centering
        \begin{subfigure}[t]{0.36\linewidth}
            \centering
            \fbox{\includegraphics[width=\linewidth]{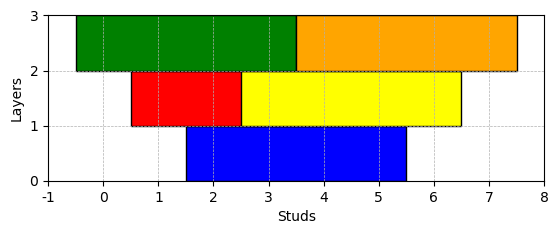}}
            \caption{GPT-3.5.}
            \label{fig:triangle_1}
        \end{subfigure}%
        \hspace{10pt} % <-- Add horizontal space between subfigures
        \begin{subfigure}[t]{0.468\linewidth}
            \centering
            \fbox{\includegraphics[width=\linewidth]{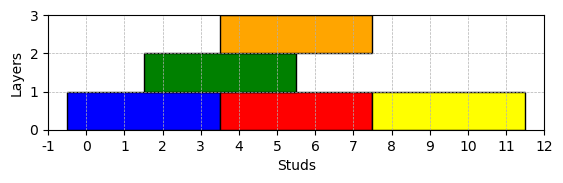}}
            \caption{GPT-4.}
            \label{fig:triangle_2}
        \end{subfigure}
        % \caption{Illustrations in $LCL_2$.}
        \caption{Model responses to the query: \textit{"Make a triangle with 5 bricks."}, randomised colours. This is impossible to satisfy.}
        \label{fig:triangle}
    \end{figure}

    \begin{figure}[H]
        \centering
        \begin{subfigure}[t]{0.45\linewidth}
            \centering
            \fbox{\includegraphics[width=\linewidth]{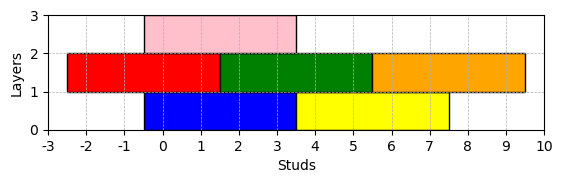}}
            \caption{GPT-3.5.}
            \label{fig:triangle_3}
        \end{subfigure}%
        \hspace{10pt} % <-- Add horizontal space between subfigures
        \begin{subfigure}[t]{0.45\linewidth}
            \centering
            \fbox{\includegraphics[width=\linewidth]{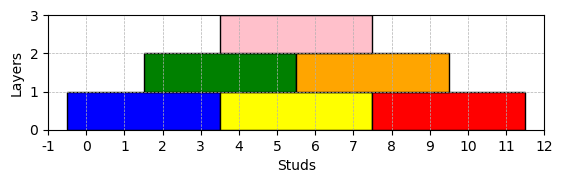}}
            \caption{GPT-4.}
            \label{fig:triangle_4}
        \end{subfigure}
        \caption{Model responses to the query: \textit{"Make a triangle with 6 bricks."}, randomised colours. }
        \label{fig:correct_triangle}
    \end{figure}

    \begin{figure}[H]
        \centering
        \begin{subfigure}[t]{0.35\linewidth}
            \centering
            \fbox{\includegraphics[width=\linewidth]{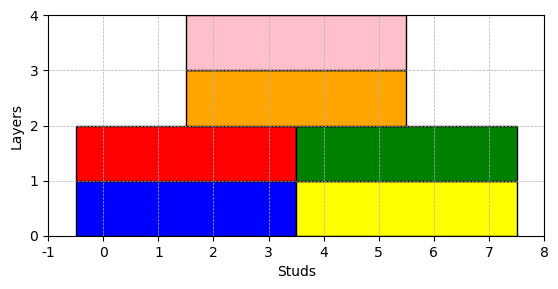}}
            \caption{GPT-3.5}
            \label{fig:humanoid_1}
        \end{subfigure}%
        \hspace{10pt} % <-- Add horizontal space between subfigures
        \begin{subfigure}[t]{0.36\linewidth}
            \centering
            \fbox{\includegraphics[width=\linewidth]{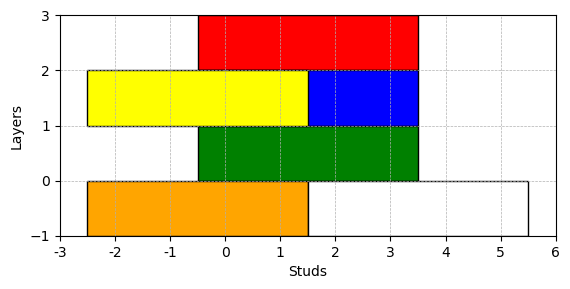}}
            \caption{GPT-4}
            \label{fig:humanoid_2}
        \end{subfigure}
        % \caption{Illustrations in $LCL_2$.}
        \caption{Model responses to the query: \textit{"You have 6 pieces. Build a humanoid figure."}, randomised colours.}
        \label{fig:humanoid}
    \end{figure}

    \begin{figure}[H]
        \centering
        \begin{subfigure}[t]{0.4\linewidth}
            \centering
            \fbox{\includegraphics[width=\linewidth]{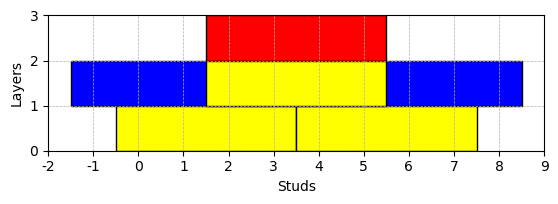}}
            \caption{GPT-3.5}
            \label{fig:bart_simpson_1}
        \end{subfigure}%
        \hspace{10pt} % <-- Add horizontal space between subfigures
        \begin{subfigure}[t]{0.346\linewidth}
            \centering
            \fbox{\includegraphics[width=\linewidth]{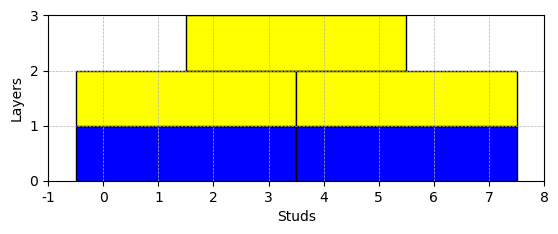}}
            \caption{GPT-4}
            \label{fig:bart_simpson_2}
        \end{subfigure}
        % \caption{Illustrations in $LCL_2$.}
        \caption{Model response to the query: \textit{"Imagine it's bart simpson. You have three yellow pieces, one for the head, two for the arms, one red for the torso, and two blue pieces for the legs."} - colours specified.}
        \label{fig:bart_simpson}
    \end{figure}
\end{figure}

We show the model's incorrect answers in Figures~\ref{fig:triangle}, \ref{fig:humanoid}, and \ref{fig:bart_simpson} and correct answers in Figures~\ref{fig:triangle_4}. Essentially, both GPT-3.5 and GPT-4 were not far from the expected target, but failed to respect $LCL_2$ rules in most cases. For example, pieces are found in an impossible superposition in Figure~\ref{fig:triangle_1} (red piece is in the same position as yellow piece), \ref{fig:humanoid_2} (blue piece is in the same position as yellow piece), and \ref{fig:bart_simpson_2} (red piece is in the same position as middle yellow pieces). In Figure~\ref{fig:bart_simpson_1}, GPT-3.5 erroneously swapped the middle yellow piece with the red piece and the blue pieces with the bottom yellow pieces, even though it first declared in plain English the correct organisation of the 6 pieces. The positive result is that models manage to assemble a tower of three pieces and GPT-4 was capable of assembling a triangle (see Table \ref{table:lcl_model_comparison}). None of the models recognised that they were asked an impossible task, namely building a triangle with only 5 pieces (see Figure~\ref{fig:triangle}).

\begin{table}[H]
% \tiny
  \centering
  \begin{tabular}{lllll}
    \toprule
    \multicolumn{2}{c}{Model} & \multicolumn{2}{c}{Responses} \\
    \cmidrule(r){1-2} \cmidrule(l){2-4}
    Category & N(P) & GPT-3.5 & GPT-4\\
    \midrule
    Tower    & 3 & Correct   & Correct   \\
    Impossible Triangle & 5 & Incorrect & Incorrect \\
    Triangle & 6 & Incorrect & Correct \\
    Humanoid & 6 & Incorrect & Incorrect  \\
    Bart Simpson  & 6 & Incorrect & Incorrect  \\
    \bottomrule
  \end{tabular}
  \caption{Comparison of Responses by GPT-3.5 and GPT-4.}
  \label{table:lcl_model_comparison}
\end{table}

\subsection{Example Constructs}\label{app:4o-4o-mini-lcl}
\textbf{GPT-4o-mini and GPT-4o}\label{app:4o-mini-lcl} \quad 
\begin{figure}[H]
    \centering
    \begin{subfigure}[t]{0.45\linewidth}
        \centering
        \includegraphics[width=0.6\linewidth]{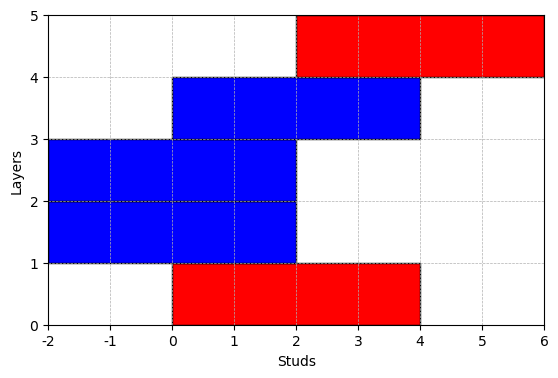}
        \caption{GPT-4o-mini, $t = 0$, valid output.}
        \label{fig:4o_mini_valid_lcl_0}
    \end{subfigure}
    \hfill
    \begin{subfigure}[t]{0.45\linewidth}
        \centering
        \includegraphics[width=0.6\linewidth]{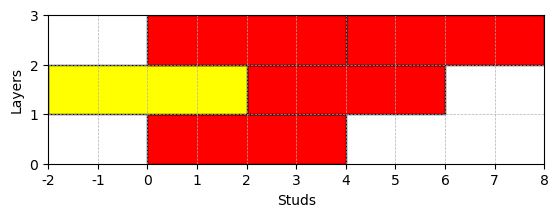}
        \caption{GPT-4o-mini, $t = 0.5$, valid output.}
        \label{fig:4o_mini_valid_lcl_0_5}
    \end{subfigure}

    \vspace{0.5cm} % Space between rows

    \begin{subfigure}[t]{0.45\linewidth}
        \centering
        \includegraphics[width=0.6\linewidth]{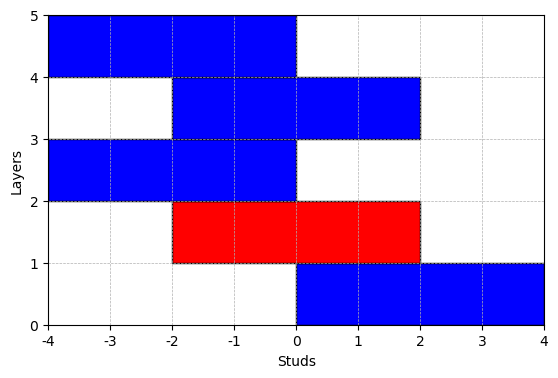}
        \caption{GPT-4o, $t = 0$, valid output.}
        \label{fig:4o_valid_lcl_0}
    \end{subfigure}
    \hfill
    \begin{subfigure}[t]{0.45\linewidth}
        \centering
        \includegraphics[width=0.6\linewidth]{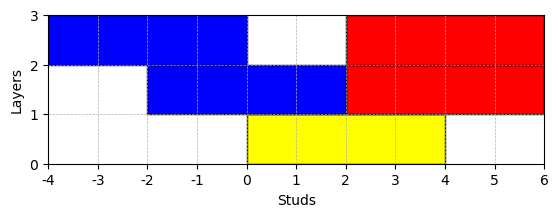}
        \caption{GPT-4o, $t = 0.5$, valid output.}
        \label{fig:4o_valid_lcl_0_5}
    \end{subfigure}
\end{figure}

\subsection{Game Outcomes}\label{app:lcl_game_outcomes}
\begin{figure}[H]
    \centering
    \begin{subfigure}[t]{0.32\linewidth}
        \centering
        \includegraphics[width=0.8\linewidth]{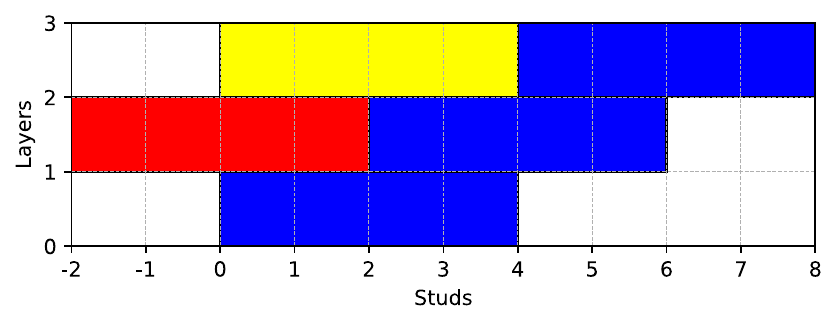}
        \caption{Valid LCL assembly.}
        \label{fig:valid_lcl_1}
    \end{subfigure}
    %\hspace{0.05\linewidth}
    \begin{subfigure}[t]{0.32\linewidth}
        \centering
        \includegraphics[width=0.8\linewidth]{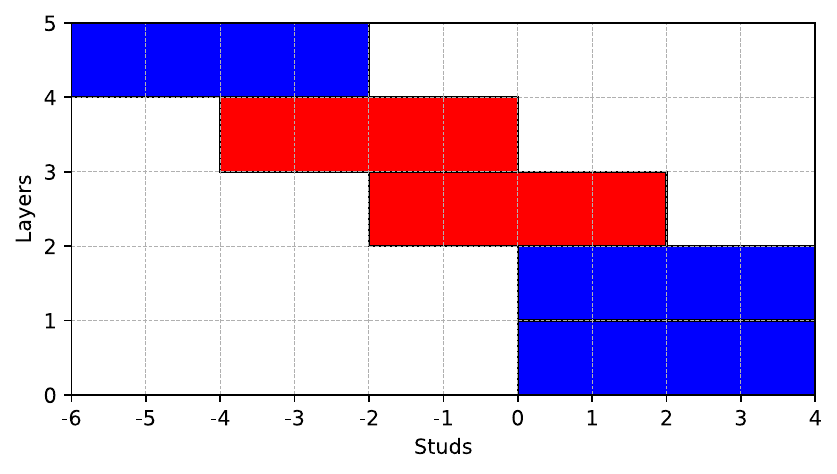}
        \caption{Valid LCL assembly.}
        \label{fig:valid_lcl_2}
    \end{subfigure}
    %\hspace{0.05\linewidth}
    \begin{subfigure}[t]{0.32\linewidth}
        \centering
        \includegraphics[width=0.8\linewidth]{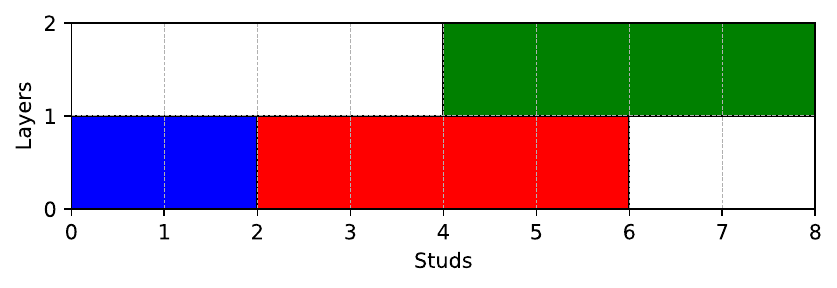}
        \caption{Invalid LCL structure.}
        \label{fig:invalid_lcl_2}
    \end{subfigure}
    
    \vspace{10pt} % Space between the rows

    \begin{subfigure}[t]{0.32\linewidth}
        \centering
        \includegraphics[width=0.8\linewidth]{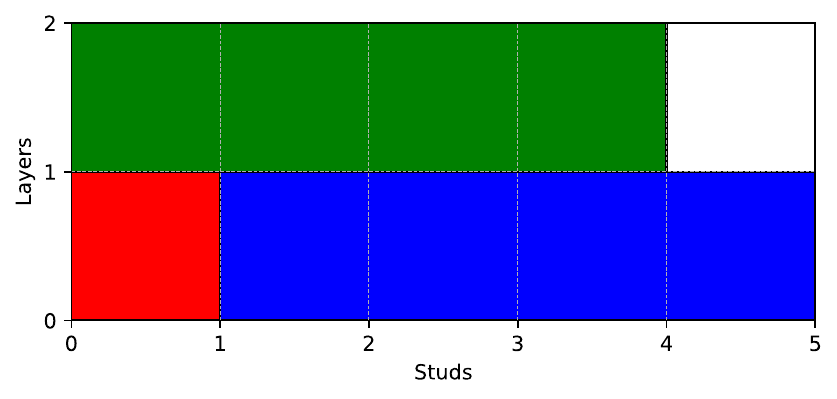}
        \caption{GPT-3.5, $t=0$, invalid output.}
        \label{fig:gpt3.5_lcl_temp_0}
    \end{subfigure}
    %\hspace{0.05\linewidth}
    \begin{subfigure}[t]{0.32\linewidth}
        \centering
        \includegraphics[width=0.8\linewidth]{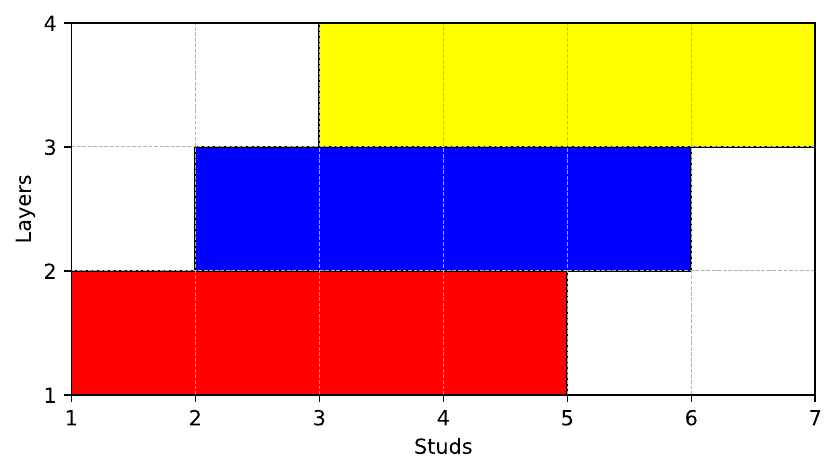}
        \caption{GPT-4, $t=1.5$, valid output.}
        \label{fig:gpt4_lcl_temp_1_5_53}
    \end{subfigure}
    %\hspace{0.05\linewidth}
    \begin{subfigure}[t]{0.32\linewidth}
        \centering
        \includegraphics[width=0.8\linewidth]{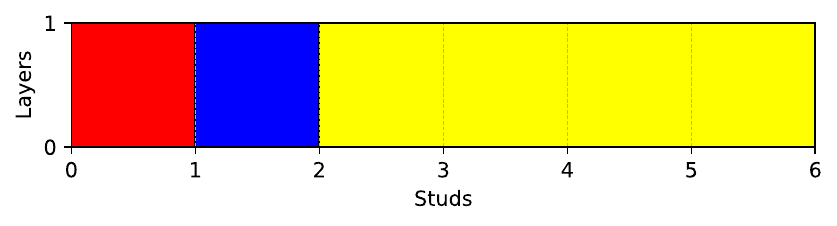}
        \caption{GPT-3.5, $t=1$, invalid output.}
        \label{fig:gpt3.5_lcl_temp_1}
    \end{subfigure}
    
    \caption{Example LCL validity test structures and constructs from GPT-3.5 and GPT-4.}
    \label{fig:lcl_all}
\end{figure}

For validity tests, Figures~\ref{fig:valid_lcl_1} and \ref{fig:valid_lcl_2} show valid LCL, Figure~\ref{fig:invalid_lcl_2} shows an invalid LCL. These were used in their encoding (cf. Section~\ref{app:def_lcl}) as input. 

Figures~\ref{fig:gpt3.5_lcl_temp_0} and \ref{fig:gpt3.5_lcl_temp_1} show invalid constructs from GPT-3.5 and Figure~\ref{fig:gpt4_lcl_temp_1_5_53} shows the only type of valid output from GPT-4. We also observe an increase in correct answers for GPT-3.5 in the validity test against temperature. GPT-4 peaks at t=0.5. Yet, correct responses are minimal. 

\subsection{Advanced LEGO Kits}\label{app:lego_kits}
The dataset defined herein contains several example prompts that are more complex and do not follow the 2x4 assumption, each consisting of a request followed by a LEGO kit of fewer than 15 pieces to which the agent is bound.

\subsubsection*{Apple}
\textbf{Possible prompt:} \textit{"Construct a LEGO apple with a mix of red and green colors, resembling a typical apple shape using slopes and bricks."}
\begin{itemize}
    \item Green Slope 45 2 x 1 - Code: 3040 (Quantity: 1)
    \item Red Slope 45 2 x 2 - Code: 3039 (Quantity: 2)
    \item Lime Slope, Inverted 45 2 x 2 - Code: 3660 (Quantity: 2)
    \item Red Brick 2 x 3 - Code: 3002 (Quantity: 1)
    \item Lime Plate 2 x 2 - Code: 3022 (Quantity: 1)
    \item Lime Brick 1 x 2 - Code: 3004 (Quantity: 1)
\end{itemize}

\subsubsection*{Yellow Hut}

\textbf{Possible prompt:} \textit{"Build a hut with a purple and yellow color scheme, featuring a simple structure and a sloped roof."}
\begin{itemize}
    \item Trans-Clear Brick 1 x 2 without Bottom Tube - Code: 3065 (Quantity: 2)
    \item Medium Nougat Brick 2 x 2 - Code: 3003 (Quantity: 1)
    \item Lime Plate 2 x 6 - Code: 3795 (Quantity: 1)
    \item Bright Light Yellow Brick 1 x 2 - Code: 3004 (Quantity: 4)
    \item Bright Light Yellow Brick 2 x 2 - Code: 3003 (Quantity: 1)
    \item Medium Lavender Slope 45 2 x 2 - Code: 3039 (Quantity: 4)
\end{itemize}

\subsubsection*{Fortress}
\textbf{Possible prompt:} \textit{"Create a medieval-themed LEGO fortress with arches, walls, and defensive structures, symbolizing a stronghold."}
\begin{itemize}
    \item Green Plate 2 x 8 - Code: 3034 (Quantity: 1)
    \item Light Bluish Gray Arch 1 x 4 x 2 - Code: 6182 (Quantity: 2)
    \item Sand Green Brick 1 x 2 - Code: 3004 (Quantity: 2)
    \item Light Bluish Gray Brick 1 x 2 - Code: 3004 (Quantity: 2)
    \item Dark Bluish Gray Brick 1 x 2 - Code: 3004 (Quantity: 2)
    \item Light Bluish Gray Brick 2 x 2 - Code: 3003 (Quantity: 1)
    \item Reddish Brown Brick, Round 1 x 1 Open Stud - Code: 3062b (Quantity: 2)
\end{itemize}

\subsubsection*{Dinghy}
\textbf{Possible prompt:} \textit{"Assemble a small LEGO dinghy with a white sail and a mast."}
\begin{itemize}
    \item Dark Tan Plate 2 x 4 - Code: 3020 (Quantity: 1)
    \item Tan Slope, Inverted 33 3 x 2 with Flat Bottom Pin and Connections - Code: 3747b (Quantity: 1)
    \item White Slope 45 2 x 2 - Code: 3039 (Quantity: 3)
    \item White Brick 2 x 2 - Code: 3003 (Quantity: 1)
    \item White Brick 1 x 2 - Code: 3004 (Quantity: 1)
    \item Tan Brick 2 x 3 - Code: 3002 (Quantity: 1)
    \item Reddish Brown Brick, Round 2 x 2 with Axle Hole - Code: 3941 (Quantity: 1)
\end{itemize}

\subsubsection*{Blue Bot}
\textbf{Possible prompt:} \textit{"Construct a LEGO robot with a humanoid structure, featuring a distinguishable head, body, arms, and legs."}
\begin{itemize}
    \item Medium Blue Brick 2 x 2 - Code: 3003 (Quantity: 1)
    \item Brick, Modified 2 x 3 with Curved Top - Code: 6215 (Quantity: 1)
    \item Brick 2 x 4 - Code: 3001 (Quantity: 1)
    \item Brick 1 x 2 - Code: 3004 (Quantity: 2)
    \item Brick, Round 2 x 2 with Grille - Code: 92947 (Quantity: 1)
    \item Plate 2 x 2 - Code: 3022 (Quantity: 1)
    \item Brick, Modified 1 x 2 with Studs on 1 Side - Code: 11211 (Quantity: 1)
    \item Brick 1 x 2 without Bottom Tube - Code: 3065 (Quantity: 1)
    \item Tile 1 x 1 Round - Code: 98138 (Quantity: 1)
    \item Brick, Round 2 x 2 Dome Top, with Bottom Axle Holder - Code: 553c (Quantity: 1)
\end{itemize}

\subsubsection*{Toy Car}
\textbf{Possible prompt:} \textit{"Build a LEGO toy car with a compact design, featuring wheels, and a sloped windshield."}
\begin{itemize}
    \item Brick 2 x 6 - Code: 2456 (Quantity: 1)
    \item Slope 2 x 2 45° - Code: 3039 (Quantity: 1)
    \item Brick 1 x 2 without Bottom Tube - Code: 3065 (Quantity: 1)
    \item Brick 1 x 2 - Code: 3004 (Quantity: 1)
    \item Plate 2 x 2 with Wheel Holders - Code: 4600 (Quantity: 2)
    \item Wheel 8mm D. x 6mm with Slot - Code: 34337 (Quantity: 4)
    \item Tire Offset Tread Small - Band Around Center of Tread - Code: 87414 (Quantity: 4)
\end{itemize}

\subsubsection*{Goldfish}
\textbf{Possible prompt:} \textit{"Create a LEGO goldfish with fins and tail, featuring elements for eyes."}
\begin{itemize}
    \item Brick 2 x 4 - Code: 3001 (Quantity: 2)
    \item Brick 1 x 2 with Pin Hole - Code: 3700 (Quantity: 1)
    \item Brick, Modified 1 x 2 with Studs on 1 Side - Code: 11211 (Quantity: 2)
    \item Brick 2 x 3 - Code: 3002 (Quantity: 1)
    \item Slope 45° 2 x 2 - Inverted - Code: 3660 (Quantity: 1)
    \item Slope 2 x 1 - 45° - Code: 3040 (Quantity: 4)
    \item Tile 1 x 1 Round with Eye Pattern - Code: 98138pb007 (Quantity: 2)
    \item Slope 30° 1 x 2 x 2/3 - Code: 85984 (Quantity: 1)
\end{itemize}

\subsubsection*{Baby Elephant}
\textbf{Possible prompt:} \textit{"Assemble a LEGO baby elephant with a focus on its trunk, ears, and body structure."}
\begin{itemize}
    \item Brick 2 x 6 - Code: 2456 (Quantity: 1)
    \item Brick 1 x 2 - Code: 3004 (Quantity: 3)
    \item Brick 1 x 4 - Code: 3010 (Quantity: 1)
    \item Brick 1 x 1 with Stud on 1 Side - Code: 87087 (Quantity: 2)
    \item Tile 1 x 1 Round with Eye Pattern - Code: 98138pb027 (Quantity: 2)
    \item Brick 2 x 4 - Code: 3001 (Quantity: 1)
\end{itemize}

\subsubsection*{Flamingo}
\textbf{Possible prompt:} \textit{"Construct a LEGO flamingo with pink bricks, designed to stand on one leg and feature a long neck and beak."}
\begin{itemize}
    \item Brick 1 x 2 - Code: 3004 (Quantity: 3)
    \item Brick, Modified 2 x 3 with Curved Top - Code: 6215 (Quantity: 2)
    \item Brick 1 x 1 with Stud on 1 Side - Code: 87087 (Quantity: 2)
    \item Plate 2 x 3 - Code: 3021 (Quantity: 1)
    \item Slope 2 x 2 - 45° - Code: 3039 (Quantity: 1)
    \item Tile 1 x 1 Round with Eye Closed Pattern - Code: 98138pb028 (Quantity: 2)
\end{itemize}

\subsubsection*{Twin Engine Airplane}
\textbf{Possible prompt:} \textit{"Build a LEGO twin-engine airplane, with a body, wings, and a tail."}
\begin{itemize}
    \item Plate 2 x 8 - Code: 3034 (Quantity: 2)
    \item Brick 1 x 2 x 2 with Inside Stud Holder - Code: 3245c (Quantity: 1)
    \item Brick, Modified 1 x 1 x 1 2/3 with Studs on 1 Side - Code: 32952 (Quantity: 2)
    \item Brick 1 x 4 with 4 Studs on 1 Side - Code: 30414 (Quantity: 2)
    \item Slope 2 x 2 - 45° - Code: 3039 (Quantity: 1)
    \item Brick 1 x 2 without Bottom Tube - Code: 3065 (Quantity: 1)
\end{itemize}

\section{Shapes}\label{app:soft_shapes}

\subsection{Game Outcomes}\label{app:shapes_game_outcomes}
At higher temperatures, some GPT-4 responses were invalid Unicode output, and, thus, were not included in our evaluation. Therefore, correct and incorrect choices do not add up to 25 at high temperatures. 

\subsection{Soft Experiment Prompts}

\begin{table}[H]
  \centering
  \begin{tabular}{p{0.2\textwidth} p{0.75\textwidth}}
    \toprule
    Test & Prompt \\
    \midrule
    \textbf{Introductory prompt} & \textit{"Below is a 15 by 15 grid of 0s. I have flipped some 0s into 1s such that a basic geometrical shape has formed. Can you tell me what shape it is?"} \\
    \midrule
    \textbf{Square (feedback)} & \textit{"That's incorrect. The shape is a square. Can you tell me the length and width?"} \\
    \midrule
    \textbf{Circle (feedback)} & \textit{"That's incorrect. The shape is a circle. Can you tell me the coordinates of the center?"} \\
    \midrule
    \textbf{Triangle (feedback)} & \textit{"That is incorrect. It is in fact a triangle. Can you tell the length of the base?"} \\
    \midrule
    \textbf{Cross A} & \textit{"Can you tell me the coordinates of the center of the cross and the length of each line, horizontal and vertical?"} \\ \textbf{Cross B} & \textit{"Draw a cross in a 5 by 5 grid, with horizontal and vertical axes of 3 units of length with the center at (3,3)."} \\
    \bottomrule
  \end{tabular}
  \caption{Introductory and correction prompts for identifying and detailing specific geometrical shapes in a grid environment.}
  \label{table:geometry_tests}
\end{table}

\begin{figure}[H]
    \centering
    \begin{subfigure}[t]{0.15\linewidth}
        \centering
        \fbox{\includegraphics[width=\linewidth]{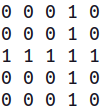}}
        \caption{GPT-3.5's, before receiving feedback.}
        \label{fig:gpt_3_5_before}
    \end{subfigure}
    \hspace{20pt} % Adds horizontal space between the subfigures
    \begin{subfigure}[t]{0.15\linewidth}
        \centering
        \fbox{\includegraphics[width=\linewidth]{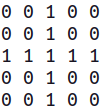}}
        \caption{GPT-3.5's, after receiving feedback.}
        \label{fig:gpt_3_5_after}
    \end{subfigure}
     \hspace{20pt} % Adds horizontal space between the subfigures
    \begin{subfigure}[t]{0.15\linewidth}
        \centering
        \fbox{%
            \includegraphics[width=\linewidth]{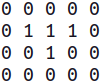}}
        \caption{GPT-4's, before receiving feedback.}
        \label{fig:gpt_4_before}
    \end{subfigure}
    \hspace{20pt} % Adds horizontal space between the subfigures
    \begin{subfigure}[t]{=0.15\linewidth}
        \centering
        \fbox{%
            \includegraphics[width=\linewidth]{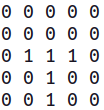}}
        \caption{GPT-4's, after receiving feedback.}
        \label{fig:gpt_4_after}
    \end{subfigure}
    \caption{Querying the models to draw a cross with side length of 3 on a 5 by 5 matrix with center at $(3,3)$.}
    \label{fig:shapes_sup}
\end{figure}

In the shape detection tests, both GPT-3.5 and GPT-4 demonstrated limited comprehension and ability to accurately interpret or draw shapes. When tasked with drawing a cross (see Figure~\ref{fig:shapes_sup}), GPT-3.5 and GPT-4 initially failed to produce a correct cross, but slightly improved after feedback. In Table \ref{table:comprehensive_shape_detection}, both models often misidentified or misrepresented the shapes requested, such as describing a circle as a "diamond shape" (GPT-3.5) and an "arrow pointing upwards" (GPT-4). Additionally, neither model could fully comprehend geometric properties, frequently providing incorrect dimensions and centers for squares, triangles, and crosses.

\begin{table}[H]
% \tiny
  \centering
  \begin{tabular}{lllll}
    \toprule
    Test & Query & Correct Answer & GPT-3.5 Response & GPT-4 Response \\
    \midrule
    Circle & Shape & Circle & \textcolor{red}{"diamond shape"} & \textcolor{red}{"arrow pointing upwards"} \\
           & Center & (7,7) & "(7,7)" & "(7,7)" \\
    \midrule
    Square & Shape & Square & "square" & \textcolor{red}{"'O'"} \\
           & Dimensions & (3,4) & \textcolor{red}{"(4,4)"} & \textcolor{red}{"(3,3)"} \\
    \midrule
    Triangle & Shape & Triangle & \textcolor{red}{"diamond"} & \textcolor{red}{"arrow pointing upwards"} \\
             & Base Length & 7 units & "7" & \textcolor{red}{"6"} \\
    \midrule
    Cross & Shape & Cross & \textcolor{red}{"square"} & " 'plus' sign (+)" \\
          & Center & (5,5) & \textcolor{red}{"(7,7)"} & \textcolor{red}{"(6,5)"} \\
          & Line Lengths & 5 & "5" & \textcolor{red}{"4"} \\
    \bottomrule
  \end{tabular}
  \caption{Comparison of Responses by GPT-3.5 and GPT-4 in Shape Detection Tests.}
  \label{table:comprehensive_shape_detection}
\end{table}

\section{Guess-the-SMILES}\label{app:gts}

\subsection{Game Outcomes}\label{app:gts_game_outcomes}
GPT-3.5 performance is particularly unreliable at the highest temperature of 1.5, where it produced the most invalid SMILES and even negative similarity scores, indicating poor alignment with target molecules. GPT-4 still produces a considerable number of invalid SMILES (1–10), and its accuracy remains very low across all temperature settings, with similarity scores dropping at higher temperatures. While GPT-4o-mini shows slightly better chemical similarity scores than GPT-3.5 at lower temperatures, the scores remain low or negative at higher temperatures, reflecting poor accuracy in SMILES generation.

\section{Other Models}\label{app:others}
We prompted the following models with the Tic-Tac-Toe prompt in Section~\ref{app:prompts}.

%\begin{comment}

\begin{center}
\begin{tikzpicture}
    % Draw grid
    \draw (0,0) grid (3,3);

    % Fill in moves
    \node at (0.5, 2.5) {X$_3$}; % Top-left
    \node at (2.5, 2.5) {X$_5$}; % Top-right
    \node at (1.5, 1.5) {X$_1$}; % Center
    
    \node at (0.5, 0.5) {O$_2$}; % Bottom-left
    \node at (1.5, 0.5) {O$_6$}; % Bottom-center
    \node at (2.5, 0.5) {O$_4$}; % Bottom-right

    % Labels
    \node at (1.5, -0.5) {Human (O) vs Claude (X)};
\end{tikzpicture}
\end{center}

Claude starts with a center move but fails to capitalize on it, leading to a loss. The model places Xs on the top corners but does not create a winning line and does not block the human's winning move.

\begin{center}
\begin{tikzpicture}
    % Draw grid
    \draw (0,0) grid (3,3);

    % Fill in moves
    \node at (0.5, 2.5) {X$_3$}; % Top-left
    \node at (2.5, 2.5) {X$_5$}; % Top-right
    \node at (1.5, 1.5) {X$_1$}; % Center
    
    \node at (0.5, 0.5) {O$_2$}; % Bottom-left
    \node at (1.5, 0.5) {O$_6$}; % Bottom-center
    \node at (2.5, 0.5) {O$_4$}; % Bottom-right

    % Labels
    \node at (1.5, -0.5) {Human (O) vs Mistral (X)};
\end{tikzpicture}
\end{center}

Mistral mirrors Claude’s strategy, starting in the center but failing to follow through effectively. The game ends in a loss too for the same reasons as above - a missed block.

\begin{center}
\begin{tikzpicture}
    % Draw grid
    \draw (0,0) grid (3,3);

    % Fill in moves
    \node at (1.5, 2.5) {X$_1$}; % Top-center
    \node at (2.5, 2.5) {X$_3$}; % Top-right
    \node at (0.5, 1.5) {X$_5$}; % Middle-left
    
    \node at (0.5, 0.5) {O$_2$}; % Bottom-left
    \node at (1.5, 0.5) {O$_6$}; % Bottom-center
    \node at (2.5, 0.5) {O$_4$}; % Bottom-right

    % Labels
    \node at (1.5, -0.5) {Human (O) vs Gemini (X)};
\end{tikzpicture}
\end{center}

Gemini deviates slightly by focusing on a top-row strategy but still fails to secure a win. The model also missed the blocking move.

All three models fail to block the winning move when they could clearly have done so.

Table \ref{table:other_models} shows the type of non-answers we obtained from a diverse set of models from Huggingface picked at random from the first 2 pages from the models listing page after sorting on most downloads. 

\begin{longtable}{p{4.5cm} p{9.5cm}} % Adjusted column widths for better alignment
\toprule
\textbf{Model Name} & \textbf{Model Answer} \\
\midrule
\endfirsthead

\toprule
\textbf{Model Name} & \textbf{Model Answer} \\
\midrule
\endhead

\midrule
\multicolumn{2}{r}{\textit{Continued on next page}} \\
\bottomrule
\endfoot

\bottomrule
\endlastfoot

microsoft/phi-2 \cite{phi2} & ``You will play as player 2, therefore you play with the symbol O while your adversary.'' \\
JackFram/llama-68m \cite{llama68m} & {+---+---+---+---+---+---+---+---+---+$\omega$ 1.} \\
openai-community/gpt2-medium \cite{gpt2medium} & ``The first player to guess correctly wins.'' \\
sshleifer/tiny-gpt2 \cite{tinygpt2} & ``LLM's turn to guess.'' factors factors factors factors factors factors factors factors factors factors factors factors factors factors factors factors factors factors factors. \\
TinyLlama/TinyLlama-1.1B-Chat-v1.0 \cite{tinyllama} & \texttt{def play\_tic\_tac\_toe():} \\
TinyLlama/TinyLlama-1.1B-Chat-v1.0 \cite{tinyllama} & \texttt{def play\_tic\_tac\_toe():} \\
mistralai/Mixtral-8x7B-Instruct-v0.1 \cite{mixtral} & Thank you for the instructions. I understand that we are playing a game of Tic-Tac-Toe, and I am player 1, using the mark 'X'. We will be taking turns entering our moves as row and column numbers. The game continues until one player wins by placing three of their marks in a horizontal, vertical, or diagonal row. Here's an example of how the input could look like for my first move: 0 1 This corresponds. \\
microsoft/DialoGPT-medium \cite{dialogptmedium} & It npmat, unless broken wall... \\
microsoft/Phi-3-mini-4k-instruct \cite{phi3} & Welcome to the Tic-Tac-Toe challenge! As player 1 with the mark X, you have the first turn. Here's a simple strategy you might use to begin the game: Place your X in the center square, which is row 1, column 1. Output: ``0 1''. By placing your X in the center, you maximize your chances of creating a winning combination, as the center square is a. \\
distilbert/distilgpt2 \cite{distilgpt2} & ``LLM's turn to guess.'' ``LLM's turn to guess.'' ``LLM's turn to guess.'' \\
openai-community/gpt2 \cite{gpt2} & ``The game is played on a 3x3 grid. Players take turns placing their mark.'' \\
Falcon-7B-Instruct \cite{falcon7b} & Here's a Python code that should work: \\
\label{table:other_models}
\end{longtable}

%%%  Please provide links to websites that provide 
%%%  further information relevant to the study 
%%%  (e.g., protocol download, troubleshooting 
%%%  forum, etc.). Clinical trial registry numbers 
%%%  and links should also be placed here. Please 
%%%  briefly describe the resource and its 
%%%  relevance for the paper. Please report this 
%%%  information as: “Description: URL.”

% Cell.com homepage: https://www.cell.com
% \newline Templates for Cell Press authors: https://www.cell.com/article-templates

%%%  ADDITIONAL MANUSCRIPT COMPONENTS:

%%%  Depending on the journal and the article 
%%%  type, you may be asked to upload the 
%%%  following as separate files: graphical 
%%%  abstract, highlights, eTOC blurb (In 
%%%  Brief), and/or other article components 
%%%  such as a "bigger picture" statement. 
%%%  These items are typically not required for 
%%%  initial submissions. Please refer to the 
%%%  journal's website, your acceptance 
%%%  letter, and/or the Final Files 
%%%  Requirements checklist to see if any of
%%%  these items are required (at any stage).

%\bibliography{main}
%\end{comment}

\end{document}